\documentclass[a4paper, 11pt]{article}

\usepackage[margin=1in]{geometry}
\usepackage{url}
\usepackage{graphicx}
\usepackage{caption}
\usepackage{booktabs}
\usepackage{xcolor}
\usepackage{colortbl}
\usepackage{graphicx}
\usepackage{subcaption}
\usepackage{amsmath, amssymb}
\usepackage{wrapfig}
\usepackage{float}
\usepackage{bm}
\usepackage{ulem}
\usepackage{titletoc}
\usepackage{tabularx}
\usepackage[T1]{fontenc}
\usepackage{helvet}
\usepackage{multicol}

\usepackage[numbers,sort&compress]{natbib}
\usepackage[bottom]{footmisc}
\usepackage{enumitem}
\usepackage{siunitx}

\usepackage{etoolbox}
\definecolor{primalcolor}{HTML}{960000}
\definecolor{contrarydarkcolor}{HTML}{006666}
\definecolor{irongray}{HTML}{6D6E71}
\definecolor{steelgray}{HTML}{F1F4F7}
\definecolor{releaseBlockColor}{HTML}{FFFFFF}
\definecolor{abcRed}{HTML}{C8102E}
\definecolor{abcGreen}{HTML}{2E7D32}
\definecolor{abcYellow}{HTML}{F2B705}
\newcommand{\linkcolor}{contrarydarkcolor}
\newcommand{\urlcolor}{contrarydarkcolor}
\newcommand{\citecolor}{contrarydarkcolor}
\newcommand{\authormarks}[1]{\textsuperscript{#1}}

\newcommand{\abcIcon}[2]{%
  \begin{tikzpicture}[x=0.16in,y=0.16in,baseline=-0.03in]
    \draw[#1,line width=0.8pt,rounded corners=0.012in] (0.06,0.06) rectangle (0.94,0.94);
    \node[inner sep=0pt] at (0.5,0.48)
      {\textcolor{#1}{\fontfamily{ptm}\bfseries\fontsize{9.5}{9.5}\selectfont #2}};
  \end{tikzpicture}%
}
\newcommand{\abcTitleIcons}{%
  \abcIcon{abcRed}{A}\hspace{0.25em}%
  \abcIcon{abcGreen}{B}\hspace{0.25em}%
  \abcIcon{abcYellow}{C}%
}

\newsavebox{\lecartitleblockbox}
\newcommand{\lecartitleblock}[1]{%
  \par\noindent
  \begingroup
  \setlength{\fboxsep}{10pt}%
  \begin{lrbox}{\lecartitleblockbox}%
  \begin{minipage}{\dimexpr\linewidth-2\fboxsep\relax}%
  \maketitle
  #1
  \lecarprintmetadata
  \end{minipage}%
  \end{lrbox}%
  \colorbox{releaseBlockColor}{\usebox{\lecartitleblockbox}}%
  \endgroup
  \par\vspace{0.45em}
}

\newcommand{\lecarseparator}{%
  \par\smallskip
  {\color{irongray}\noindent\rule{\linewidth}{0.4pt}}%
  \par\smallskip
}

\makeatletter

\renewcommand{\maketitle}{%
  \begingroup
  \raggedright
  {\abcTitleIcons\par}
  \vskip -0.25em
  {\LARGE\bfseries \@title \par}
  \vskip 1em
  {\@author\par}
  \endgroup
  \par
  \vskip 1.4em
}

\newcommand\addtometadatalist[5][]{%
  \begingroup
  \if\relax#3\relax\def\sep{}\else\def\sep{#5}\fi
  \let\protect\@unexpandable@protect
  \xdef#3{\expandafter{#3}\sep #4[#1]{#2}}%
  \endgroup
}
\newcommand\metadatalist{}
\newcommand\metadataformat[2][]{{\small \textbf{#1:} #2}}
\newcommand\metadata[2][]{\addtometadatalist[#1]{#2}{\metadatalist}{\metadataformat}{\\}}
\newcommand{\paperwebsite}[1]{\metadata[Website]{\url{#1}}}
\newcommand{\lecarprintmetadata}{%
  \ifdefempty{\metadatalist}{}{\lecarseparator\metadatalist\par}%
}
\makeatother

\newcommand{\lecarAuthorBlock}{%
  \begin{minipage}{\linewidth}
  \raggedright
  {\scriptsize
  \mbox{\textbf{Arthur Allshire}\authormarks{1*},
  \textbf{Himanshu Gaurav Singh}\authormarks{1*},
  \textbf{Ritvik Singh}\authormarks{1*},
  \textbf{Adam Rashid}\authormarks{2*},
  \textbf{Hongsuk Choi}\authormarks{1*},
  \textbf{David McAllister}\authormarks{1*},}\par
  Justin Yu\authormarks{1,4},
  Yiyuan Chen\authormarks{4},
  Huang Huang\authormarks{4},
  Pieter Abbeel\authormarks{1,3},
  Xi Chen\authormarks{3},
  Rocky Duan\authormarks{3},
  Phillip Isola\authormarks{2},
  Jitendra Malik\authormarks{1,3},\par
  Fred Shentu\authormarks{1,4},
  Guanya Shi\authormarks{3,5},
  Philipp Wu\authormarks{4},
  Angjoo Kanazawa\authormarks{1,3}\par}
  \end{minipage}%
}

\newcommand{\releaseAffiliationsFootnote}{%
  \begingroup
  \renewcommand{\thefootnote}{}%
  \footnotetext{%
    \authormarks{*}Core Contributor; work performed during an internship at Amazon FAR.\\
    \authormarks{1}UC Berkeley \quad
    \authormarks{2}MIT \quad
    \authormarks{3}Amazon FAR \quad
    \authormarks{4}XDOF \quad
    \authormarks{5}Carnegie Mellon University%
  }%
  \endgroup
}

\newcommand{\releaseAbstractText}{%
We introduce ABC, a fully open-source stack for manipulation with behavior cloning. At its core is ABC-130K: the largest open-source teleoperation dataset to date, featuring 3,500 hours of data spanning over 130K episodes across 195 diverse tasks. Furthermore, we open-source our accessible hardware setup, training infrastructure, and simulation pipeline. We also release 400 hours of sim-teleop data and provide a co-training recipe that produces correlated simulation and real-world evaluation, offering a reliable proxy for ablating model-design and training decisions before costly real-world evaluation. 
We explore various training recipes and compare common architectural choices for Diffusion Transformers (DiT) and Vision-Language-Action (VLA) models, grounding our findings in real-world evaluations. The resulting policies successfully execute dexterous tasks such as box folding and extracting credit cards from wallets. By providing a reproducible toolkit, we aim to place researchers on an equal footing, establishing the necessary foundation to learn the ABCs of Behavior Cloning together as a community.%
}

\newcommand{\releaseTitleAbstract}{%
  \vspace{0.4em}
  {\small\textbf{Abstract.} \releaseAbstractText\par}
}

\newlength{\releaseTeaserWidth}
\newlength{\releaseTeaserCellX}
\newlength{\releaseTeaserCellY}
\newcommand{\releaseTeaserFigure}{%
  \begin{center}
    \setlength{\releaseTeaserWidth}{1.0\linewidth}
    \setlength{\releaseTeaserCellX}{\dimexpr\releaseTeaserWidth/4\relax}
    \setlength{\releaseTeaserCellY}{\dimexpr\releaseTeaserWidth/6\relax}
    \begin{tikzpicture}[
        x=\releaseTeaserCellX,
        y=\releaseTeaserCellY,
        inner sep=0pt,
        outer sep=0pt
    ]
        \node[anchor=south west] at (0, 2)
            {\includegraphics[width=\dimexpr2\releaseTeaserCellX\relax,
                              height=\dimexpr2\releaseTeaserCellY\relax]
                {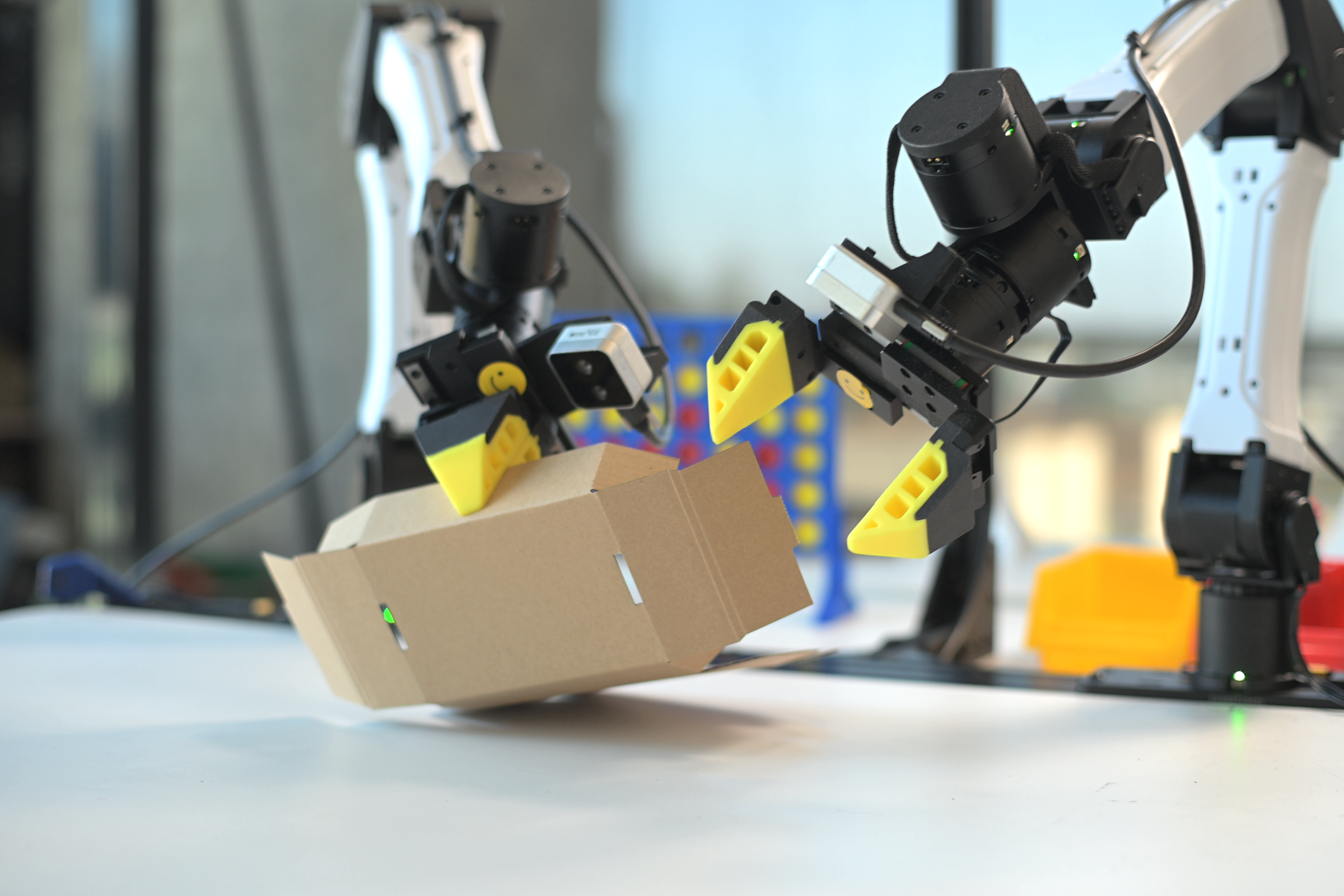}};
        \node[anchor=south west] at (2, 2)
            {\includegraphics[width=\dimexpr2\releaseTeaserCellX\relax,
                              height=\dimexpr2\releaseTeaserCellY\relax]
                {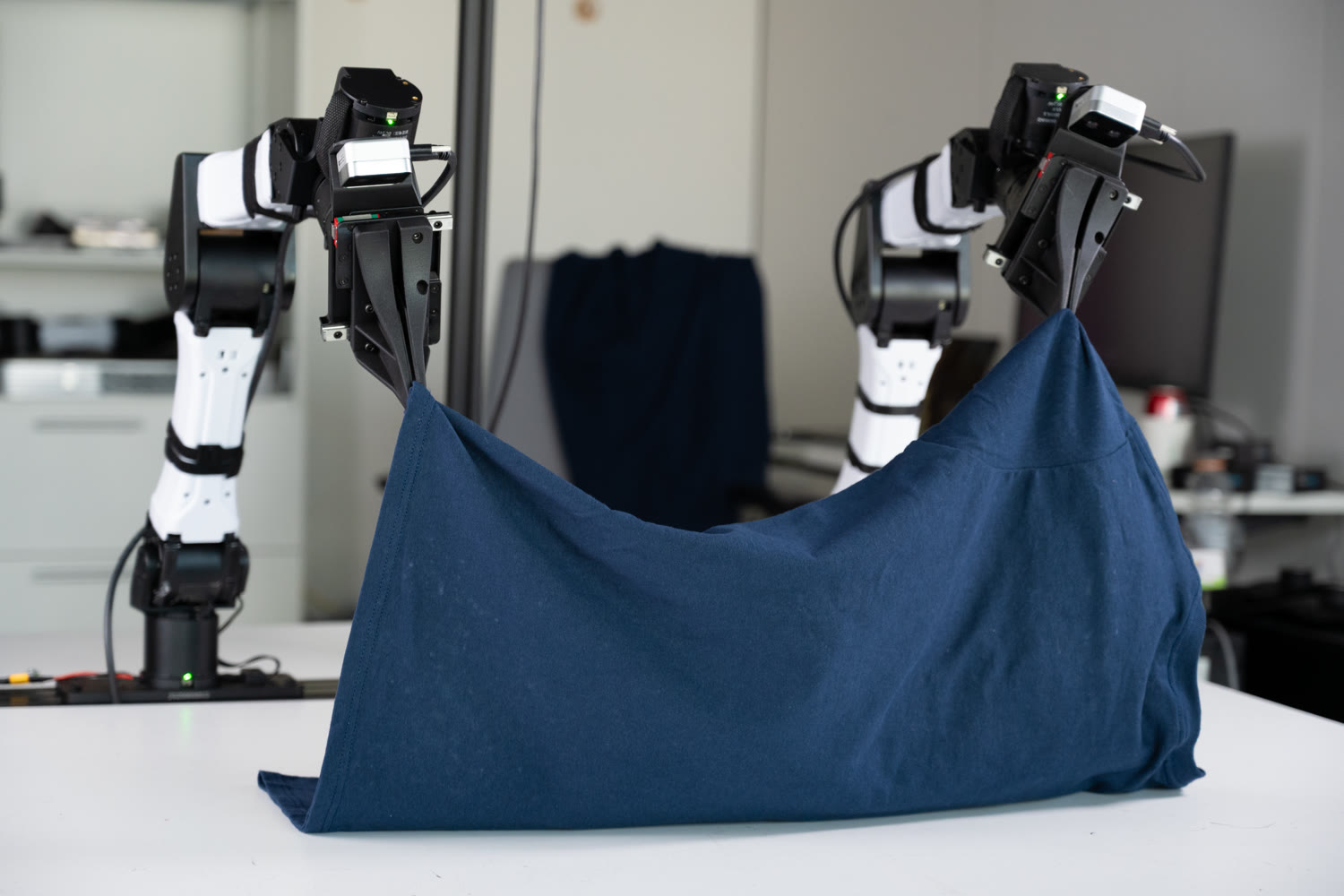}};
        \node[anchor=south west] at (0, 1)
            {\includegraphics[width=\releaseTeaserCellX,
                              height=\releaseTeaserCellY]
                {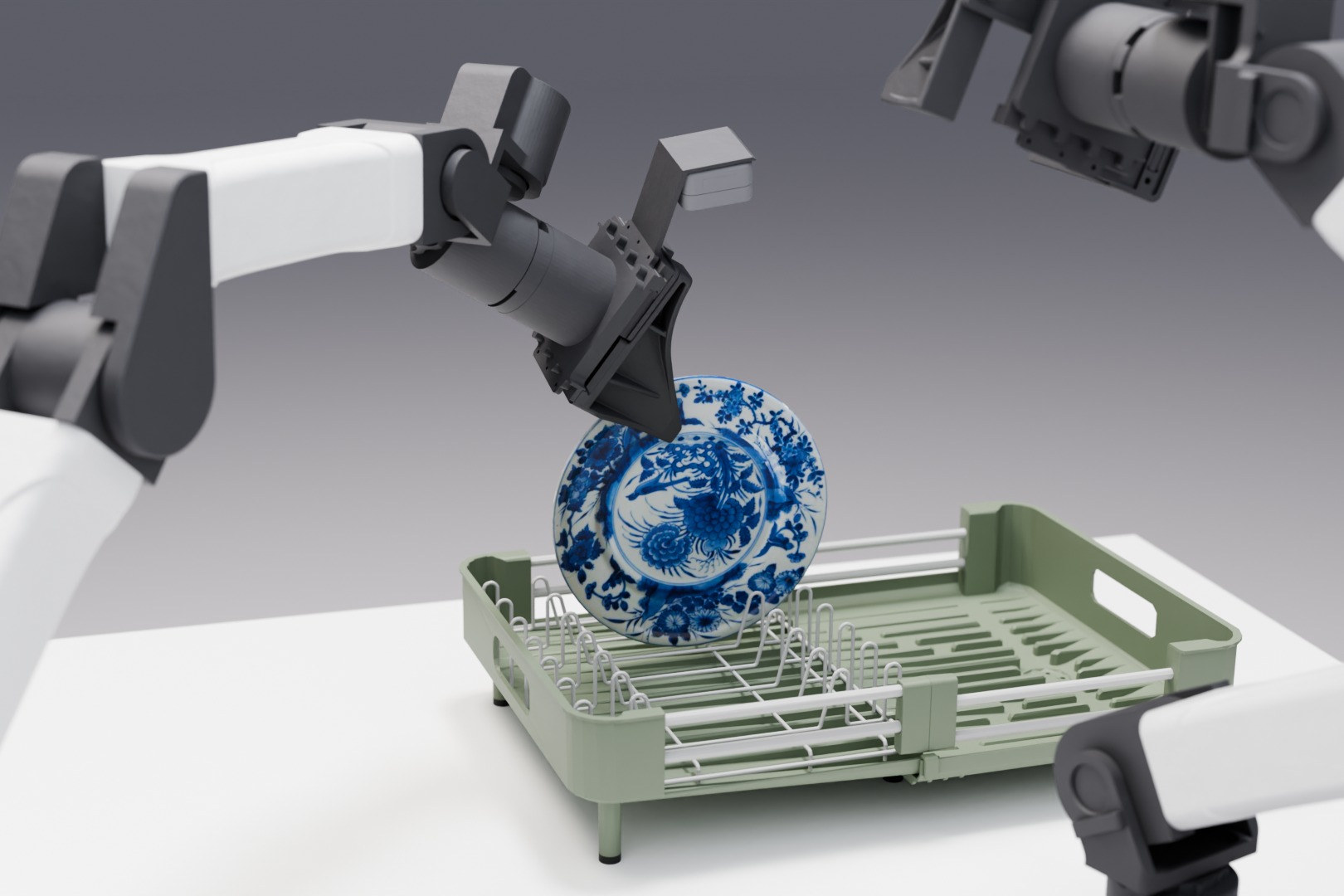}};
        \node[anchor=south west] at (1, 1)
            {\includegraphics[width=\releaseTeaserCellX,
                              height=\releaseTeaserCellY]
                {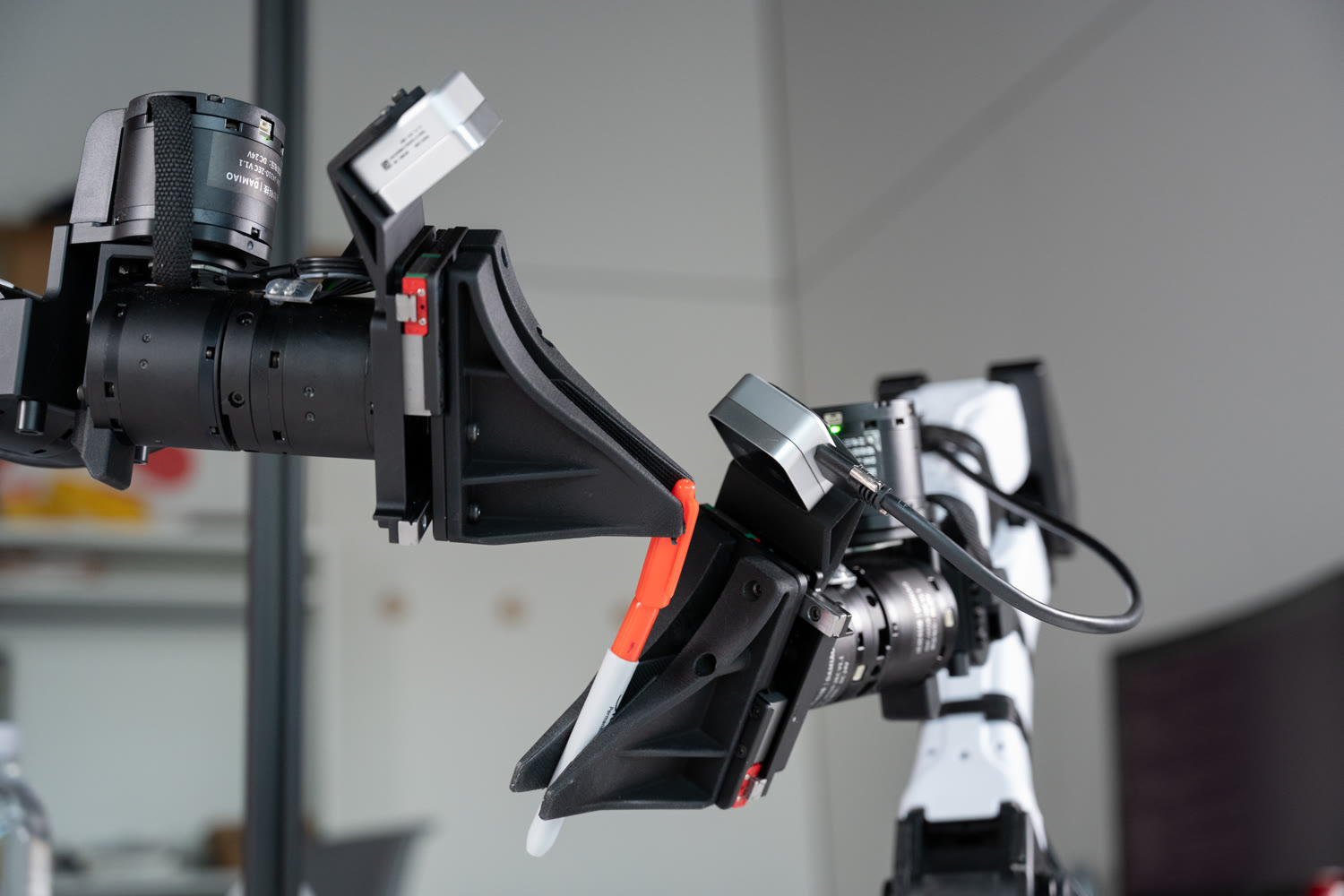}};
        \node[anchor=south west] at (2, 1)
            {\includegraphics[width=\releaseTeaserCellX,
                              height=\releaseTeaserCellY]
                {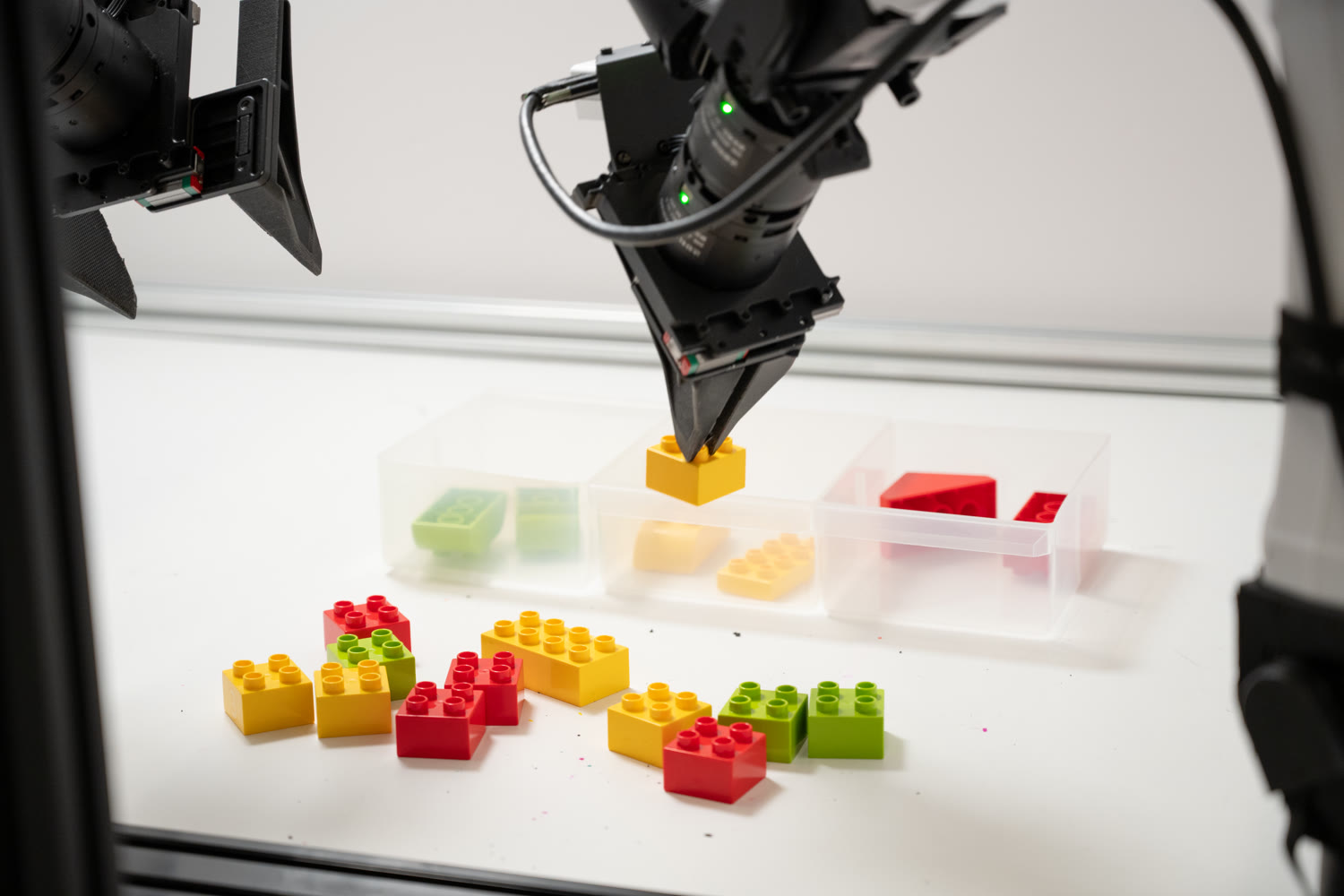}};
        \node[anchor=south west] at (3, 1)
            {\includegraphics[width=\releaseTeaserCellX,
                              height=\releaseTeaserCellY]
                {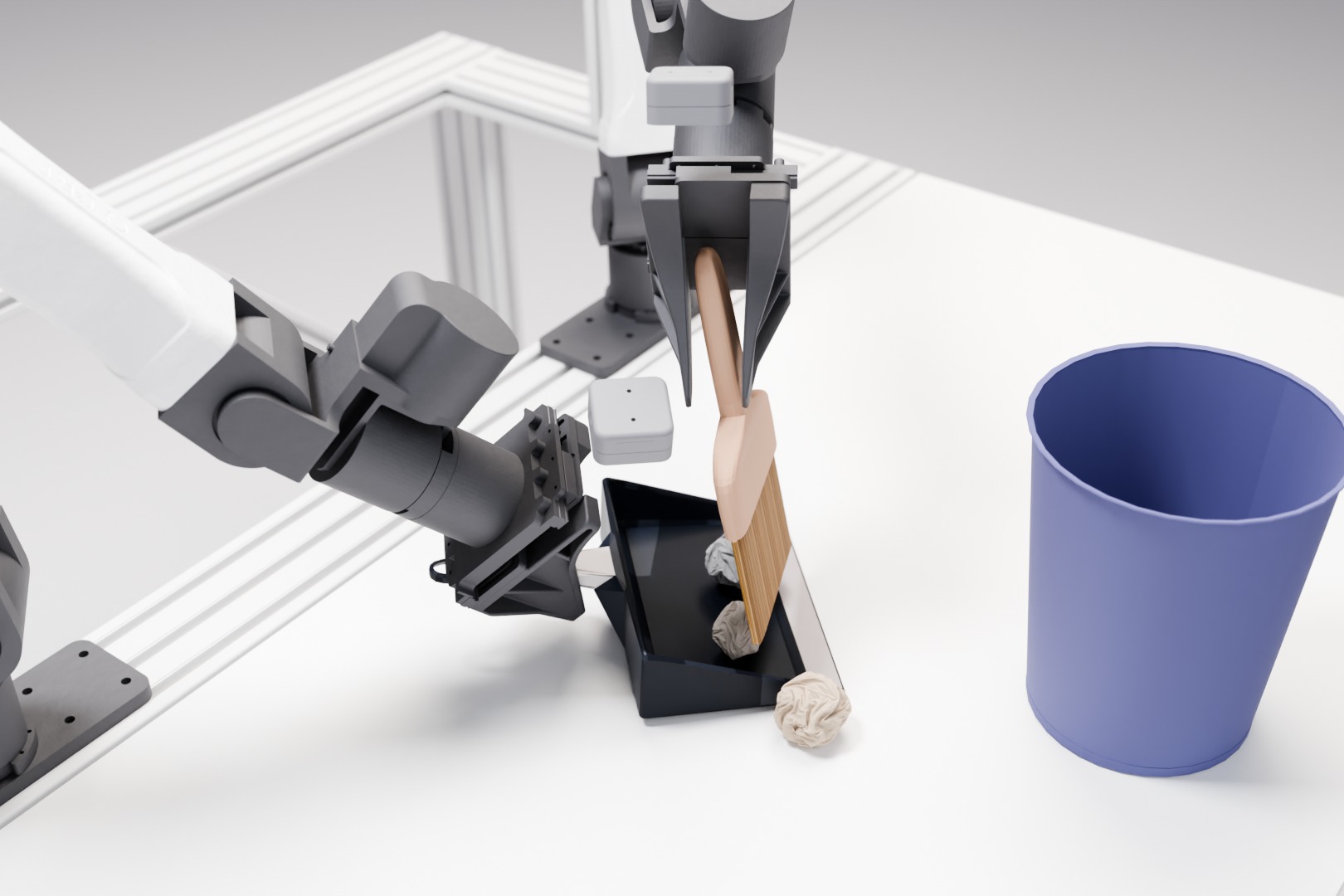}};
        \node[anchor=south west] at (0, 0)
            {\includegraphics[width=\releaseTeaserCellX,
                              height=\releaseTeaserCellY]
                {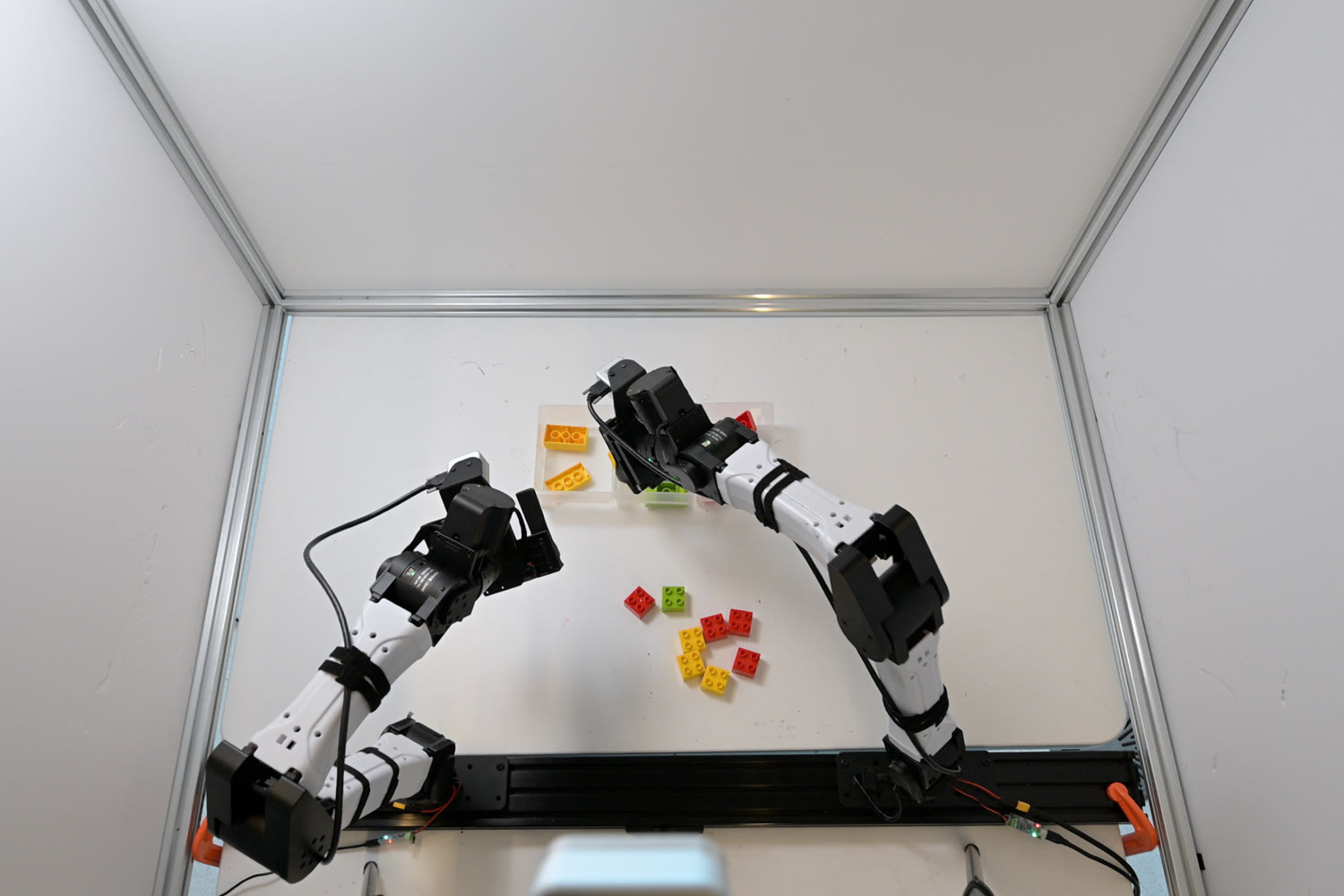}};
        \node[anchor=south west] at (1, 0)
            {\includegraphics[width=\releaseTeaserCellX,
                              height=\releaseTeaserCellY]
                {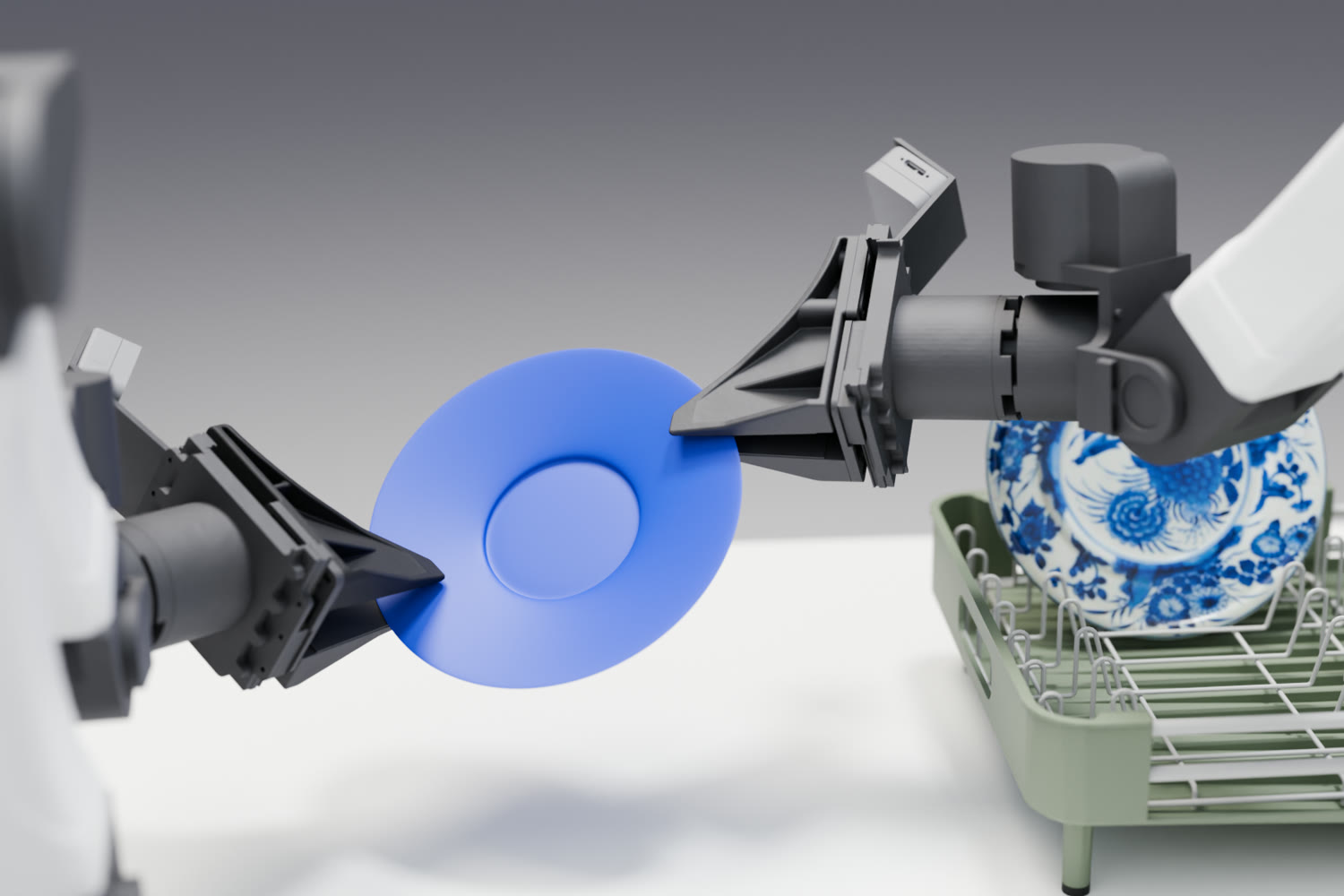}};
        \node[anchor=south west] at (2, 0)
            {\includegraphics[width=\releaseTeaserCellX,
                              height=\releaseTeaserCellY]
                {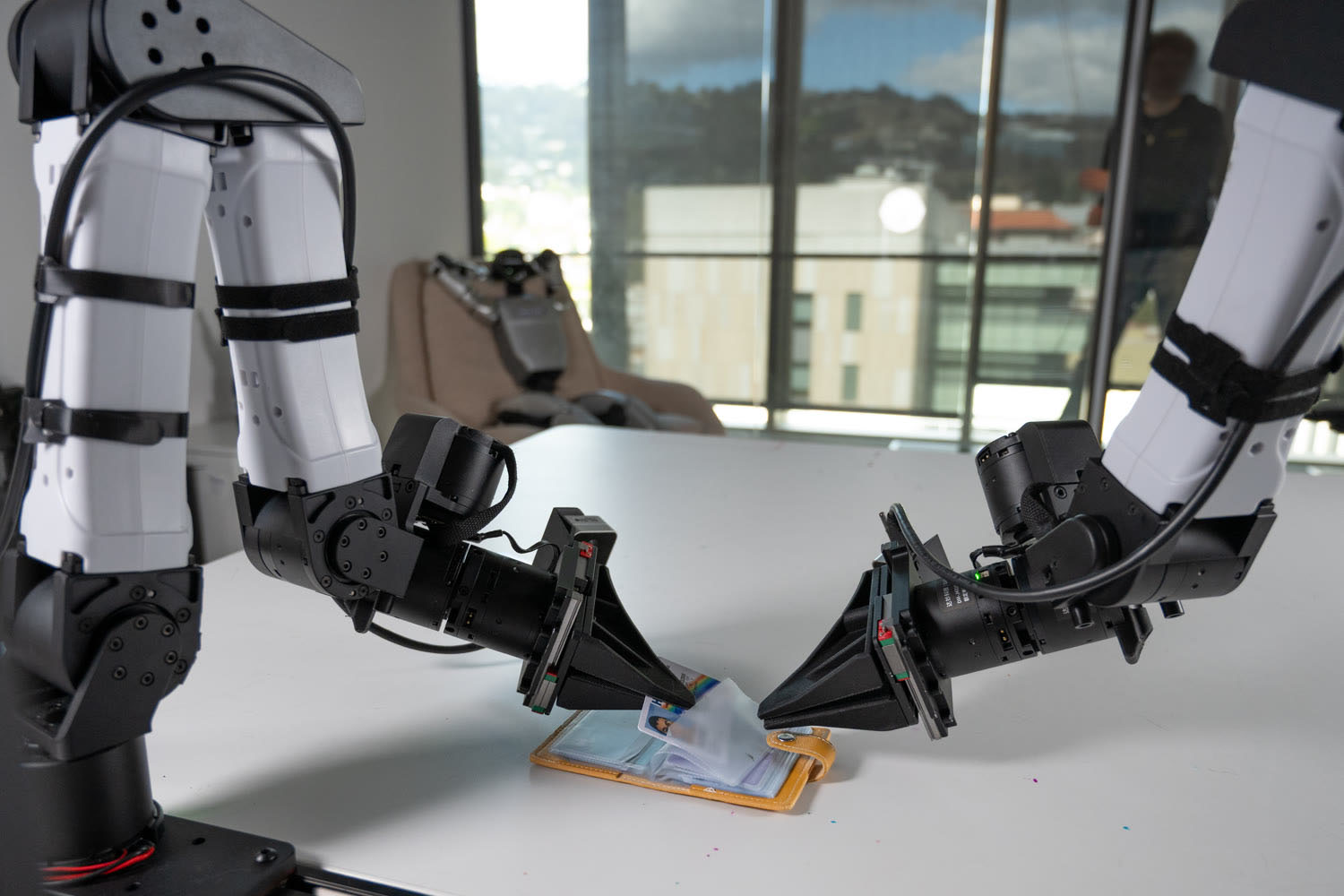}};
        \node[anchor=south west] at (3, 0)
            {\includegraphics[width=\releaseTeaserCellX,
                              height=\releaseTeaserCellY]
                {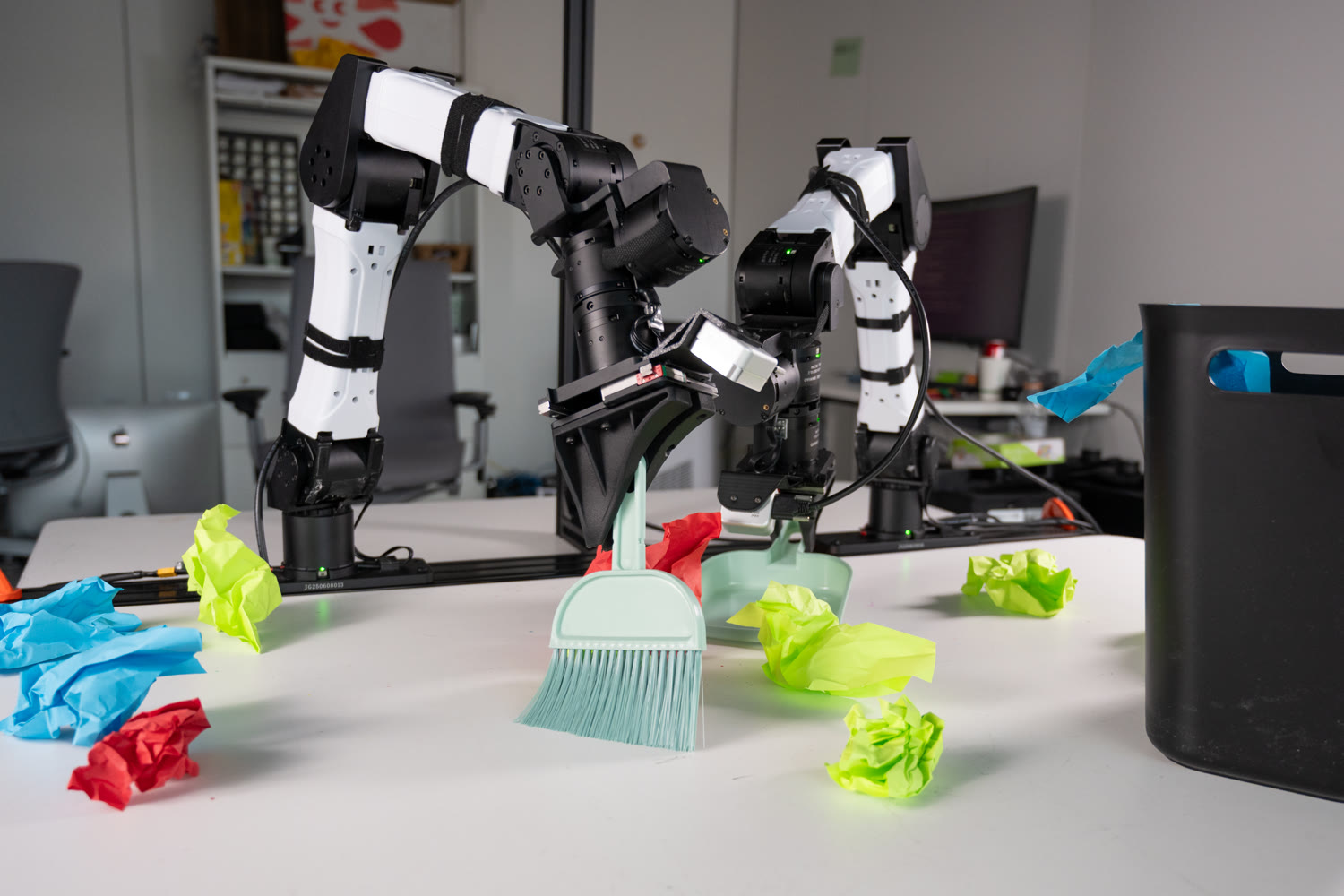}};
    \end{tikzpicture}
    {\captionsetup{font=small,labelformat=empty}
    \captionof{figure}{Policies trained on ABC-130K being rolled out. More videos at \url{https://abc.bot}}
    \label{fig:teaser}}
  \end{center}
  \vspace{-0.2em}
}

\usepackage[
  colorlinks=true,
  linkcolor=\linkcolor,
  urlcolor=\urlcolor,
  citecolor=\citecolor,
  breaklinks
]{hyperref}

\usepackage{pgfplots,pgfplotstable}
\usetikzlibrary{calc}
\usepgfplotslibrary{groupplots}
\pgfplotsset{compat=1.18}
\pgfplotsset{
  tick label style={font=\scriptsize},
  label style={font=\small},
  legend style={font=\scriptsize},
  title style={font=\small},
  tick style={draw=none},
  major grid style={gray!35},
  every axis plot/.append style={line width=0.75pt, mark size=1.25pt},
}
\definecolor{LBMColor}{rgb}{0.121569,0.466667,0.705882}
\definecolor{VLAColor}{rgb}{0.839216,0.152941,0.156863}
\definecolor{BSLow}{rgb}{1.0,0.498039,0.054902}
\definecolor{BSMid}{rgb}{0.172549,0.627451,0.172549}
\definecolor{BSHigh}{rgb}{0.121569,0.466667,0.705882}
\definecolor{BSOther}{rgb}{0.498039,0.498039,0.498039}
\definecolor{DivOne}{rgb}{0.121569,0.466667,0.705882}
\definecolor{DivTwo}{rgb}{1.0,0.498039,0.054902}
\definecolor{MultTwo}{rgb}{0.839216,0.152941,0.156863}
\definecolor{PrettyText}{rgb}{0.105882,0.105882,0.105882}
\definecolor{PrettyDivOne}{rgb}{0.105882,0.105882,0.105882}
\definecolor{PrettyDivTwo}{rgb}{0.356863,0.541176,0.447059}
\definecolor{PrettyMultTwo}{rgb}{0.780392,0.317647,0.278431}
\definecolor{NPGBlueLight}{HTML}{B7CEE6}
\definecolor{NPGBlueMid}{HTML}{6E94C4}
\definecolor{NPGBlueDark}{HTML}{2E558E}
\definecolor{DiTSColor}{rgb}{0.839216,0.152941,0.156863}
\definecolor{DiTBColor}{rgb}{1.000000,0.498039,0.054902}
\definecolor{DiTLColor}{rgb}{0.172549,0.627451,0.172549}
\definecolor{DiTxLColor}{rgb}{0.121569,0.466667,0.705882}
\usetikzlibrary{patterns}

\makeatletter
\newcommand{\tightintrosection}[1]{%
  \@startsection{section}{1}{\z@}%
                {-0.8ex \@plus -0.2ex \@minus -0.1ex}%
                {0.35ex \@plus 0.1ex \@minus 0.1ex}%
                {\large\bf\raggedright}%
                {#1}%
}
\makeatother

\makeatletter
  \newcommand{\tightcapssubsection}[1]{%
    \@startsection{subsection}{2}{\z@}%
                  {-0.7ex \@plus -0.2ex \@minus -0.1ex}%
                  {0.25ex \@plus 0.1ex \@minus 0.1ex}%
                  {\normalsize\bf\raggedright}%
                  {#1}%
}
\makeatother

\makeatletter
\renewcommand\section{\@startsection{section}{1}{\z@}%
  {-1.2ex \@plus -0.3ex \@minus -0.2ex}%
  {0.5ex \@plus 0.2ex \@minus 0.1ex}%
  {\large\bf\raggedright}}
\renewcommand\subsection{\@startsection{subsection}{2}{\z@}%
  {-1.0ex \@plus -0.3ex \@minus -0.1ex}%
  {0.4ex \@plus 0.1ex \@minus 0.1ex}%
  {\normalsize\bf\raggedright}}
\renewcommand\subsubsection{\@startsection{subsubsection}{3}{\z@}%
  {-0.8ex \@plus -0.2ex \@minus -0.1ex}%
  {0.3ex \@plus 0.1ex \@minus 0.1ex}%
  {\normalsize\bfseries\raggedright}}
\makeatother

\title{
Scalable Behavior Cloning with\\Open Data, Training, and Evaluation
}

\author{
  \lecarAuthorBlock
}

\begin{document}

\lecartitleblock{\releaseTeaserFigure\releaseTitleAbstract}
\releaseAffiliationsFootnote
\begin{figure}[H]
    \centering
    \newlength{\fdw}\setlength{\fdw}{0.24\textwidth}
    \newlength{\fdh}\setlength{\fdh}{0.75\fdw}
    \newcommand{\fdcell}[3]{%
        \begin{minipage}[t]{\fdw}%
            \centering
            \parbox[c][0.34cm][c]{\fdw}{\centering\bfseries #1\par}\\[2pt]
            \includegraphics[width=\fdw, height=\fdh]{#2}\\[3pt]
            {\footnotesize #3\par}%
        \end{minipage}%
    }
    \fdcell{ABC-130K}{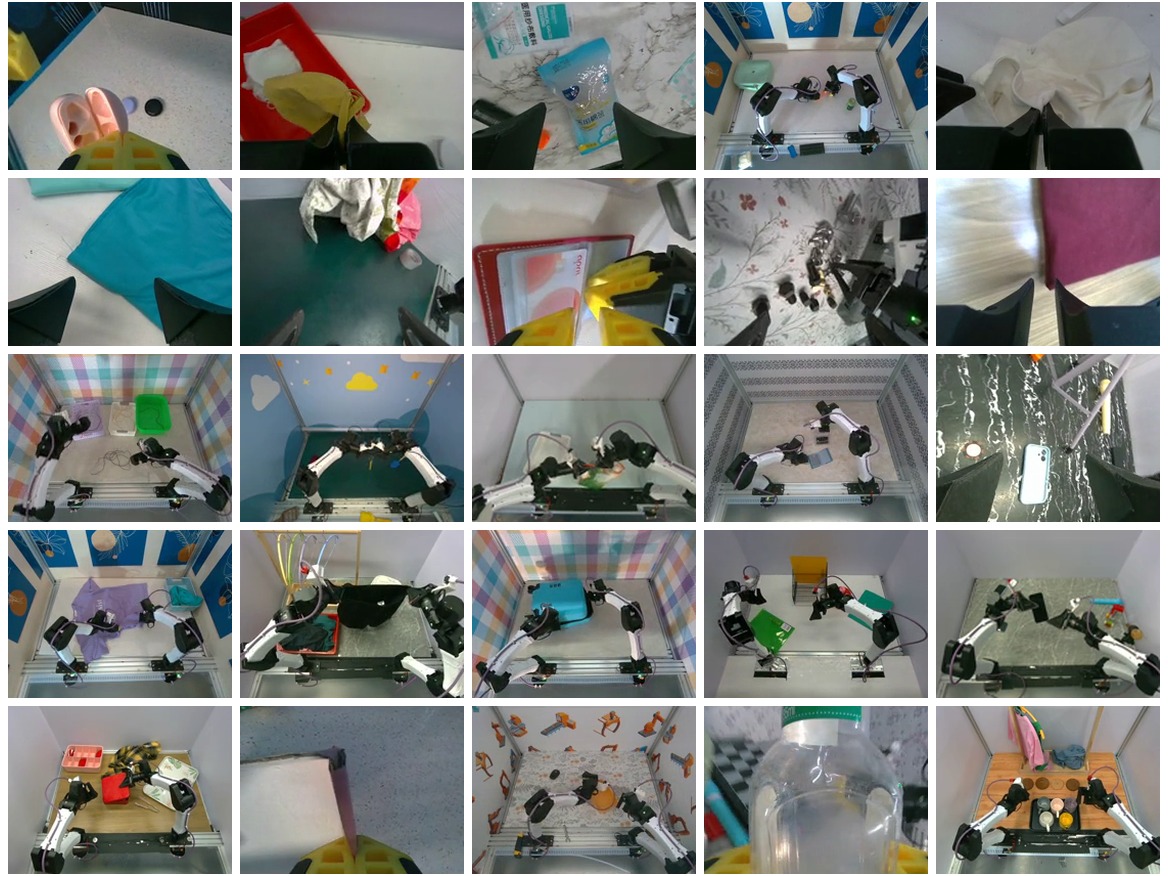}{\textit{3.5K hours of real-world teleop-data across 130K trajectories and 195 tasks.}}\hfill
    \fdcell{ABC-Models}{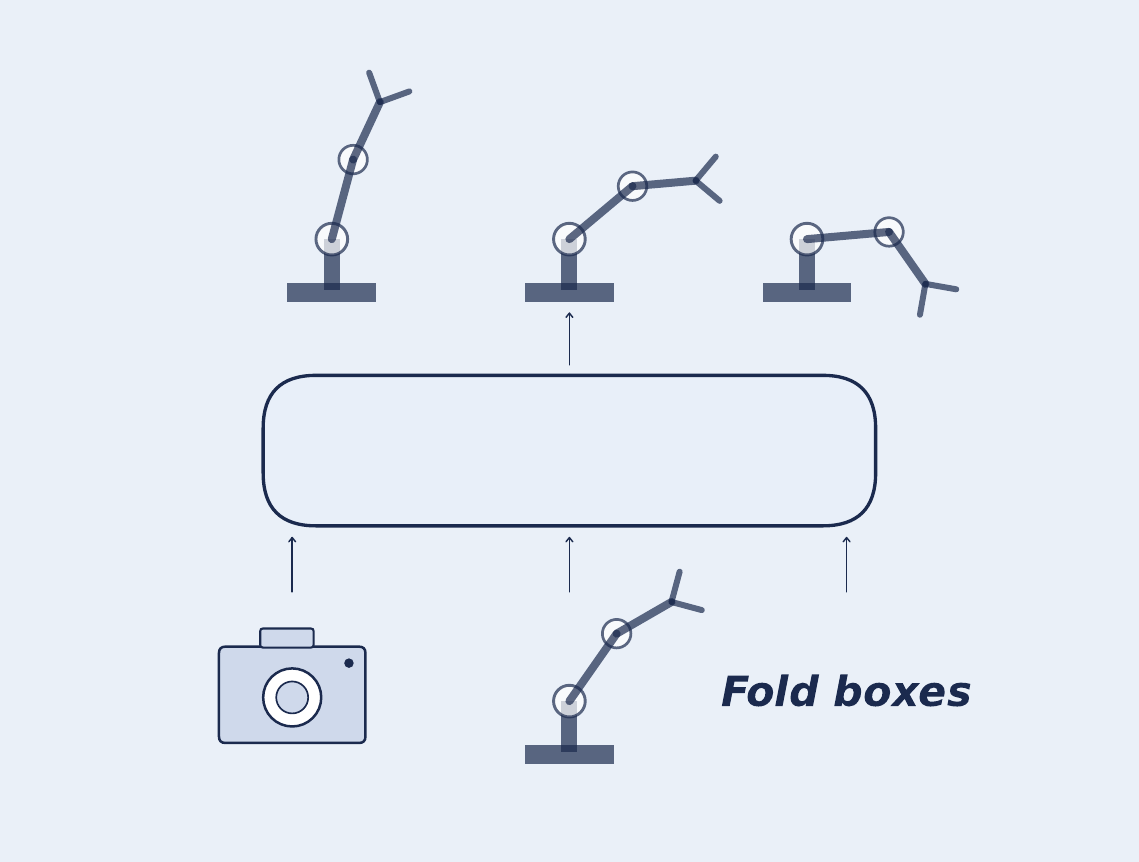}{\textit{DiT $\&$ VLA training, with real-world ablations over architecture and compute.}}\hfill
    \fdcell{ABC-Sim}{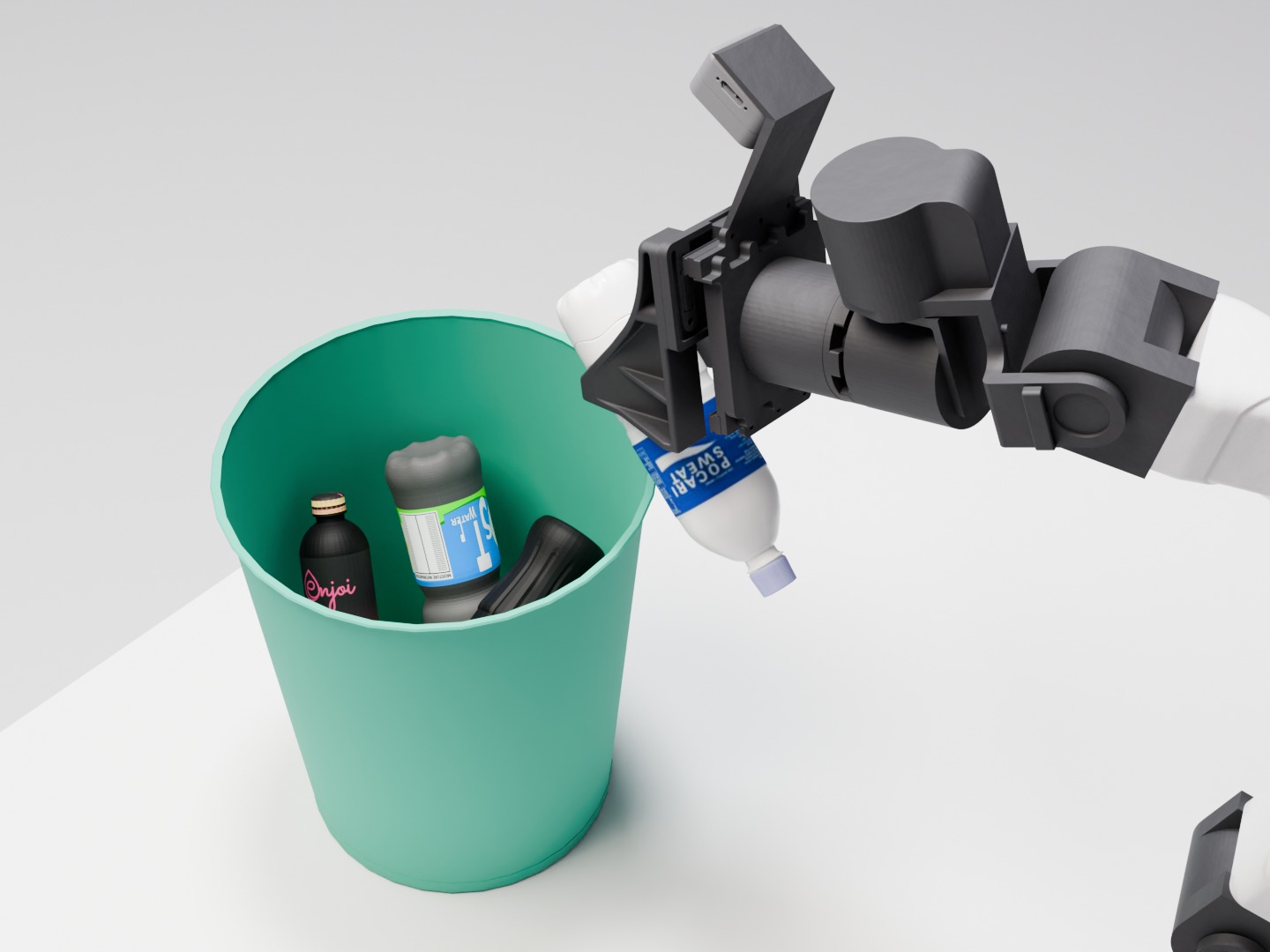}{\textit{400 hours of simulation teleop-data across 20 tasks for sim-to-real correlation.}}\hfill
    \fdcell{ABC-Eval}{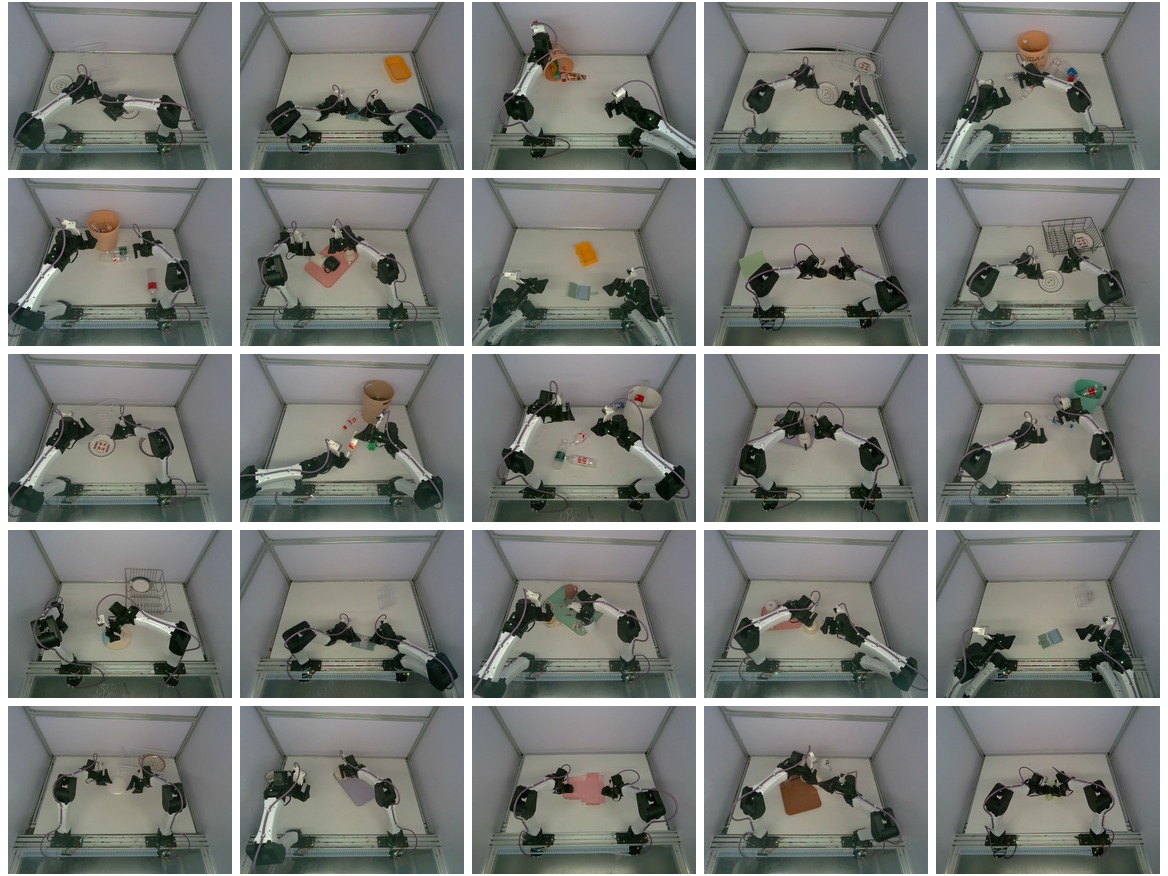}{$>$\textit{100 hours of real-world policy rollouts, released with evaluation scores}}
    \caption{\textbf{An overview of the ABC stack.} ABC-130K provides large-scale real-world teleoperation data. ABC-Models instantiates DiT and VLA policies and studies architecture and compute-scaling choices through real-world ablations. ABC-Sim provides simulation environments and data for studying sim-to-real correlation. ABC-Eval provides a large-scale real-world evaluation suite with rollouts and rubrics. The full stack will be released for the community to build upon.}
    \label{fig:abc-stack}
\end{figure}

\tightintrosection{Introduction}

The stack for behavior cloning (BC) is deep. In order to make progress, a practitioner must first set up robot hardware and collect demonstrations. Policy learning on this data requires both an efficient training system over heterogeneous data streams and an effective learning recipe, which the community has yet to converge on. The resulting pretrained policy may be sufficient for simple tasks, but more complex skills require further iteration, such as DAgger~\cite{ross2011reductionimitationlearningstructured} finetuning, to achieve success. Rigorously evaluating design decisions for BC policies typically requires scaling up real-world evaluation, which is expensive and logistically demanding. 
As a result, progress at the frontier of robot manipulation remains opaque. State-of-the-art systems are often developed in industry labs on proprietary datasets, and their training recipes, when described, rarely provide enough detail to reproduce the results or determine which design choices drive performance.

To address this bottleneck, we present an open, end-to-end stack for BC. At the core of our release is ABC-130K, the largest open teleoperation dataset, comprising 3,500 hours of real-world interaction. It spans over 130{,}000 episodes and 195 distinct tasks, with a wide spectrum of manipulation primitives---including pick-and-place, handovers, tool use, and assembly---as well as highly dexterous behaviors like folding paper planes, folding boxes, and retrieving credit cards from wallets. We also release the accompanying infrastructure required to leverage this corpus, including our hardware setup, training and deployment code, and a simulation pipeline paired with 400 hours of simulation teleoperation data. We demonstrate sim-to-real evaluation correlation, enabling those without physical robots to iterate on their design choices and easily integrate custom task environments.

Using our pipeline, we conduct an exploration of training recipes and architectural design choices for two classes of models: diffusion transformers~\cite{peebles2023scalablediffusionmodelstransformers} and vision-language-action (VLA) models~\cite{brohan2023rt2visionlanguageactionmodelstransfer}. We evaluate core components such as the vision encoder, conditioning schemes, and various pretraining recipes. Our findings are grounded in offline validation metrics (e.g. loss, action error, etc.), simulation performance, and real-world evaluations. The resulting baseline policies successfully execute dexterous tasks out-of-the-box, and we publicly release these model weights alongside our real-world evaluation rollouts. Figure~\ref{fig:abc-stack} summarizes the full ABC stack, including ABC-130K, ABC-Models, ABC-Sim, and ABC-Eval. With the data, models, and recipe we release, our policies are able to perform long-horizon and dexterous tasks such as inserting 6 AirPods in a row, inserting a key into a lockbox to unlock it, folding a cardboard box, and packing a student bag.

Robot learning for complex manipulation is still in its early stages, and optimal training paradigms are far from established. Existing open datasets~\cite{droid,openx} have driven the field forward, but they offer insufficient scale, rely on bespoke and expensive hardware platforms, or lack the necessary training infrastructure and evaluation suites. Our main goal is to accelerate this ongoing exploration by releasing a comprehensive and extensible manipulation stack.

\section{ABC-130K Dataset}

Publicly available robot data for policy learning has grown substantially in recent years. DROID~\cite{droid} and BridgeData-V2~\cite{bridgedatav2} provide high-quality single-arm data on the Franka and WidowX robots, respectively, but on tasks that are largely single-step pick-and-place. Open X-Embodiment~\cite{openx} spans many embodiments and tasks, but its wide variation in quality along with noisy language and action-space metadata makes it hard to use. AgiBot-World~\cite{bu2025agibot_iros} offers 3{,}000 hours of bimanual data, although on a \$30{,}000 robot. Closest to ours is MolmoAct2~\cite{fang2026molmoact2actionreasoningmodels}, which offers 720 hours of data on the YAM platform. Because episodes are stored as encoded video across multiple camera views, with some ranging up to 469 s, naively decoding frames on the fly becomes the dominant throughput cost at this scale. We detail the data-loading challenges this introduces and our optimized dataloader design in Appendix~\ref{app:data_loading}.

\begin{figure*}[h]
    \centering
    \includegraphics[width=\textwidth]{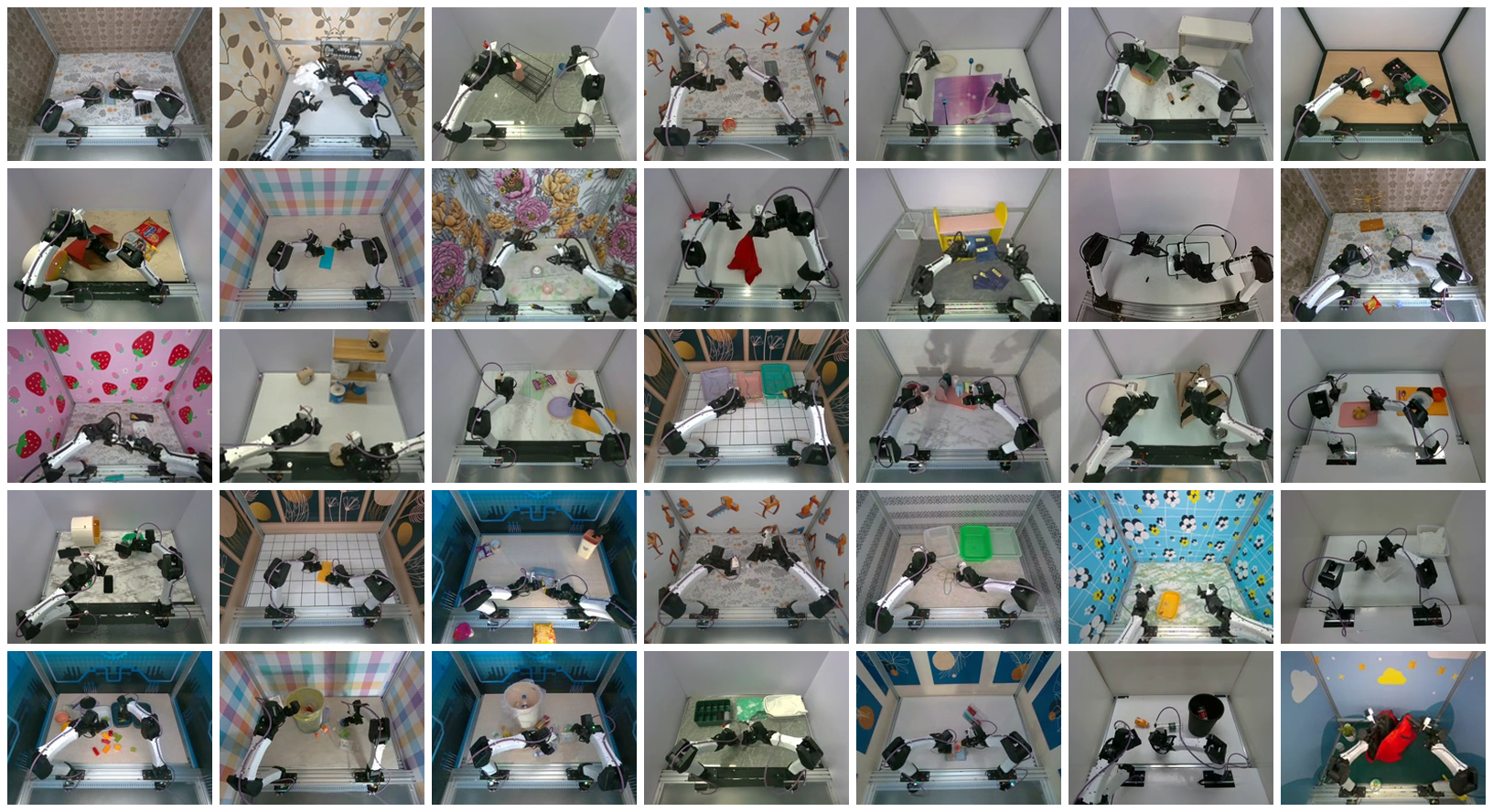}
    \caption{\textbf{Dataset overview.} Random top camera frame samples from ABC-130K. We recommend zooming in to see the diverse tasks, backgrounds, and objects present in the data.}
    \label{fig:task_gallery}
\end{figure*}

With \textbf{ABC-130K}, our goal is to bring the best of these works together:  high-quality data for complex, diverse tasks on a bimanual, inexpensive robot at scale.
Our dataset comprises \textbf{134{,}806 episodes} across \textbf{195 tasks}, totaling \textbf{3{,}553 hours} of bimanual manipulation (see Fig.~\ref{fig:task_gallery}). It spans diverse manipulation primitives such as pick-and-place, folding, handover, insertion, tool use, and assembly. We organize the 195 tasks into 7 primitive categories that capture the dominant contact modes and control strategies. Within each task, we vary the objects and initial configurations. Descriptions of our primitive categories can be found in Appendix~\ref{appendix: data}. Figure~\ref{fig:per_task_hours_grid} provides the distribution of data across tasks within each category. Figure~\ref{fig:task_gallery} shows random samples of the top camera frames from episodes in the data.

We also provide teleoperation metadata, including anonymized teleoperator IDs and collection timestamps. In addition, a 1,552-hour subset of \textbf{ABC-130K} includes subgoal annotations: contiguous sub-trajectories labeled with subtask descriptions. In Appendix~\ref{app:conditioning}, we show how these metadata and annotations can be used for policy conditioning.

\begin{figure*}[h]
    \centering
    \makebox[\textwidth][c]{\definecolor{CatPnP}{rgb}{0.121569,0.466667,0.705882}%
\definecolor{CatFinePnP}{rgb}{0.839216,0.152941,0.156863}%
\definecolor{CatFolding}{HTML}{2CA02C}%
\definecolor{CatInsertion}{HTML}{FF7F0E}%
\definecolor{CatToolUse}{HTML}{9467BD}%
\definecolor{CatSorting}{HTML}{17BECF}%
\definecolor{CatTying}{HTML}{8C564B}%
\begin{tikzpicture}

\begin{axis}[
  name=topL,
  at={(0,0)}, anchor=north west, xshift=8mm,
  width=70mm, height=26mm,
  ymode=log, ymin=0.1, ymax=400,
  log origin=infty,
  ytick={0.1, 1, 10, 100},
  log basis y=10,
  xmin=0.3, xmax=67.7,
  xtick={1, 67},
  xticklabels={1, 67},
  x tick label style={font=\scriptsize, color=PrettyText, yshift=1pt},
  y tick label style={font=\scriptsize, color=PrettyText},
  axis line style={color=PrettyText!60},
  tick style={color=PrettyText!60},
  major grid style={gray!20},
  ymajorgrids,
  ylabel={Hours per task (log)},
  ylabel near ticks,
  ylabel shift={-8pt},
  label style={font=\small, color=PrettyText},
  title={Pick-and-Place},
  title style={font=\small, color=PrettyText, yshift=-1.5mm,
                inner sep=0pt},
  scale only axis,
  enlarge x limits=false,
]
\addplot+[
  ybar, fill=CatPnP, draw=none, mark=none, bar width=1.4835820895522387pt,
] table [x=idx, y=hours, col sep=tab] {figures/pgfplots/data/per_task_hours_Pick_and_Place.tsv};
\end{axis}

\begin{axis}[
  name=topR,
  at={(topL.outer east)}, anchor=outer west, xshift=6mm,
  width=70mm, height=26mm,
  ymode=log, ymin=0.1, ymax=400,
  log origin=infty,
  ytick={0.1, 1, 10, 100},
  log basis y=10,
  xmin=0.3, xmax=39.7,
  xtick={1, 39},
  xticklabels={1, 39},
  x tick label style={font=\scriptsize, color=PrettyText, yshift=1pt},
  y tick label style={font=\scriptsize, color=PrettyText},
  axis line style={color=PrettyText!60},
  tick style={color=PrettyText!60},
  major grid style={gray!20},
  ymajorgrids,
  yticklabels={},
  ylabel near ticks,
  ylabel shift={-2pt},
  label style={font=\small, color=PrettyText},
  title={Fine Pick-and-Place},
  title style={font=\small, color=PrettyText, yshift=-1.5mm,
                inner sep=0pt},
  scale only axis,
  enlarge x limits=false,
]
\addplot+[
  ybar, fill=CatFinePnP, draw=none, mark=none, bar width=2.5487179487179485pt,
] table [x=idx, y=hours, col sep=tab] {figures/pgfplots/data/per_task_hours_Fine_Pick_and_Place.tsv};
\end{axis}

\begin{axis}[
  name=bot0,
  at={(topL.outer south west)}, anchor=outer north west, xshift=-8mm, yshift=-7mm,
  width=54.76056338028169mm, height=26mm,
  ymode=log, ymin=0.1, ymax=400,
  log origin=infty,
  ytick={0.1, 1, 10, 100},
  log basis y=10,
  xmin=0.3, xmax=36.7,
  xtick={1, 36},
  xticklabels={1, 36},
  x tick label style={font=\scriptsize, color=PrettyText, yshift=1pt},
  y tick label style={font=\scriptsize, color=PrettyText},
  axis line style={color=PrettyText!60},
  tick style={color=PrettyText!60},
  major grid style={gray!20},
  ymajorgrids,
  ylabel={Hours per task (log)},
  ylabel near ticks,
  ylabel shift={-8pt},
  label style={font=\small, color=PrettyText},
  title={Folding},
  title style={font=\small, color=PrettyText, yshift=-1.5mm,
                inner sep=0pt},
  scale only axis,
  enlarge x limits=false,
]
\addplot+[
  ybar, fill=CatFolding, draw=none, mark=none, bar width=2.1599999999999997pt,
] table [x=idx, y=hours, col sep=tab] {figures/pgfplots/data/per_task_hours_Folding.tsv};
\end{axis}

\begin{axis}[
  name=bot1,
  at={(bot0.outer east)}, anchor=outer west, xshift=1.5mm,
  width=28.901408450704224mm, height=26mm,
  ymode=log, ymin=0.1, ymax=400,
  log origin=infty,
  ytick={0.1, 1, 10, 100},
  log basis y=10,
  xmin=0.3, xmax=19.7,
  xtick={1, 19},
  xticklabels={1, 19},
  x tick label style={font=\scriptsize, color=PrettyText, yshift=1pt},
  y tick label style={font=\scriptsize, color=PrettyText},
  axis line style={color=PrettyText!60},
  tick style={color=PrettyText!60},
  major grid style={gray!20},
  ymajorgrids,
  yticklabels={},
  ylabel near ticks,
  ylabel shift={-2pt},
  label style={font=\small, color=PrettyText},
  title={Insertion/Ejection},
  title style={font=\small, color=PrettyText, yshift=-1.5mm,
                inner sep=0pt},
  scale only axis,
  enlarge x limits=false,
]
\addplot+[
  ybar, fill=CatInsertion, draw=none, mark=none, bar width=2.16pt,
] table [x=idx, y=hours, col sep=tab] {figures/pgfplots/data/per_task_hours_Insertion_Ejection.tsv};
\end{axis}

\begin{axis}[
  name=bot2,
  at={(bot1.outer east)}, anchor=outer west, xshift=1.5mm,
  width=24.338028169014084mm, height=26mm,
  ymode=log, ymin=0.1, ymax=400,
  log origin=infty,
  ytick={0.1, 1, 10, 100},
  log basis y=10,
  xmin=0.3, xmax=16.7,
  xtick={1, 16},
  xticklabels={1, 16},
  x tick label style={font=\scriptsize, color=PrettyText, yshift=1pt},
  y tick label style={font=\scriptsize, color=PrettyText},
  axis line style={color=PrettyText!60},
  tick style={color=PrettyText!60},
  major grid style={gray!20},
  ymajorgrids,
  yticklabels={},
  ylabel near ticks,
  ylabel shift={-2pt},
  label style={font=\small, color=PrettyText},
  title={Tool Use},
  title style={font=\small, color=PrettyText, yshift=-1.5mm,
                inner sep=0pt},
  scale only axis,
  enlarge x limits=false,
]
\addplot+[
  ybar, fill=CatToolUse, draw=none, mark=none, bar width=2.1599999999999997pt,
] table [x=idx, y=hours, col sep=tab] {figures/pgfplots/data/per_task_hours_Tool_Use.tsv};
\end{axis}

\begin{axis}[
  name=bot3,
  at={(bot2.outer east)}, anchor=outer west, xshift=1.5mm,
  width=16.0mm, height=26mm,
  ymode=log, ymin=0.1, ymax=400,
  log origin=infty,
  ytick={0.1, 1, 10, 100},
  log basis y=10,
  xmin=0.3, xmax=8.7,
  xtick={1, 8},
  xticklabels={1, 8},
  x tick label style={font=\scriptsize, color=PrettyText, yshift=1pt},
  y tick label style={font=\scriptsize, color=PrettyText},
  axis line style={color=PrettyText!60},
  tick style={color=PrettyText!60},
  major grid style={gray!20},
  ymajorgrids,
  yticklabels={},
  ylabel near ticks,
  ylabel shift={-2pt},
  label style={font=\small, color=PrettyText},
  title={Sorting},
  title style={font=\footnotesize, color=PrettyText, yshift=-1.5mm,
                inner sep=0pt},
  scale only axis,
  enlarge x limits=false,
]
\addplot+[
  ybar, fill=CatSorting, draw=none, mark=none, bar width=2.84pt,
] table [x=idx, y=hours, col sep=tab] {figures/pgfplots/data/per_task_hours_Sorting.tsv};
\end{axis}

\begin{axis}[
  name=bot4,
  at={(bot3.outer east)}, anchor=outer west, xshift=1.5mm,
  width=16.0mm, height=26mm,
  ymode=log, ymin=0.1, ymax=400,
  log origin=infty,
  ytick={0.1, 1, 10, 100},
  log basis y=10,
  xmin=0.3, xmax=10.7,
  xtick={1, 10},
  xticklabels={1, 10},
  x tick label style={font=\scriptsize, color=PrettyText, yshift=1pt},
  y tick label style={font=\scriptsize, color=PrettyText},
  axis line style={color=PrettyText!60},
  tick style={color=PrettyText!60},
  major grid style={gray!20},
  ymajorgrids,
  yticklabels={},
  ylabel near ticks,
  ylabel shift={-2pt},
  label style={font=\small, color=PrettyText},
  title={Tying/Untying},
  title style={font=\footnotesize, color=PrettyText, yshift=-1.5mm,
                inner sep=0pt},
  scale only axis,
  enlarge x limits=false,
]
\addplot+[
  ybar, fill=CatTying, draw=none, mark=none, bar width=2.272pt,
] table [x=idx, y=hours, col sep=tab] {figures/pgfplots/data/per_task_hours_Untying_Tying.tsv};
\end{axis}
\end{tikzpicture}}
    \caption{\textbf{Per-task hours by primitive category.} Hours-per-task on a log scale in descending order within each category. Top row: \textit{Pick-and-Place} and \textit{Fine Pick-and-Place} (the two largest categories in terms of number of tasks). Bottom row: the five remaining categories. Descriptions of each task category along with example images from the data can be found in Appendix~\ref{appendix: data}.}
    \label{fig:per_task_hours_grid}
\end{figure*}
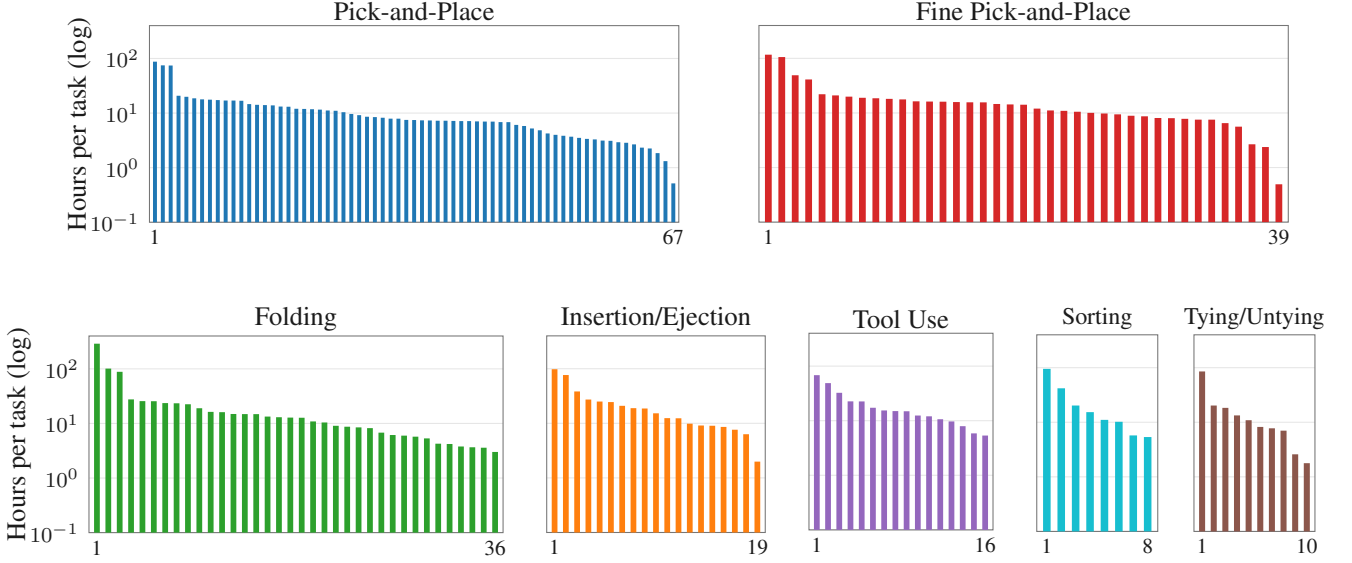

\section{ABC-Models}
\label{sec:abc_models}

\begin{table}[t]
    \centering
    \small
    \setlength{\tabcolsep}{3.5pt}
    \renewcommand{\arraystretch}{1.08}
    \resizebox{\columnwidth}{!}{%
    \begin{tabular}{@{}llccc|ccc@{}}
        \toprule
        & & \multicolumn{3}{c|}{\textbf{VLA variants}} 
          & \multicolumn{3}{c}{\textbf{DiT variants}} \\
        \cmidrule(lr){3-5} \cmidrule(lr){6-8}
        Task & Metric 
        & Pooled adaLN & FAST + X-Attn & X-Attn
        & DINOv3-xattn & CLIP-adaln & CLIP-xattn \\
        \midrule

        Bottles 
        & Strict 
        & $\mathbf{67.3}\%$ 
        & $8.0\%$ 
        & $0.0\%$ 
        & $\mathbf{73.5}\%$ 
        & $11.8\%$ 
        & $38.9\%$ \\
        & Progress 
        & $\mathbf{83.0}\%$ 
        & $44.0\%$ 
        & $5.2\%$
        & $\mathbf{93.1}\%$ 
        & $53.6\%$ 
        & $74.1\%$ \\

        \addlinespace[0.15em]

        Dishrack 
        & Strict 
        & $\mathbf{30.0}\%$ 
        & $2.8\%$ 
        & $0.0\%$
        & $23.2\%$ 
        & $28.6\%$ 
        & $\mathbf{34.7}\%$ \\
        & Progress 
        & $\mathbf{75.7}\%$ 
        & $47.2\%$ 
        & $25.0\%$
        & $74.1\%$ 
        & $72.3\%$ 
        & $\mathbf{81.3}\%$ \\

        \addlinespace[0.15em]

        Mugs 
        & Strict 
        & $\mathbf{1.0}\%$ 
        & $0.0\%$ 
        & $0.0\%$
        & $\mathbf{2.0}\%$ 
        & $0.0\%$ 
        & $0.0\%$ \\
        & Progress 
        & $\mathbf{25.5}\%$ 
        & $6.5\%$ 
        & $5.0\%$
        & $\mathbf{35.3}\%$ 
        & $15.9\%$ 
        & $21.0\%$ \\

        \midrule

        Aggregate
        & Mean strict 
        & $\mathbf{32.8}\%$ 
        & $3.6\%$ 
        & $0.0\%$
        & $\mathbf{32.9}\%$ 
        & $13.4\%$ 
        & $24.5\%$ \\
        & Mean progress 
        & $\mathbf{61.4}\%$ 
        & $32.6\%$ 
        & $11.7\%$
        & $\mathbf{67.5}\%$ 
        & $47.3\%$ 
        & $58.8\%$ \\
        & Latency (ms) 
        & $17.24$ 
        & $19.24$ 
        & $19.24$
        & $37.4$ 
        & $27.5$
        & $37.5$ \\

        \bottomrule
    \end{tabular}%
    }
    \vspace{0.35em}
    \caption{\textbf{Architecture ablations.}
    Real-world evaluation of VLA and DiT architecture variants after 200k training steps. 
    Strict success reports the percentage of tasks completed; progress reports average task progress. Latency is measured in ms/chunk for 10 diffusion steps. The models are trained on an internal corpus of 7{,}000 hours of robot data.}
    \label{tab:architecture_ablations}
\end{table}

\label{sec:abc-models}

We study two policy families for large-scale behavior cloning: \textbf{ABC-DiT}, a diffusion-transformer policy with lightweight vision and language encoders, and \textbf{ABC-VLA}, a VLM-backed policy with a diffusion action head. These two models reflect common design choices in recent robot learning systems: diffusion policies built on pretrained visual representations~\cite{chi2025diffusion, barreiros2025careful}, and vision-language-action models that adapt pretrained VLM backbones for robot control~\cite{groot_n1, pi0, pi05, brohan2022rt}. Below, we study how base encoders, interfaces between backbones and action heads, and training hyperparameters affect performance across these two families. Unless otherwise noted, the comparison models in this section are trained on an internal corpus of 7{,}000 hours, which we had for development before the release dataset was finalized. We take the modeling decisions from this study to train and evaluate our final ABC models on the publicly released \textbf{ABC-130K} in Section~\ref{sec:capabilities}. All models take as input three $224 \times 224$ images from the top camera and two gripper cameras, along with a 14D proprioceptive joint state. Non-square images are padded using letterbox padding. The policy outputs a 30-step chunk of absolute joint-position targets, and states and actions are normalized with z-score normalization. The architecture details for both models can be found in Appendix~\ref{app:arch-details} and the training details in Appendix~\ref{app:training_details}. All models are trained on NVIDIA H200 GPUs.

All real-world evaluations in this section are on three tasks: \textit{throwing plastic bottles into a bin}, \textit{loading plates into a tabletop dish rack}, and \textit{turning mugs right side up}. Each task is evaluated for 50 trials. Our policies are executed at 30 Hz. Following \cite{black2025trainingtimeactionconditioningefficient}, we execute our policies asynchronously with an action prefix. This is done to prevent mode switching, thereby ensuring smooth trajectories. As noted in Appendix~\ref{sec:action-prefix-conditioning}, depending on the task, performance can improve by modulating the length of the prefix as well as the execution horizon of the action chunk.

\subsection{ABC-DiT}

We study various architectural choices for diffusion transformers~\cite{peebles2023scalablediffusionmodelstransformers}. We specifically investigate how the DiT head communicates with various vision encoders. All our DiTs use 32 layers, 24 attention heads, and a hidden dimension of 1536, which is larger than the DiT-XL architecture. We chose this large model size to ensure that architecture choices were not bottlenecked by network capacity, allowing us to better understand what matters at scale (comparisons of various DiT sizes are in Appendix~\ref{app:arch_scaling}). We compare the following three variants below.

\textit{CLIP-AdaLN}: We begin with a baseline inspired by the LBM architecture~\cite{barreiros2025careful}. The policy backbone is a DiT~\cite{peebles2023scalablediffusionmodelstransformers} that takes as input a noised action chunk and predicts the velocity of the conditional probability flow. Each image is encoded independently using a CLIP ViT-B image encoder~\cite{radford2021learningtransferablevisualmodels}, and the resulting CLS tokens are concatenated to form a compact visual representation. The task language instruction is embedded using the CLIP text encoder and projected with an MLP. This vector is concatenated with the visual representation, an MLP projection of the robot proprioception, and the diffusion timestep embedding to form a single conditioning vector. The DiT is then conditioned on this vector through a standard adaLN conditioning mechanism~\cite{peebles2023scalablediffusionmodelstransformers}. 

\textit{CLIP-Cross-Attention}: In the previous variant, the AdaLN conditioning compresses each image into a single token. This compression might lose fine-grained spatial information that is important for manipulation. Alternatively, we can have the DiT attend to the sequence of visual tokens. Since directly attending to all tokens is expensive, we use a set of learned latent queries to attend over all vision tokens. For all our experiments, we use 12 latent query tokens per image as it offers a good balance of expressivity and memory overhead. In the DiT block, after each self-attention layer, we add a cross-attention layer where the query tokens are taken from the noised action tokens and the keys and values are the pooled visual tokens. The non-visual conditioning variables remain in the AdaLN pathway. 

\textit{DINO-Cross-Attention}: Image-text alignment is not necessarily the objective most aligned with low-level robot actions. Manipulation policies often require precise spatial cues that may not be optimally preserved by CLIP but may be captured with more pixel-aligned representations like DINOv3~\cite{simeoni2025dinov3}. We use the cross-attention architecture and replace the CLIP image encoder with a DINOv3 ViT-B encoder.

All models are trained for 200k steps with a global batch size of 4608. We linearly warm up the learning rate for the first 1000 iterations to $1 \times 10^{-4}$ and then keep it constant throughout training. We use the AdamW optimizer~\cite{loshchilov2019decoupledweightdecayregularization} with a weight decay of 0.01. We use a learning rate scale of 0.1 for the vision encoder. We use gradient clipping with a maximum gradient norm of 10. The proprioception input is dropped out with a probability of 0.1.

The results are shown in Table~\ref{tab:architecture_ablations}. Replacing global AdaLN-based visual conditioning with pooled cross-attention improves task progress, suggesting that preserving token-level visual information is important for action generation. Replacing CLIP with DINOv3 further improves performance, indicating that the choice of visual representation is also a key design factor for real-world manipulation. For the rest of this paper, we use DINOv3 and this pooled cross-attention conditioning scheme as our \textit{ABC-DiT} model. This model has 2B parameters in total.

\subsection{ABC-VLA}

For \textit{ABC-VLA}, we build on Gemma 3~\cite{kamath2025gemma}, a 4B-parameter VLM with a SigLIP vision encoder and a Gemma language decoder, which we pair with a small  DiT action head to generate action chunks. We consistently find that freezing the VLM backbone performs poorly, so we finetune the backbone using the action prediction objective. The connector between the VLM and action head routes vision-language features into the action expert and can provide gradients back to update the VLM. Existing VLAs adopt many different connector mechanisms, but their effects are typically entangled with differences in data and training recipe, so their relative merits are difficult to isolate. We therefore compare three representative connectors directly on our data, fixing the backbone, action head, and training recipe so that only the connector varies.

\textit{Cross Attention}: We run the VLM on the image, language, and proprioceptive context, extract final-layer token features, and project them to the DiT token dimension. These projected VLM features serve as the keys and values for a cross-attention layer in each DiT block, while the noisy action token stream inside the DiT queries them. The VLM backbone is finetuned end-to-end by the diffusion objective, so diffusion gradients flow through the DiT's cross-attention back into the VLM.

\textit{Cross Attention + FAST}: Following~\cite{driess2025knowledgeinsulatingvisionlanguageactionmodels}, we concatenate FAST-tokenized~\cite{pertsch2025fastefficientactiontokenization} actions in the VLM token sequence and train with next-token-prediction CE to finetune the VLM. Simultaneously, the DiT cross-attends to the context token features at the final VLM layer, which we detach to ``insulate'' the backbone from diffusion gradients. In our experiments, cross attention slightly outperforms $\pi_{0.5}$-style~\cite{pi05} reuse of the VLM's keys and values.

\textit{Adaptive LayerNorm (AdaLN)}: We adapt the LBM-2~\cite{barreiros2025careful} conditioning scheme, where VLM features are compressed into a fixed-length conditioning vector that modulates the DiT through adaLN~\cite{peebles2023scalablediffusionmodelstransformers}. Unlike the cross-attention variants, the DiT action tokens do not directly attend to VLM tokens. Instead, VLM token features from the final layer are summarized by an attention-pooling module into 8 512-dimensional tokens, which are flattened and projected to produce the DiT's adaLN shifts, scales, and residual gates. We use attention pooling with QK-Norm~\cite{henry2020querykeynormalizationtransformers} in place of LBM-2's register tokens.

\noindent \textbf{Variance Reduction.} For \textit{AdaLN} and \textit{Cross Attention}, the diffusion training loss is an expectation over three random variables: the data sample $x$, the noise $\epsilon$, and the timestep $\tau$ (i.e., $\nabla \mathcal{L} = \mathbb{E}_{x, \epsilon, \tau}[\nabla \ell(x, \epsilon, \tau)]$). Since our DiT is small relative to the VLM (42M vs. 4B parameters), it is efficient to amortize a single VLM pass across many ($\epsilon$, $\tau$) draws. Specifically, in the forward pass, we replicate the VLM features and clean action $k$ times within the batch and pair each copy with an independent ($\epsilon$, $\tau$) sample. In the backward pass, the gradient expansion averages at the conditioning interface, so the VLM backward cost is independent of $k$. This yields a strictly lower-variance gradient estimator with negligible overhead: 1.346 s/step at $k=1$ and 1.366 s/step at $k=8$ on H200. Figure~\ref{fig:vla-diffusion-draws-train-loss} shows the improvement that the multiple draws trick provides.

Table~\ref{tab:architecture_ablations} compares the three variants. Vanilla cross-attention performs worst on aggregate progress (11.7\%), consistent with prior studies. FAST co-training helps (32.6\%) by replacing the diffusion gradient on the VLM with a cross-entropy gradient. AdaLN beats both variants on our data (61.4\%) despite exposing the VLM to denoising gradients. This suggests that diffusion gradients are not intrinsically incompatible with VLM features if given a proper interface.

\subsection{Scaling compute across batch size and training steps}

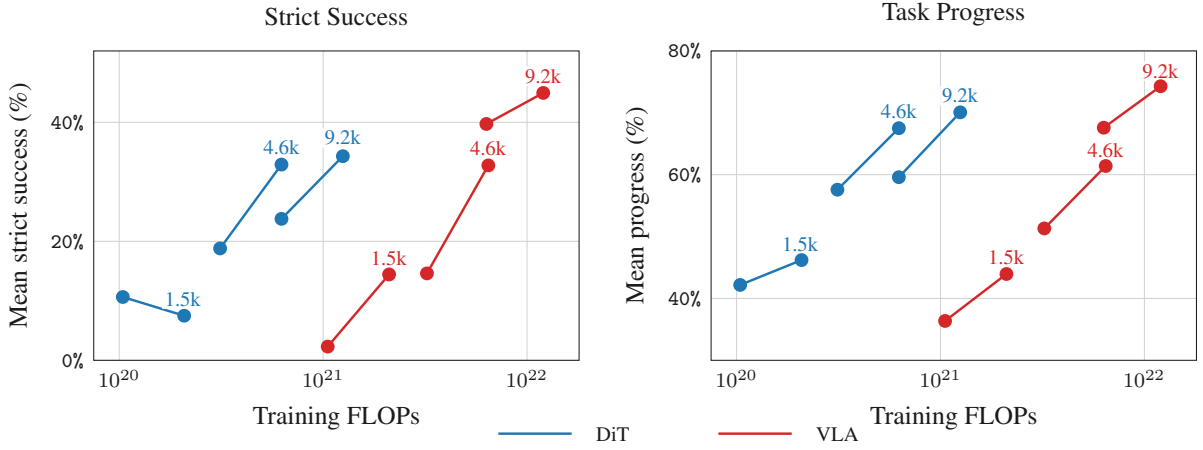
\begin{figure*}[t]
    \centering
    \resizebox{\textwidth}{!}{%
        \begin{tikzpicture}
\begin{axis}[
  name=computeSuccess,
  width=82mm, height=58mm,
  xmode=log,
  xmin=7.5e19, xmax=1.8e22,
  ymin=0, ymax=52,
  xtick={1e20,1e21,1e22},
  xticklabels={$10^{20}$,$10^{21}$,$10^{22}$},
  ytick={0,20,40},
  yticklabels={0\%,20\%,40\%},
  grid=major,
  xlabel={Training FLOPs},
  ylabel={Mean strict success (\%)},
  title={Strict Success},
  axis line style={PrettyText},
  tick label style={font=\ttfamily\scriptsize, PrettyText},
  label style={font=\small, PrettyText},
  title style={font=\small, PrettyText},
]
\addplot[color=LBMColor, no marks, solid, line width=0.95pt]
  table [x expr=\thisrow{flops_ef}*1e18, y=agg_success] {figures/pgfplots/data/flops_progress_3tasks_lbm_1536.tsv};
\addplot[color=VLAColor, no marks, solid, line width=0.95pt]
  table [x expr=\thisrow{flops_ef}*1e18, y=agg_success] {figures/pgfplots/data/flops_progress_3tasks_vla_1500.tsv};
\addplot[color=LBMColor, no marks, solid, line width=0.95pt]
  table [x expr=\thisrow{flops_ef}*1e18, y=agg_success] {figures/pgfplots/data/flops_progress_3tasks_lbm_4608.tsv};
\addplot[color=VLAColor, no marks, solid, line width=0.95pt]
  table [x expr=\thisrow{flops_ef}*1e18, y=agg_success] {figures/pgfplots/data/flops_progress_3tasks_vla_4600.tsv};
\addplot[color=LBMColor, no marks, solid, line width=0.95pt]
  table [x expr=\thisrow{flops_ef}*1e18, y=agg_success] {figures/pgfplots/data/flops_progress_3tasks_lbm_9216.tsv};
\addplot[color=VLAColor, no marks, solid, line width=0.95pt]
  table [x expr=\thisrow{flops_ef}*1e18, y=agg_success] {figures/pgfplots/data/flops_progress_3tasks_vla_9000.tsv};

\addplot[color=LBMColor, only marks, mark=*, mark size=2.3pt, solid,
  mark options={draw=none, fill=LBMColor}]
  table [x expr=\thisrow{flops_ef}*1e18, y=agg_success] {figures/pgfplots/data/flops_progress_3tasks_lbm_1536.tsv};
\addplot[color=VLAColor, only marks, mark=*, mark size=2.3pt, solid,
  mark options={draw=none, fill=VLAColor}]
  table [x expr=\thisrow{flops_ef}*1e18, y=agg_success] {figures/pgfplots/data/flops_progress_3tasks_vla_1500.tsv};
\addplot[color=LBMColor, only marks, mark=*, mark size=2.3pt, solid,
  mark options={draw=none, fill=LBMColor}]
  table [x expr=\thisrow{flops_ef}*1e18, y=agg_success] {figures/pgfplots/data/flops_progress_3tasks_lbm_4608.tsv};
\addplot[color=VLAColor, only marks, mark=*, mark size=2.3pt, solid,
  mark options={draw=none, fill=VLAColor}]
  table [x expr=\thisrow{flops_ef}*1e18, y=agg_success] {figures/pgfplots/data/flops_progress_3tasks_vla_4600.tsv};
\addplot[color=LBMColor, only marks, mark=*, mark size=2.3pt, solid,
  mark options={draw=none, fill=LBMColor}]
  table [x expr=\thisrow{flops_ef}*1e18, y=agg_success] {figures/pgfplots/data/flops_progress_3tasks_lbm_9216.tsv};
\addplot[color=VLAColor, only marks, mark=*, mark size=2.3pt, solid,
  mark options={draw=none, fill=VLAColor}]
  table [x expr=\thisrow{flops_ef}*1e18, y=agg_success] {figures/pgfplots/data/flops_progress_3tasks_vla_9000.tsv};

  \node[anchor=south, font=\scriptsize, text=LBMColor, fill=white,
    fill opacity=0.82, text opacity=1, inner sep=1pt]
    at (axis cs:2.05e20,8.6) {1.5k};
  \node[anchor=south, font=\scriptsize, text=LBMColor, fill=white,
    fill opacity=0.82, text opacity=1, inner sep=1pt]
    at (axis cs:6.25e20,34.2) {4.6k};
  \node[anchor=south, font=\scriptsize, text=LBMColor, fill=white,
    fill opacity=0.82, text opacity=1, inner sep=1pt]
    at (axis cs:1.25e21,35.7) {9.2k};
  \node[anchor=south, font=\scriptsize, text=VLAColor, fill=white,
    fill opacity=0.82, text opacity=1, inner sep=1pt]
    at (axis cs:2.10e21,15.4) {1.5k};
  \node[anchor=south, font=\scriptsize, text=VLAColor, fill=white,
    fill opacity=0.82, text opacity=1, inner sep=1pt]
    at (axis cs:6.45e21,34.0) {4.6k};
  \node[anchor=south, font=\scriptsize, text=VLAColor, fill=white,
    fill opacity=0.82, text opacity=1, inner sep=1pt]
    at (axis cs:1.20e22,46.1) {9.2k};
\end{axis}

\begin{axis}[
  name=computeProgress,
  at={($(computeSuccess.south east)+(18mm,0)$)},
  anchor=south west,
  width=82mm, height=58mm,
  xmode=log,
  xmin=7.5e19, xmax=1.8e22,
  ymin=30, ymax=80,
  xtick={1e20,1e21,1e22},
  xticklabels={$10^{20}$,$10^{21}$,$10^{22}$},
  ytick={40,60,80},
  yticklabels={40\%,60\%,80\%},
  grid=major,
  xlabel={Training FLOPs},
  ylabel={Mean progress (\%)},
  title={Task Progress},
  axis line style={PrettyText},
  tick label style={font=\ttfamily\scriptsize, PrettyText},
  label style={font=\small, PrettyText},
  title style={font=\small, PrettyText},
]
\addplot[color=LBMColor, no marks, solid, line width=0.95pt]
  table [x expr=\thisrow{flops_ef}*1e18, y=agg_progress] {figures/pgfplots/data/flops_progress_3tasks_lbm_1536.tsv};
\addplot[color=VLAColor, no marks, solid, line width=0.95pt]
  table [x expr=\thisrow{flops_ef}*1e18, y=agg_progress] {figures/pgfplots/data/flops_progress_3tasks_vla_1500.tsv};
\addplot[color=LBMColor, no marks, solid, line width=0.95pt]
  table [x expr=\thisrow{flops_ef}*1e18, y=agg_progress] {figures/pgfplots/data/flops_progress_3tasks_lbm_4608.tsv};
\addplot[color=VLAColor, no marks, solid, line width=0.95pt]
  table [x expr=\thisrow{flops_ef}*1e18, y=agg_progress] {figures/pgfplots/data/flops_progress_3tasks_vla_4600.tsv};
\addplot[color=LBMColor, no marks, solid, line width=0.95pt]
  table [x expr=\thisrow{flops_ef}*1e18, y=agg_progress] {figures/pgfplots/data/flops_progress_3tasks_lbm_9216.tsv};
\addplot[color=VLAColor, no marks, solid, line width=0.95pt]
  table [x expr=\thisrow{flops_ef}*1e18, y=agg_progress] {figures/pgfplots/data/flops_progress_3tasks_vla_9000.tsv};

\addplot[color=LBMColor, only marks, mark=*, mark size=2.3pt, solid,
  mark options={draw=none, fill=LBMColor}]
  table [x expr=\thisrow{flops_ef}*1e18, y=agg_progress] {figures/pgfplots/data/flops_progress_3tasks_lbm_1536.tsv};
\addplot[color=VLAColor, only marks, mark=*, mark size=2.3pt, solid,
  mark options={draw=none, fill=VLAColor}]
  table [x expr=\thisrow{flops_ef}*1e18, y=agg_progress] {figures/pgfplots/data/flops_progress_3tasks_vla_1500.tsv};
\addplot[color=LBMColor, only marks, mark=*, mark size=2.3pt, solid,
  mark options={draw=none, fill=LBMColor}]
  table [x expr=\thisrow{flops_ef}*1e18, y=agg_progress] {figures/pgfplots/data/flops_progress_3tasks_lbm_4608.tsv};
\addplot[color=VLAColor, only marks, mark=*, mark size=2.3pt, solid,
  mark options={draw=none, fill=VLAColor}]
  table [x expr=\thisrow{flops_ef}*1e18, y=agg_progress] {figures/pgfplots/data/flops_progress_3tasks_vla_4600.tsv};
\addplot[color=LBMColor, only marks, mark=*, mark size=2.3pt, solid,
  mark options={draw=none, fill=LBMColor}]
  table [x expr=\thisrow{flops_ef}*1e18, y=agg_progress] {figures/pgfplots/data/flops_progress_3tasks_lbm_9216.tsv};
\addplot[color=VLAColor, only marks, mark=*, mark size=2.3pt, solid,
  mark options={draw=none, fill=VLAColor}]
  table [x expr=\thisrow{flops_ef}*1e18, y=agg_progress] {figures/pgfplots/data/flops_progress_3tasks_vla_9000.tsv};

  \node[anchor=south, font=\scriptsize, text=LBMColor, fill=white,
    fill opacity=0.82, text opacity=1, inner sep=1pt]
    at (axis cs:2.05e20,47.2) {1.5k};
  \node[anchor=south, font=\scriptsize, text=LBMColor, fill=white,
    fill opacity=0.82, text opacity=1, inner sep=1pt]
    at (axis cs:6.25e20,68.6) {4.6k};
  \node[anchor=south, font=\scriptsize, text=LBMColor, fill=white,
    fill opacity=0.82, text opacity=1, inner sep=1pt]
    at (axis cs:1.25e21,71.1) {9.2k};
  \node[anchor=south, font=\scriptsize, text=VLAColor, fill=white,
    fill opacity=0.82, text opacity=1, inner sep=1pt]
    at (axis cs:2.10e21,45.0) {1.5k};
  \node[anchor=south, font=\scriptsize, text=VLAColor, fill=white,
    fill opacity=0.82, text opacity=1, inner sep=1pt]
    at (axis cs:6.45e21,62.4) {4.6k};
  \node[anchor=south, font=\scriptsize, text=VLAColor, fill=white,
    fill opacity=0.82, text opacity=1, inner sep=1pt]
    at (axis cs:1.20e22,75.1) {9.2k};
\end{axis}

\coordinate (legendbase) at ($(computeSuccess.south)!0.5!(computeProgress.south)+(0,-10mm)$);
\begin{scope}[font=\scriptsize, text=PrettyText]
  \draw[LBMColor, line width=0.95pt] ($(legendbase)+(-20mm,0mm)$) -- ($(legendbase)+(-10mm,0mm)$);
  \node[anchor=west] at ($(legendbase)+(-8mm,0mm)$) {DiT};
  \draw[VLAColor, line width=0.95pt] ($(legendbase)+(10mm,0mm)$) -- ($(legendbase)+(20mm,0mm)$);
  \node[anchor=west] at ($(legendbase)+(22mm,0mm)$) {VLA};
\end{scope}
\end{tikzpicture}%
    }

    \caption{\textbf{More training steps and larger batch size give better performance.}
    Real-world task success (left) and task progress (right) with training compute for DiT and VLA policies; each connected pair uses the same effective batch size (approximately 1.5K, 4.6K, or 9K) evaluated at different checkpoints. We consistently find that the DiT is more flop-efficient.}
    \label{fig:abc_lite_large}
\end{figure*}

When training policies, there is no consensus on the optimal batch size. Models like OpenVLA~\cite{kim2024openvlaopensourcevisionlanguageactionmodel} and LBM~\cite{barreiros2025careful} use 2048 and 2560, respectively, whereas GR00T N1~\cite{groot_n1} uses 16{,}384. In order to more precisely understand the effect of increasing the batch size, we train three different variants of \textit{ABC-DiT} and \textit{ABC-VLA}. We train both on batch sizes of 1536, 4608, and 9216.
Figure~\ref{fig:abc_lite_large} shows the plot of real-world task success and progress for these variants for 100K and 200K training steps. For the relatively smaller batch sizes, the DiT outperforms the VLA. However, as we scale the batch size to 9216, the DiT does not improve as much, whereas we see the VLA has a much larger jump in performance. Furthermore, we see that the VLA, at higher batch sizes, is able to outperform the DiT in real-world performance, though the DiT is more compute-efficient.

  \begin{figure*}[t]
      \centering
      \begin{minipage}[t]{0.49\linewidth}
          \centering
          \resizebox{\linewidth}{!}{%
              \begin{tikzpicture}
\begin{axis}[
  width=122mm, height=67mm,
  xmin=0, xmax=3000,
  ymin=0.045, ymax=0.16,
  ymode=log,
  ytick={0.05,0.06,0.08,0.1,0.14},
  yticklabels={0.05,0.06,0.08,0.10,0.14},
  xtick={0,500,1000,1500,2000,2500,3000},
  scaled x ticks=false,
  grid=major,
  xlabel={H200 GPU-hours},
  ylabel={Train loss},
  axis line style={PrettyText},
  tick label style={font=\ttfamily\scriptsize, PrettyText},
  label style={font=\small, PrettyText},
  legend style={
    draw=none,
    fill=white,
    fill opacity=0.92,
    text opacity=1,
    at={(0.97,0.97)},
    anchor=north east,
    font=\scriptsize,
  },
]
\addplot+[color=VLAColor, no marks, line width=0.55pt, opacity=0.78]
  table [x=h200_gpu_hours, y=loss]
  {figures/pgfplots/data/vla_diffusion_draws_train_loss_h200_raw_8draw.tsv};
\addlegendentry{8 Draws}
\addplot+[color=LBMColor, no marks, line width=0.55pt, opacity=0.78]
  table [x=h200_gpu_hours, y=loss]
  {figures/pgfplots/data/vla_diffusion_draws_train_loss_h200_raw_1draw.tsv};
\addlegendentry{1 Draw}
\end{axis}
\end{tikzpicture}%
          }
          \captionof{figure}{\textbf{Eight diffusion draws reduce VLA train loss at fixed accelerator time.}
          Training loss versus GPU-hours for one- and eight-draw VLA diffusion training.}
          \label{fig:vla-diffusion-draws-train-loss}
      \end{minipage}\hfill
      \begin{minipage}[t]{0.49\linewidth}
          \centering
          \resizebox{\linewidth}{!}{%
              \begin{tikzpicture}
\begin{axis}[
  name=lossVlaOnly,
  width=122mm, height=67mm,
  xmin=0, xmax=220000,
  ymin=0.035, ymax=0.105,
  ymode=log,
  ytick={0.04,0.05,0.06,0.08,0.1},
  yticklabels={0.04,0.05,0.06,0.08,0.1},
  xtick={0,100000,200000},
  xticklabels={0,100k,200k},
  scaled x ticks=false,
  grid=major,
  xlabel={Optimization step},
  ylabel={Loss (Train/Val.)},
  axis line style={PrettyText},
  tick label style={font=\ttfamily\scriptsize, PrettyText},
  label style={font=\small, PrettyText},
]
\addplot+[color=VLAColor, no marks, line width=0.9pt]
  table [x=step, y=loss_smooth] {figures/pgfplots/data/vla_loss_history_9k.tsv};
\addplot+[color=VLAColor, no marks, dashed, line width=1.0pt]
  table [x=step, y=val_loss_smooth] {figures/pgfplots/data/vla_loss_history_9k.tsv};
\end{axis}

\begin{axis}[
  at={(lossVlaOnly.south west)},
  anchor=south west,
  width=122mm, height=67mm,
  xmin=0, xmax=220000,
  ymin=0, ymax=70,
  xtick=\empty,
  axis x line=none,
  axis y line*=right,
  ytick={0,20,40,60},
  yticklabels={0\%,20\%,40\%,60\%},
  ylabel={Real progress (\%)},
  axis background/.style={fill=none},
  axis line style={DivOne},
  tick style={DivOne},
  yticklabel style={font=\ttfamily\scriptsize, DivOne},
  ylabel style={font=\small, DivOne},
]
\addplot+[color=DivOne, mark=diamond*, mark size=2.0pt, line width=1.0pt]
  table [x=step, y=progress] {figures/pgfplots/data/vla_progress_9k.tsv};
\end{axis}

\coordinate (legendbase) at ($(lossVlaOnly.south)+(0,-11mm)$);
\begin{scope}[font=\scriptsize, text=PrettyText]
  \draw[VLAColor, line width=0.9pt] ($(legendbase)+(-36mm,0mm)$) -- ($(legendbase)+(-28mm,0mm)$);
  \node[anchor=west] at ($(legendbase)+(-26mm,0mm)$) {Train Loss};
  \draw[VLAColor, dashed, line width=1.0pt] ($(legendbase)+(-1mm,0mm)$) -- ($(legendbase)+(7mm,0mm)$);
  \node[anchor=west] at ($(legendbase)+(9mm,0mm)$) {Val Loss};
  \draw[DivOne, line width=1.0pt] ($(legendbase)+(28mm,0mm)$) -- ($(legendbase)+(36mm,0mm)$);
  \filldraw[draw=DivOne, fill=DivOne] ($(legendbase)+(32mm,0mm)$) circle[radius=1.7pt];
  \node[anchor=west] at ($(legendbase)+(38mm,0mm)$) {Real Progress};
\end{scope}
\end{tikzpicture}%
          }
          \captionof{figure}{\textbf{VLA progress and offline metrics over time.} VLA training and validation loss with real-world progress across checkpoints.}
          \label{fig:vla-loss-progress}
      \end{minipage}
  \end{figure*}

\subsection{Offline Metrics for Policy Evaluation}
Real-world evaluation does not scale to the number of minor architecture and hyperparameter decisions involved in developing a policy, so in practice we make many of these using offline metrics and reserve real-world evaluation for the major architectural comparisons. This raises the question of whether offline metrics are predictive of real-world performance at all. 

For all the models trained in the previous sections, we compare the real-world strict success rates and progress rates against training loss, validation loss, and validation action error to assess whether offline metrics predict real-world performance. While training and validation loss are defined as the conditional flow matching loss of the models on their respective data splits, validation action error is the L2 distance between predicted and ground-truth action chunks, computed by running 10 diffusion steps per model with no action prefix. 

\begin{figure*}[t]
      \centering
      \resizebox{0.98\textwidth}{!}{%
          \input{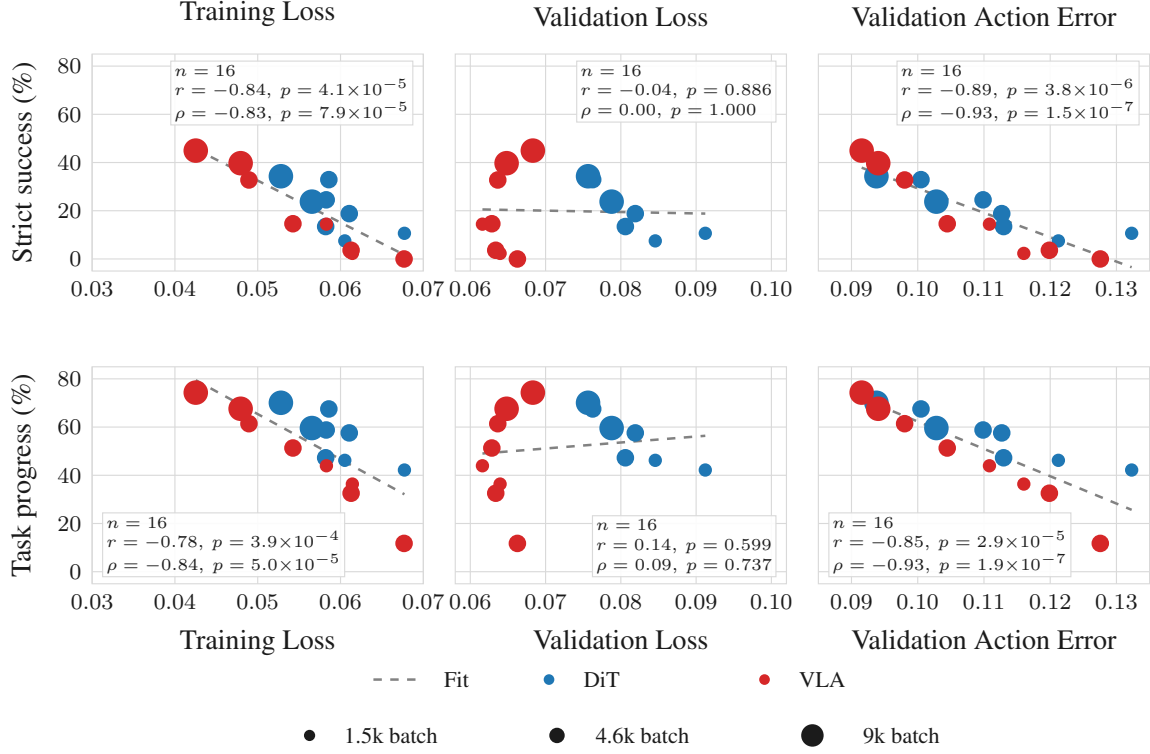}
      }
      \caption{\textbf{Offline metrics versus real-world policy performance.}
        We analyze correlation between checkpoint training diagnostics and real-world success. Training loss and validation action error are both negatively correlated with real-world performance. Each point is one checkpoint, with real-world success averaged across the three evaluation tasks.
      }
      \label{fig:loss_success_correlation}
  \end{figure*}

\textbf{Training loss and validation action error correlate with real-world performance.} The resulting plots are shown in Figure~\ref{fig:loss_success_correlation}. Training loss and validation action error are both negatively correlated with real-world performance, with statistically significant Pearson and Spearman correlations. Validation action error shows the strongest negative correlation. In contrast, validation loss is not significantly correlated with real-world performance. We note that validation action error is only meaningful when the number of diffusion steps is held fixed, since the error can be trivially reduced by lowering the number of diffusion steps, but this does not improve real-world performance. These correlations hold true across different model architectures and training batch sizes. Figure~\ref{fig:vla-loss-progress} shows an example of the phenomenon with the VLA---the training loss goes down over time as real-world progress improves; meanwhile, validation loss initially goes down but then increases even though performance improves over time. Offline metrics, such as validation action error and training loss, thus give us a cheap, scalar proxy for evaluating the many small design decisions in this section without having to test everything in the real world. While they cannot reliably predict task success rates or reveal failure modes, we still find them to be helpful for iterating on modeling decisions.
For a higher-fidelity proxy, we turn to simulation (Section ~\ref{sec:abc-sim}).

\section{ABC-Sim}
\label{sec:abc-sim}

\begin{center}
  \begin{minipage}{0.95\textwidth}
    \centering
    \resizebox{\linewidth}{!}{%
      \begin{tikzpicture}
\begin{groupplot}[
  group style={
    group name=combinedSimRealAgg,
    group size=2 by 1,
    horizontal sep=14mm,
  },
  set layers=standard,
  width=0.56\linewidth,
  height=54mm,
  grid=major,
  clip=false,
  axis line style={PrettyText},
  tick style={PrettyText},
  tick label style={font=\rmfamily\scriptsize, PrettyText},
  label style={font=\small, PrettyText},
  title style={font=\small, PrettyText},
  major grid style={gray!30},
]
\nextgroupplot[
  title={Strict success},
  xlabel={Sim strict success (\%)},
  ylabel={Real strict success (\%)},
  xmin=2.7366667, xmax=66.763333, ymin=5.3356073, ymax=45.64905,
  xtick distance=10, ytick distance=10,
]
\addplot+[color=PrettyText!55, dashed, no marks, line width=1.05pt]
  table [x=sim, y=real] {figures/pgfplots/data/sim_real_7k_1x_plus_3p5k_strict_success_all_fit.tsv};
\node[
  anchor=north west, align=left, fill=white, fill opacity=0.94, text opacity=1,
  draw=gray!35, inner sep=1.8pt, font=\rmfamily\tiny, text=PrettyText,
  on layer=axis foreground
] at (rel axis cs:0.025,0.975) {$n=12$\\$r=0.85$, $p=4.2{\times}10^{-4}$};
\addplot+[
  only marks, mark=*, mark size=2.45pt, color=LBMColor,
  mark options={solid, draw=PrettyText, fill=LBMColor, line width=0.3pt}
] table [x=sim, y=real] {figures/pgfplots/data/sim_real_7k_1x_plus_3p5k_strict_success_7k_1x.tsv};
\addplot+[
  only marks, mark=*, mark size=2.45pt, color=LBMColor,
  mark options={solid, draw=PrettyText, fill=LBMColor, line width=0.3pt}
] table [x=sim, y=real] {figures/pgfplots/data/sim_real_7k_1x_plus_3p5k_strict_success_3p5k_dit.tsv};
\addplot+[
  only marks, mark=*, mark size=2.45pt, color=VLAColor,
  mark options={solid, draw=PrettyText, fill=VLAColor, line width=0.3pt}
] table [x=sim, y=real] {figures/pgfplots/data/sim_real_7k_1x_plus_3p5k_strict_success_3p5k_vla.tsv};
\nextgroupplot[
  title={Task progress},
  xlabel={Sim task progress (\%)},
  ylabel={Real task progress (\%)},
  xmin=35.117222, xmax=81.410556, ymin=40.183128, ymax=73.333333,
  xtick distance=10, ytick distance=10,
 ]
\addplot+[color=PrettyText!55, dashed, no marks, line width=1.05pt]
  table [x=sim, y=real] {figures/pgfplots/data/sim_real_7k_1x_plus_3p5k_progress_all_fit.tsv};
\node[
  anchor=north west, align=left, fill=white, fill opacity=0.94, text opacity=1,
  draw=gray!35, inner sep=1.8pt, font=\rmfamily\tiny, text=PrettyText,
  on layer=axis foreground
] at (rel axis cs:0.025,0.975) {$n=12$\\$r=0.91$, $p=5.0{\times}10^{-5}$};
\addplot+[
  only marks, mark=*, mark size=2.45pt, color=LBMColor,
  mark options={solid, draw=PrettyText, fill=LBMColor, line width=0.3pt}
] table [x=sim, y=real] {figures/pgfplots/data/sim_real_7k_1x_plus_3p5k_progress_7k_1x.tsv};
\addplot+[
  only marks, mark=*, mark size=2.45pt, color=LBMColor,
  mark options={solid, draw=PrettyText, fill=LBMColor, line width=0.3pt}
] table [x=sim, y=real] {figures/pgfplots/data/sim_real_7k_1x_plus_3p5k_progress_3p5k_dit.tsv};
\addplot+[
  only marks, mark=*, mark size=2.45pt, color=VLAColor,
  mark options={solid, draw=PrettyText, fill=VLAColor, line width=0.3pt}
] table [x=sim, y=real] {figures/pgfplots/data/sim_real_7k_1x_plus_3p5k_progress_3p5k_vla.tsv};
\end{groupplot}

\coordinate (legendbase) at ($(combinedSimRealAgg c1r1.south)!0.5!(combinedSimRealAgg c2r1.south)+(0,-13mm)$);
\begin{scope}[font=\scriptsize, text=PrettyText]
  \draw[PrettyText!55, dashed, line width=1.05pt] ($(legendbase)+(-26mm,0mm)$) -- ($(legendbase)+(-21mm,0mm)$);
  \node[anchor=west] at ($(legendbase)+(-19.5mm,0mm)$) {Fit};
  \filldraw[draw=PrettyText, fill=LBMColor, line width=0.45pt] ($(legendbase)+(-3mm,0mm)$) circle[radius=1.1mm];
  \node[anchor=west] at ($(legendbase)+(-1mm,0mm)$) {DiT};
  \filldraw[draw=PrettyText, fill=VLAColor, line width=0.45pt] ($(legendbase)+(12mm,0mm)$) circle[radius=1.1mm];
  \node[anchor=west] at ($(legendbase)+(14mm,0mm)$) {VLA};
\end{scope}

\path
  ($(combinedSimRealAgg c1r1.north west)+(-2mm,2mm)$)
  rectangle
  ($(combinedSimRealAgg c2r1.south east)+(2mm,-21mm)$);
\end{tikzpicture}%
    }
    
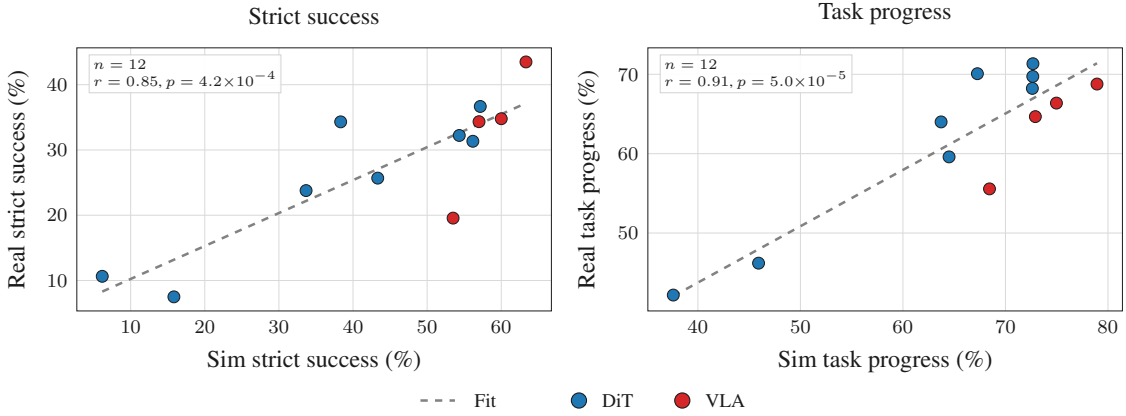
\captionof{figure}{\textbf{Sim-to-real performance correlation.}
    Task-level simulation progress correlation with real-world progress across DiT and VLA checkpoints.}
    \label{fig:sim_real_correlation}
  \end{minipage}
\end{center}

Offline metrics allow us to assess many decisions cheaply but cannot report task success or reveal failure modes. Simulation closes this gap as it yields actual success rates and watchable rollouts, and lets researchers without hardware iterate. Our release includes \textbf{ABC-Sim}---a suite of simulation environments and accompanying teleoperation data.

\begin{figure*}[t]
    \centering
    \newlength{\simw}\setlength{\simw}{0.31\textwidth}
    \newlength{\simg}\setlength{\simg}{0.005\textwidth}
    \newcommand{\simcell}[1]{\includegraphics[width=\simw]{#1}}
    \simcell{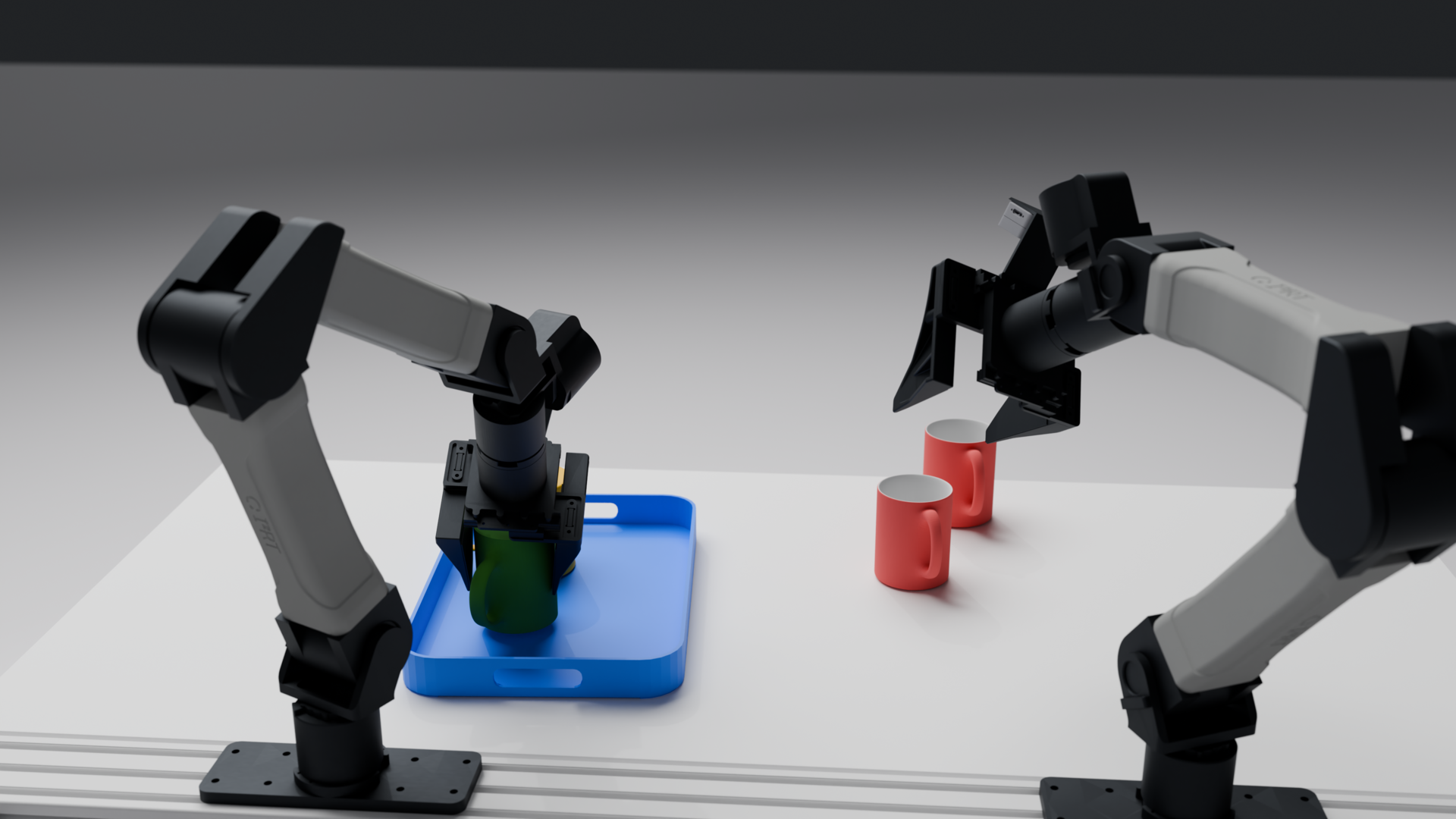}\hspace{\simg}%
    \simcell{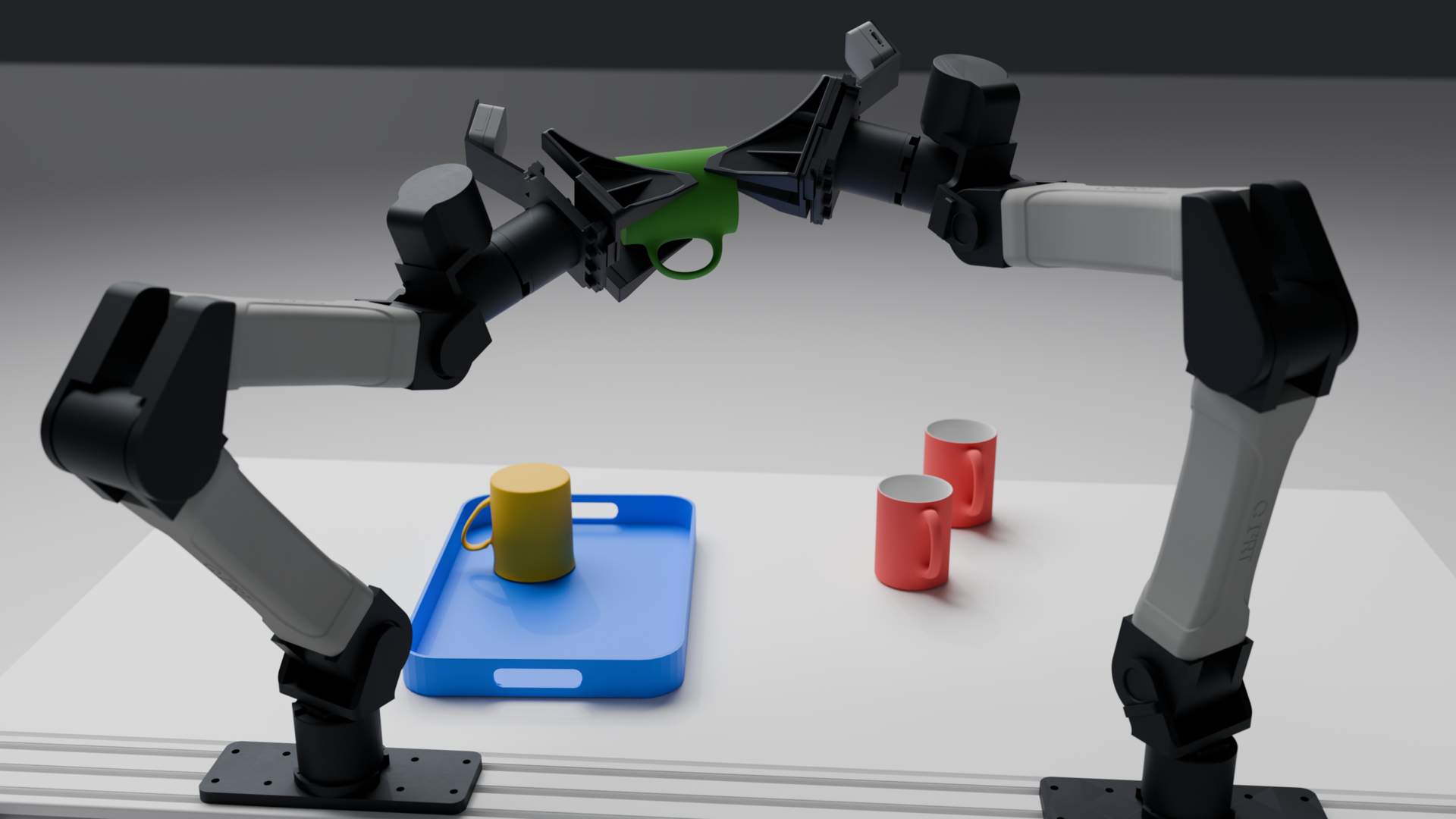}\hspace{\simg}%
    \simcell{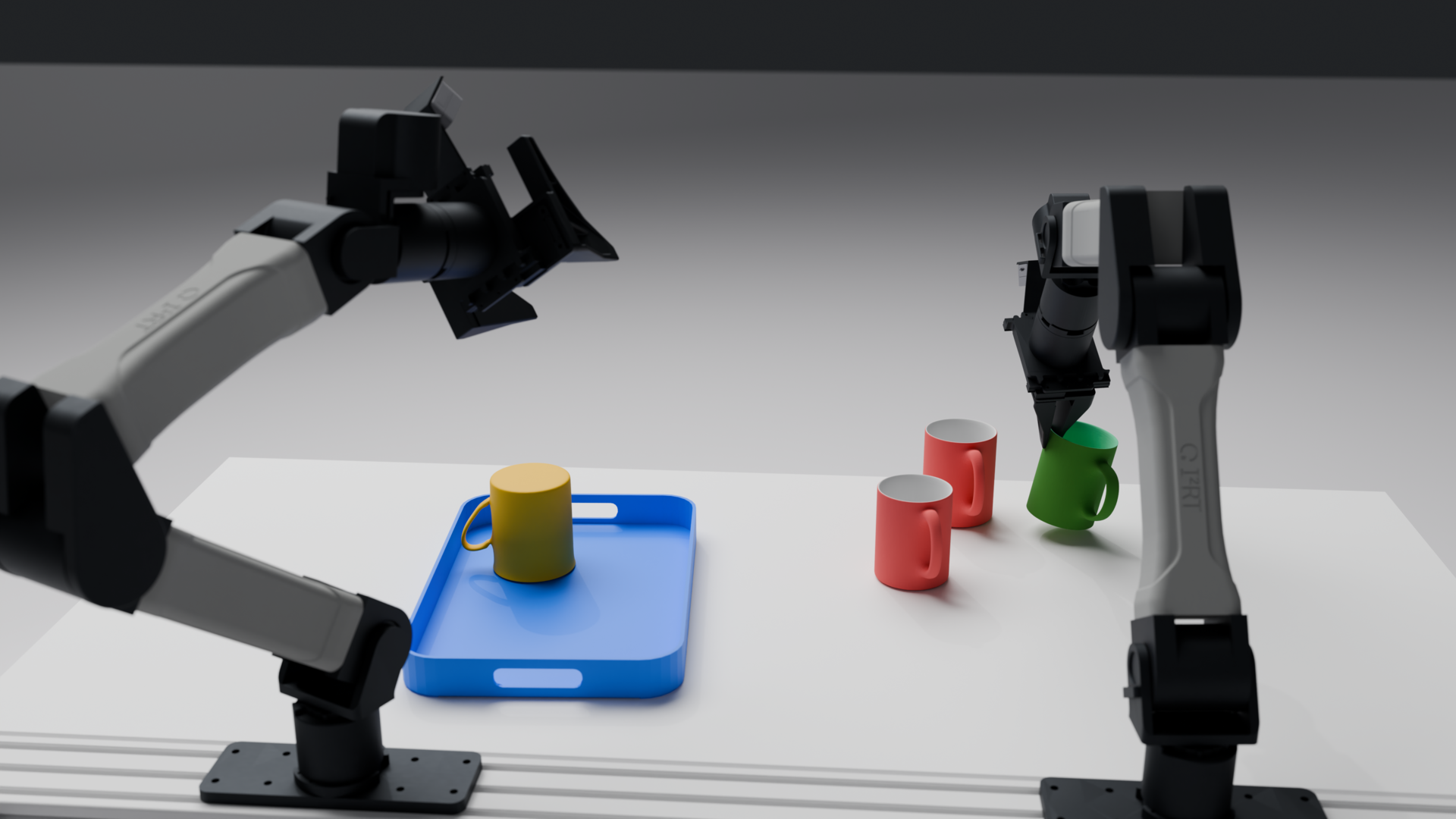}\\[2pt]
    \caption{\textbf{ABC-Sim tasks.} Policy rollouts for our simulated tasks built in MuJoCo for sim-to-real evaluation. The multi-step \textit{turning mugs right-side up} task, shown across three stages (grasp, flip, place). Rollouts on more simulated tasks can be found in the Appendix. By default we render the rollouts in MuJoCo but we also provide a Blender pipeline for generating high-quality renders.}
    \label{fig:abc-sim-tasks}

    \vspace{-1em}
\end{figure*}

We re-create the real robot setup in MuJoCo~\cite{todorov2012mujoco}. The images are rendered inside the simulation, leading to low-fidelity captures. We also expose a pipeline for re-rendering these scenes in Blender to create higher-fidelity images with path tracing. This allows users to both generate and train on higher-quality simulation renderings. We release
simulation environments for the following ten tasks (Fig.~\ref{fig:abc-sim-tasks}):
\begin{multicols}{2}
\begin{enumerate}[
    label=\textbf{\arabic*.},
    leftmargin=1.6em,
    itemsep=0.15em,
    topsep=0.25em,
    parsep=0pt
]
    \item throwing plastic bottles into a bin
    \item sweeping paper scraps off the table
    \item turning mugs right-side up
    \item loading plates into tabletop dishrack
    \item hanging a mug on a mug rack
    \item setting up chess pieces on a board
    \item spelling ``abc'' with blocks
    \item placing markers in a drawer
    \item pouring beads
    \item in-hand object handover
\end{enumerate}
\end{multicols}

The simulation runs at 240Hz and visual observations are sent to a VR headset at 60Hz. The current framework supports both Meta Quest and Apple Vision Pro headsets. The teleoperator uses a pair of GELLO~\cite{wu2024gello} leader arms in order to teleoperate the YAM arms in simulation to complete the various tasks. VR-based teleoperation for extended periods of time can lead to motion sickness in operators. To mitigate this, we enable pass-through rendering so that the operators view the real world behind the simulated robot setup. We release over 400 hours of VR teleoperation data in simulation on all of these tasks, as well as the VR teleoperation stack to collect teleop data similar to \cite{Park2024DexHubAD, ravan2026lucidxrextendedrealitydataengine, jiang2025irisimmersiverobotinteraction}.

To assess whether simulation performance is indicative of real-world performance, we conduct evaluations for sim-to-real correlations. We train \textit{ABC-DiT} and \textit{ABC-VLA} on ABC-130K and select four checkpoints per model. We also select four different \textit{ABC-DiT} checkpoints trained on a mixture of our internal 7K hour dataset and the simulation data for a total of 12 checkpoints. We evaluate each of them across the same three tasks used in Section~\ref{sec:abc-models}: \textit{throwing
plastic bottles into a bin}, \textit{loading plates into tabletop dishrack}, and
\textit{turning mugs right-side up}. Each task was evaluated for 50 trials. For both simulation and real results, we average the success rates and progress across the tasks for each checkpoint. 

\textbf{Simulation evaluations correlate strongly with real-world performance}. The results of the evaluation are shown in Figure~\ref{fig:sim_real_correlation}. For the strict success metric, we observe a Pearson correlation of $r=0.85$ ($p=4.2\times10^{-4}$, $n=12$) and for task progress we observe $r=0.91$ ($p=5.0\times10^{-5}$, $n=12$). This indicates that simulation is a meaningful predictor of real performance. We hope that given this correlation, our sim stack can provide a way for researchers to debug design decisions in their policies prior to conducting expensive real-world evaluations, or for those without hardware to contribute to the research frontier. 

\section{Real-World Capabilities}
\label{sec:capabilities}

\newsavebox{\realbarsbox}
\newsavebox{\simprogbox}
\newlength{\barsimmaxh}
\begin{figure*}[t]
      \centering
      \captionsetup{justification=raggedright,singlelinecheck=false}

      \sbox{\realbarsbox}{\resizebox{0.39\textwidth}{!}{%
        \begin{tikzpicture}
\begin{axis}[
  width=64mm,
  height=45mm,
  ybar=2pt,
  bar width=8pt,
  enlarge x limits={abs=11pt},
  ymin=0, ymax=100,
  ytick={0,20,40,60,80,100},
  ylabel={Real progress (\%)},
  symbolic x coords={Bottles,Dishrack,Mug flip},
  xtick=data,
  tick pos=left,
  axis line style={PrettyText},
  tick style={PrettyText},
  tick label style={font=\ttfamily\scriptsize, PrettyText},
  x tick label style={font=\scriptsize, PrettyText},
  label style={font=\small, PrettyText},
  major grid style={gray!35},
  ymajorgrids,
  legend style={
    at={(0.98,0.98)}, anchor=north east,
    legend columns=1, draw=gray!35, fill=white, fill opacity=0.92,
    text opacity=1.0, font=\tiny, text=PrettyText,
    inner sep=1.5pt, inner xsep=2pt, inner ysep=2pt, row sep=1pt,
    nodes={inner sep=1pt, anchor=west}, legend cell align=left,
  },
  legend image code/.code={
    \draw [#1, draw=none] (0cm,-0.08cm) rectangle (0.32cm,0.18cm);
  },
]
\addplot+[fill=LBMColor, draw=none]
  table [x=task, y=dit, col sep=tab] {figures/pgfplots/data/real_200k_progress_bars.tsv};
\addlegendentry{ABC-DiT}

\addplot+[fill=VLAColor, draw=none]
  table [x=task, y=vla, col sep=tab] {figures/pgfplots/data/real_200k_progress_bars.tsv};
\addlegendentry{ABC-VLA}
\end{axis}
\end{tikzpicture}}}%
      \sbox{\simprogbox}{\resizebox{0.57\textwidth}{!}{%
        \begin{tikzpicture}
\begin{groupplot}[
  group style={group size=3 by 1, horizontal sep=3.5mm},
  width=49mm, height=43mm,
  xmin=0, xmax=210,
  ymin=0, ymax=100,
  xtick={0,50,100,150,200},
  ytick={0,25,50,75,100},
  grid=major,
  xlabel={Checkpoint step (k)},
  axis line style={PrettyText},
  tick style={PrettyText},
  tick label style={font=\ttfamily\scriptsize, PrettyText},
  label style={font=\small, PrettyText},
  title style={font=\small, PrettyText},
]
\nextgroupplot[
  title={Bottles},
  ylabel={Sim progress (\%)},
]
\addplot+[
  color=LBMColor,
  mark=square*,
  mark size=1.35pt,
  line width=0.95pt,
  mark options={draw=none, fill=LBMColor},
] table [x=step_k, y=progress] {figures/pgfplots/data/sim_eval_progress_lbm9216_bottles_sparse2x.tsv};
\addplot+[
  color=VLAColor,
  mark=triangle*,
  mark size=1.5pt,
  line width=0.95pt,
  mark options={draw=none, fill=VLAColor},
] table [x=step_k, y=progress] {figures/pgfplots/data/sim_eval_progress_vla9216_bottles_sparse2x.tsv};

\nextgroupplot[
  title={Dishrack},
  yticklabels={},
]
\addplot+[
  color=LBMColor,
  mark=square*,
  mark size=1.35pt,
  line width=0.95pt,
  mark options={draw=none, fill=LBMColor},
] table [x=step_k, y=progress] {figures/pgfplots/data/sim_eval_progress_lbm9216_dishrack_sparse2x.tsv};
\addplot+[
  color=VLAColor,
  mark=triangle*,
  mark size=1.5pt,
  line width=0.95pt,
  mark options={draw=none, fill=VLAColor},
] table [x=step_k, y=progress] {figures/pgfplots/data/sim_eval_progress_vla9216_dishrack_sparse2x.tsv};

\nextgroupplot[
  title={Mug flip},
  yticklabels={},
]
\addplot+[
  color=LBMColor,
  mark=square*,
  mark size=1.35pt,
  line width=0.95pt,
  mark options={draw=none, fill=LBMColor},
] table [x=step_k, y=progress] {figures/pgfplots/data/sim_eval_progress_lbm9216_mug_sparse2x.tsv};
\addplot+[
  color=VLAColor,
  mark=triangle*,
  mark size=1.5pt,
  line width=0.95pt,
  mark options={draw=none, fill=VLAColor},
] table [x=step_k, y=progress] {figures/pgfplots/data/sim_eval_progress_vla9216_mug_sparse2x.tsv};
\end{groupplot}

\coordinate (legendbase) at ($(group c2r1.south)+(0,-12mm)$);
\begin{scope}[font=\scriptsize, text=PrettyText]
  \draw[LBMColor, line width=0.95pt] ($(legendbase)+(-28mm,0mm)$) -- ($(legendbase)+(-18mm,0mm)$);
  \node[
    rectangle,
    fill=LBMColor,
    draw=none,
    minimum size=2.9pt,
    inner sep=0pt
  ] at ($(legendbase)+(-23mm,0mm)$) {};
  \node[anchor=west] at ($(legendbase)+(-16mm,0mm)$) {DiT 9.2kBS};

  \draw[VLAColor, line width=0.95pt] ($(legendbase)+(11mm,0mm)$) -- ($(legendbase)+(21mm,0mm)$);
  \path[fill=VLAColor, draw=none]
    ($(legendbase)+(16mm,1.65pt)$) --
    ($(legendbase)+(16mm,0mm)+(1.65pt,-1.65pt)$) --
    ($(legendbase)+(16mm,0mm)+(-1.65pt,-1.65pt)$) -- cycle;
  \node[anchor=west] at ($(legendbase)+(23mm,0mm)$) {VLA 9.2kBS};
\end{scope}
\end{tikzpicture}}}%
      \setlength{\barsimmaxh}{\dimexpr\ht\realbarsbox+\dp\realbarsbox\relax}%
      \ifdim\dimexpr\ht\simprogbox+\dp\simprogbox\relax>\barsimmaxh
        \setlength{\barsimmaxh}{\dimexpr\ht\simprogbox+\dp\simprogbox\relax}%
      \fi

      \begin{minipage}[t]{0.39\textwidth}
          \vspace*{0pt}
          \centering
          \parbox[t][\barsimmaxh][t]{\linewidth}{\centering\usebox{\realbarsbox}}
          \caption{\textbf{ABC-DiT/ABC-VLA real-world performance.}
          Task progress across the three real-world tasks.}
          \label{fig:real-200k-bars}
      \end{minipage}\hfill%
      \begin{minipage}[t]{0.57\textwidth}
          \vspace*{0pt}
          \centering
          \parbox[t][\barsimmaxh][t]{\linewidth}{\centering\usebox{\simprogbox}}
          \caption{\textbf{Pretraining simulation progress.}
          Mean task progress across three tasks in simulation at multiple checkpoints during pretraining of \textit{ABC-DiT} and \textit{ABC-VLA} on ABC-130K.}
          \label{fig:final_pt_progress}
      \end{minipage}

\end{figure*}

In this section, we demonstrate the capabilities of our trained policies on the publicly released ABC-130K set. We begin by showing the performance of the pretrained base policies in Section~\ref{sec:results_pretrain}. Section~\ref{sec:finetuning} shows the importance of pretraining when finetuning on dexterous tasks. Lastly, in Section~\ref{sec:dagger}, we study the case of box folding, a task for which naive finetuning is insufficient for real-world performance, and how DAgger can be used to specifically collect recovery behaviors. 
Appendix~\ref{app:conditioning} studies how conditioning can steer policy behavior at inference time---selecting an operator's style, trading off motion smoothness against visual responsiveness, and resolving temporal ambiguity within a task. Two of these are enabled directly by metadata we release with ABC-130K---per-episode operator identities and subtask annotations---illustrating how richer dataset labels, not just more trajectories, translate into finer-grained control at test-time.

\subsection{Pretraining Capabilities}
\label{sec:results_pretrain}
We train the \textit{ABC-DiT} and \textit{ABC-VLA} models, which were, respectively, the best DiT and VLA architectures found in Section~\ref{sec:abc-models}, on ABC-130K for 200{,}000 steps with the same training settings as in Section~\ref{sec:abc_models}.  In Figure~\ref{fig:final_pt_progress}, we plot the progress of three tasks across various points during training. We can see that for both models, as training iterations continue, simulation performance tends to increase. We also evaluate each task 50 times in the real world in Figure~\ref{fig:real-200k-bars}.

\subsection{Downstream single-task Finetuning}
\label{sec:finetuning}
We test the effectiveness of our pretrained policies as a useful initialization for
downstream finetuning. We choose four high-precision, dexterous tasks for this
comparison: i) extracting a credit card from a wallet, ii) sorting LEGO bricks, iii) inserting a pen cap, and iv) unscrewing a bottle cap. For each task, we train three
single-task policies that differ only in their initialization: (i) trained from
scratch on the target task; (ii) finetuned from a checkpoint pretrained on the public
ABC-130K (3{,}500 hours); and (iii) finetuned from a checkpoint pretrained on a
larger internal corpus of 7{,}000 hours, included to test whether the returns
from pretraining continue to scale. Figure~\ref{fig:finetuning-pretraining}
reports these results and Appendix~\ref{sec:evaluation} details our evaluation rubrics. We use ABC-DiT for the LEGO sorting, pen-cap insertion, and bottle-cap unscrewing, and ABC-VLA for the credit-card extraction task. We found that ABC-VLA was markedly better as a base model compared to ABC-DiT for this task. 
Pretraining quality has a substantial effect on downstream performance. Across
all tasks, finetuning from a checkpoint pretrained on more data yields a
stronger single-task policy. More details about the task, rollouts, evaluation rubric, and finetuning recipe can be found in Appendix~\ref{app:single-task-finetuning}.
\begin{figure*}[!b]
    \centering
    \captionsetup{justification=raggedright,singlelinecheck=false}
      \begin{minipage}[t]{0.58\textwidth}
          \vspace*{0pt}
          \centering
          \begin{tikzpicture}
\begin{axis}[
  width=\linewidth, height=45mm,
  ybar=2pt,
  bar width=8pt,
  enlarge x limits={abs=18pt},
  ymin=0, ymax=1.0,
  ytick={0, 0.2, 0.4, 0.6, 0.8, 1.0},
  yticklabel={\pgfmathprintnumber[fixed,fixed zerofill,precision=1]{\tick}},
  ylabel={Mean progress},
  symbolic x coords={Average, Sort_LEGOs, Unscrew_Bottle_Caps, Pen_Caps, Credit_Cards},
  xtick={Average, Sort_LEGOs, Unscrew_Bottle_Caps, Pen_Caps, Credit_Cards},
  xticklabels={Average,{Sort\\LEGOs},{Unscrew\\bottle caps},{Pen\\caps},{Credit\\cards}},
  xmin=Average, xmax=Credit_Cards,
  axis line style={PrettyText},
  tick style={PrettyText},
  tick label style={font=\ttfamily\scriptsize, PrettyText},
  x tick label style={
    font=\scriptsize,
    PrettyText,
    align=center,
    text width=16mm,
    yshift=-1pt,
  },
  label style={font=\small, PrettyText},
  major grid style={gray!35},
  ymajorgrids,
  legend style={
    at={(0.98,0.98)}, anchor=north east,
    legend columns=1, draw=gray!35, fill=white, fill opacity=0.92,
    text opacity=1.0,
    font=\tiny,
    text=PrettyText,
    inner sep=1.5pt,
    inner xsep=2pt, inner ysep=2pt,
    row sep=1pt,
    nodes={inner sep=1pt, anchor=west},
    legend cell align=left,
  },
  legend image code/.code={
    \draw [#1, draw=none] (0cm,-0.08cm) rectangle (0.32cm,0.18cm);
  },
]
\addplot+[fill=PrettyText!25, draw=none,
          skip coords between index={4}{5}]
  table [x=task, y=scratch, col sep=tab]
  {figures/pgfplots/data/finetuning_pretraining_effect.tsv};
\addlegendentry{Scratch}

\addplot+[fill=VLAColor, draw=none,
          skip coords between index={4}{5}]
  table [x=task, y=base_3p5k, col sep=tab]
  {figures/pgfplots/data/finetuning_pretraining_effect.tsv};
\addlegendentry{ABC-Data (3.5K h)}

\addplot+[fill=LBMColor, draw=none,
          skip coords between index={4}{5}]
  table [x=task, y=base_7k, col sep=tab]
  {figures/pgfplots/data/finetuning_pretraining_effect.tsv};
\addlegendentry{Pretrain 7K h}

\fill[PrettyText!25]
  ([xshift=-14pt] axis cs:Average,0)
  rectangle ([xshift=-6pt] axis cs:Average,0.24125);
\draw[
  pattern=north east lines, pattern color=PrettyText!50,
  draw=PrettyText!35, line width=0.35pt,
] ([xshift=-14pt] axis cs:Average,0)
  rectangle ([xshift=-6pt] axis cs:Average,0.24125);

\fill[VLAColor]
  ([xshift=-4pt] axis cs:Average,0)
  rectangle ([xshift=4pt] axis cs:Average,0.43);
\draw[
  pattern=north east lines, pattern color=VLAColor!50!black,
  draw=PrettyText!35, line width=0.35pt,
] ([xshift=-4pt] axis cs:Average,0)
  rectangle ([xshift=4pt] axis cs:Average,0.43);

\fill[LBMColor]
  ([xshift=6pt] axis cs:Average,0)
  rectangle ([xshift=14pt] axis cs:Average,0.61);
\draw[
  pattern=north east lines, pattern color=LBMColor!50!black,
  draw=PrettyText!35, line width=0.35pt,
] ([xshift=6pt] axis cs:Average,0)
  rectangle ([xshift=14pt] axis cs:Average,0.61);
\end{axis}
\end{tikzpicture}%
          \caption{\textbf{Effect of pretraining on downstream finetuning.}
          Single-task finetuning performance improves with more diverse pretraining across tasks.}
          \label{fig:finetuning-pretraining}
      \end{minipage}\hspace{0.05\textwidth}
    \begin{minipage}[t]{0.32\textwidth}
        \vspace*{0pt}
        \centering
        \begin{tikzpicture}
\begin{axis}[
  width=\linewidth, height=45mm,
  ybar=2pt,
  bar width=10pt,
  xmin=0.5,
  xmax=1.5,
  ymin=0, ymax=1.0,
  ytick={0, 0.2, 0.4, 0.6, 0.8, 1.0},
  yticklabel={\pgfmathprintnumber[fixed,fixed zerofill,precision=1]{\tick}},
  ylabel={Mean progress},
  xtick={0.8, 1.2},
  xticklabels={%
    {Finetuning},%
    {\shortstack{After\\DAgger}}%
  },
  axis line style={PrettyText},
  tick style={PrettyText},
  tick label style={font=\ttfamily\scriptsize, PrettyText},
  x tick label style={font=\scriptsize, PrettyText, align=center, yshift=-1pt},
  label style={font=\small, PrettyText},
  major grid style={gray!35},
  ymajorgrids,
]
\addplot+[fill=LBMColor, draw=none, bar shift=0pt]
  coordinates {(0.8, 0.24)};
\addplot+[fill=LBMColor, draw=none, bar shift=0pt]
  coordinates {(1.2, 0.85)};
\end{axis}
\end{tikzpicture}
        \caption{\textbf{Box folding: effect of DAgger.}
        Average task progress for a finetuned policy, before and after DAgger.}
        \label{fig:box-folding-dagger}
    \end{minipage}
    
\end{figure*}

\subsection{DAgger}
\label{sec:dagger}

\begin{center}
    \begingroup
    \setlength{\tabcolsep}{0pt}
    \renewcommand{\arraystretch}{0}
    \newlength{\boxfoldgutter}
    \setlength{\boxfoldgutter}{0.004\linewidth}
    \begin{tabular}{@{}c@{\hspace{\boxfoldgutter}}c@{\hspace{\boxfoldgutter}}c@{\hspace{\boxfoldgutter}}c@{}}
        \includegraphics[width=0.247\linewidth]{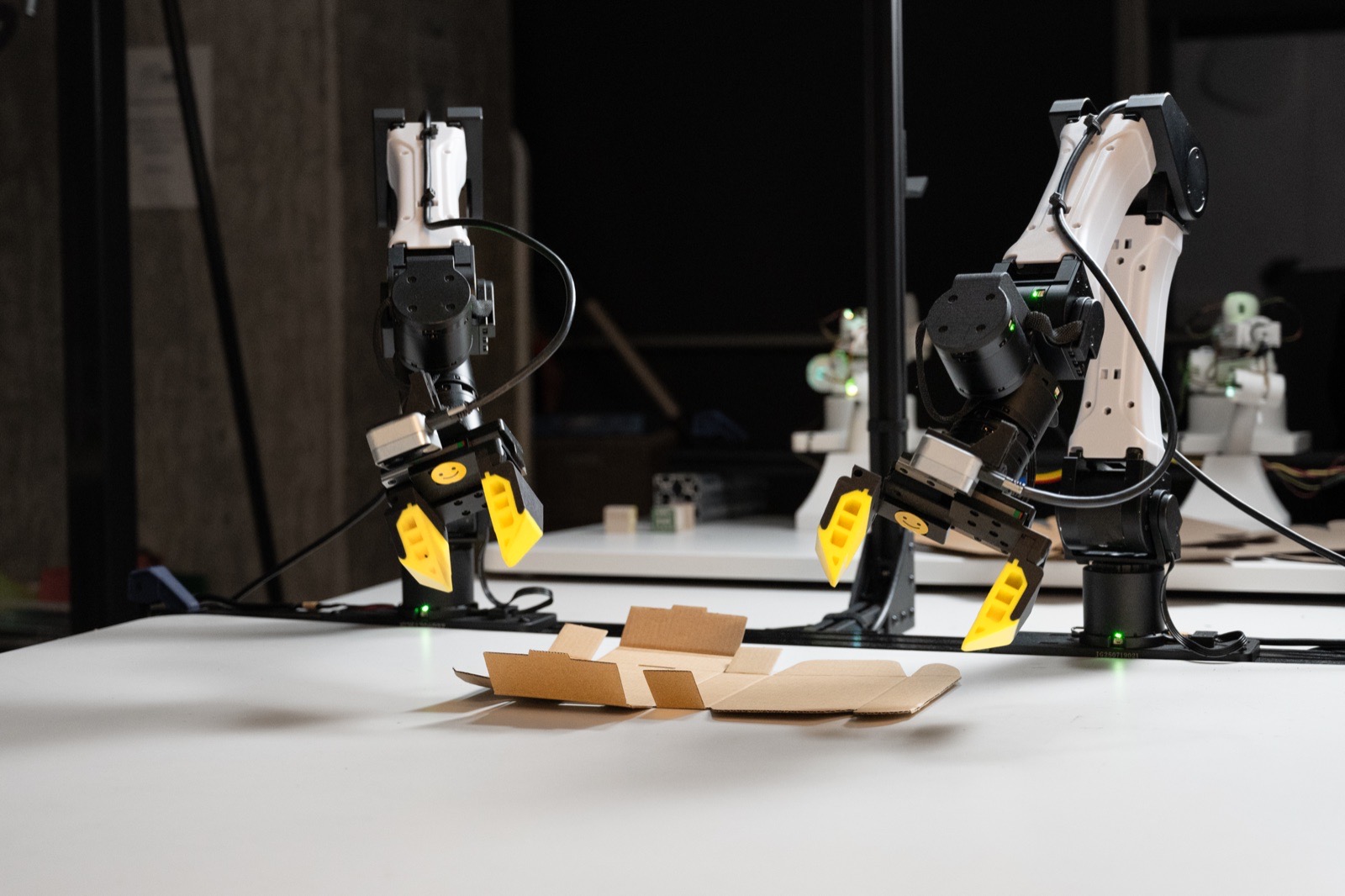} &
        \includegraphics[width=0.247\linewidth]{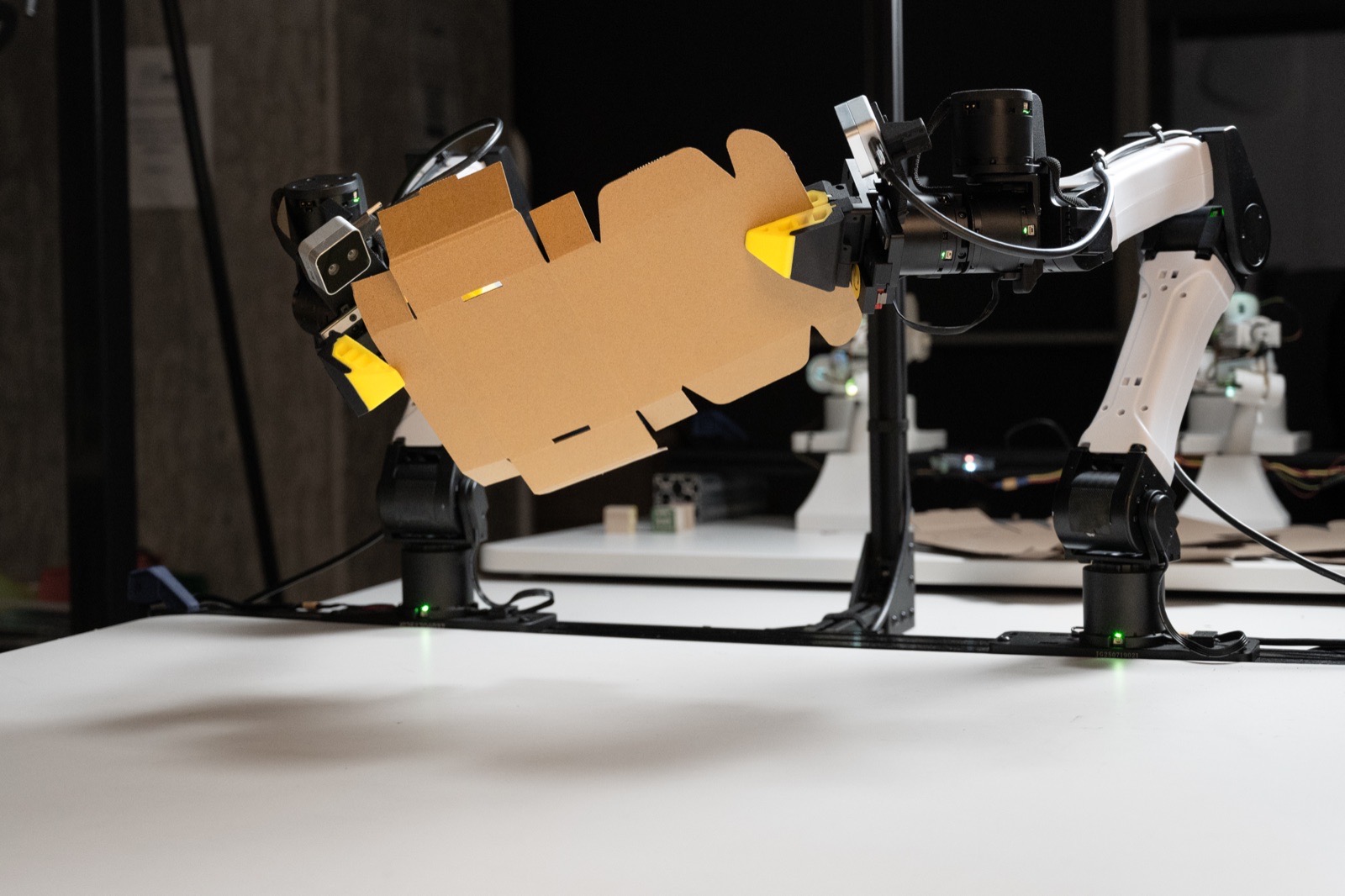} &
        \includegraphics[width=0.247\linewidth]{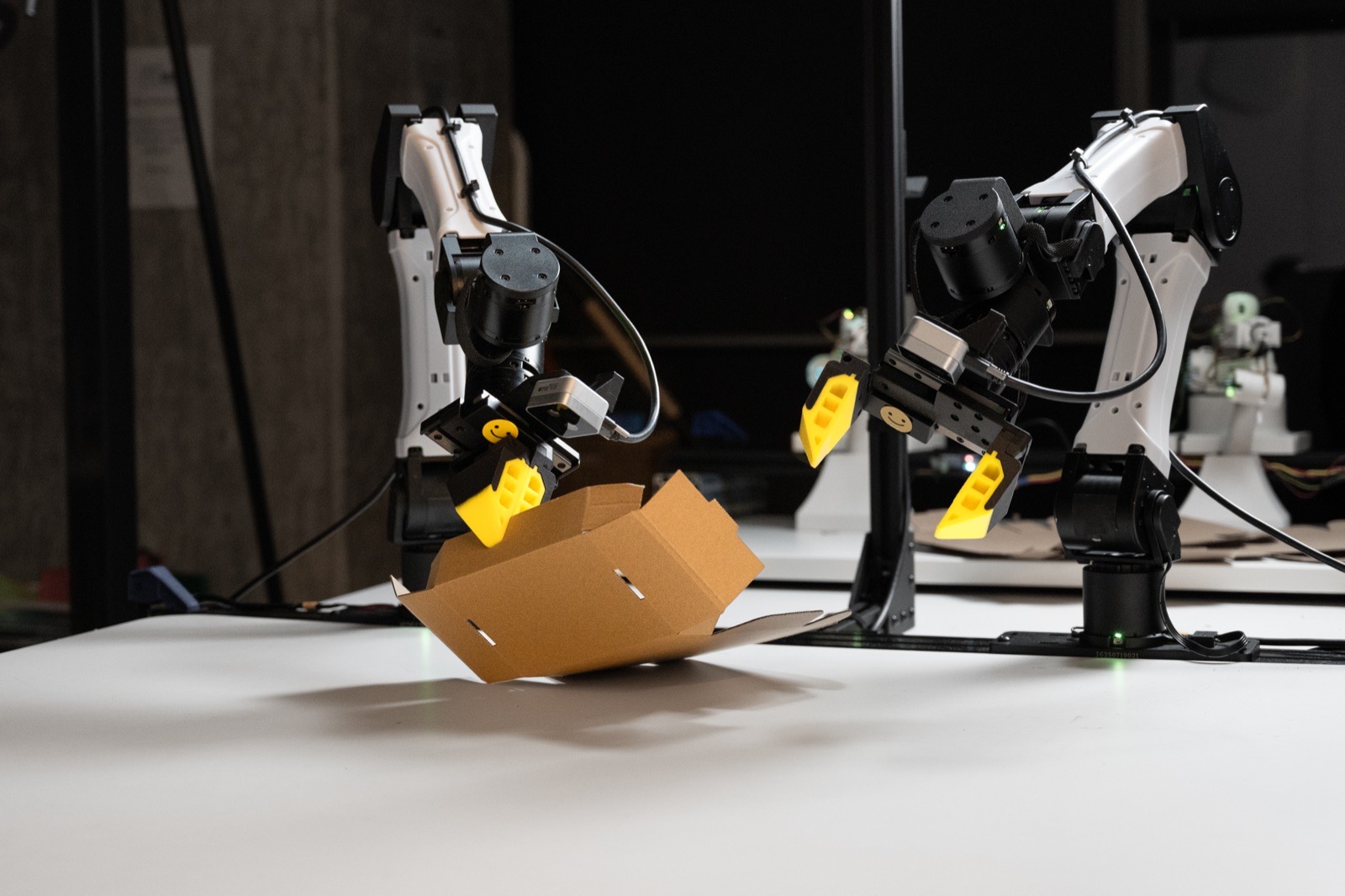} &
        \includegraphics[width=0.247\linewidth]{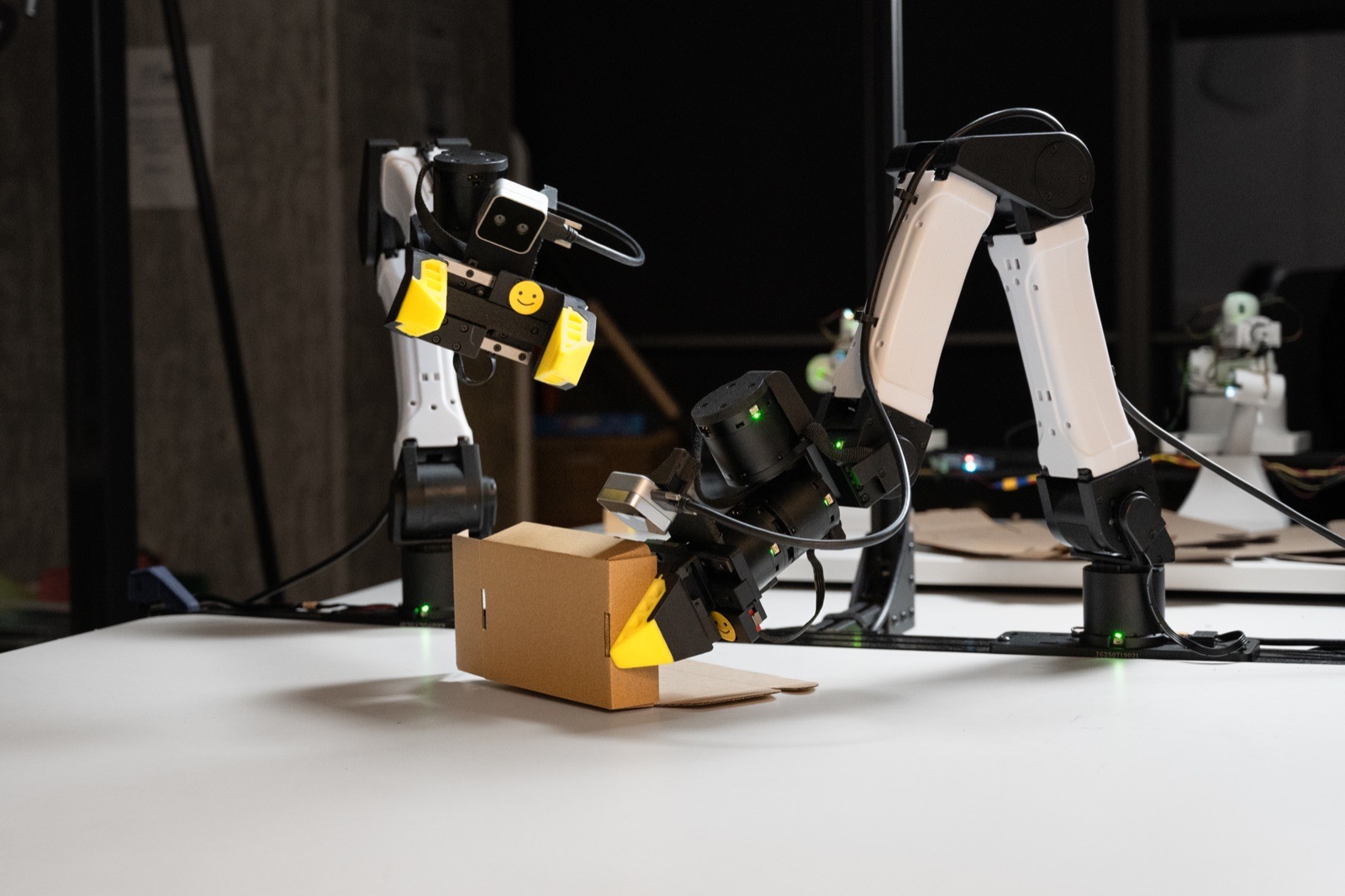} \\[\boxfoldgutter]
        \includegraphics[width=0.247\linewidth]{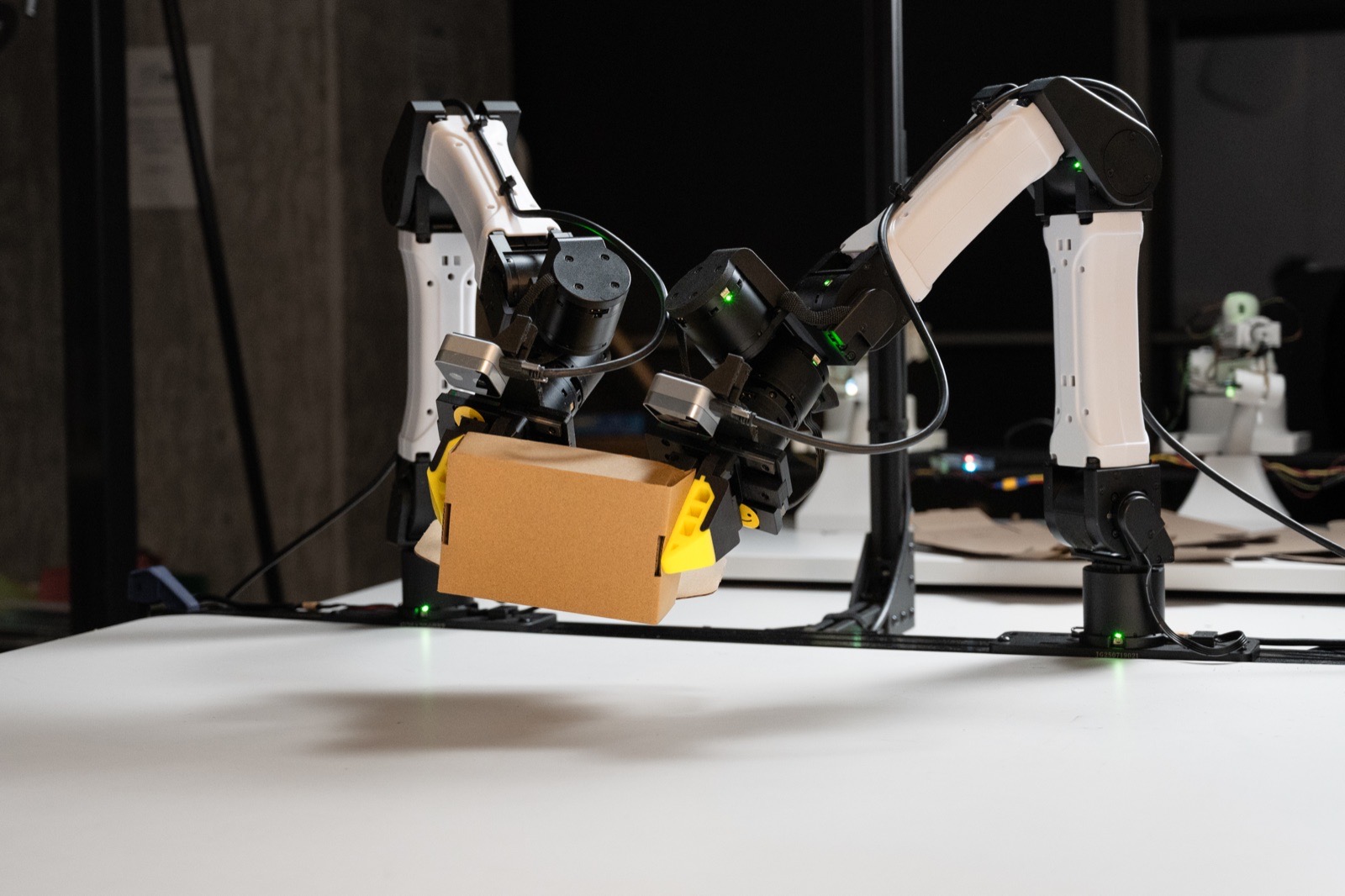} &
        \includegraphics[width=0.247\linewidth]{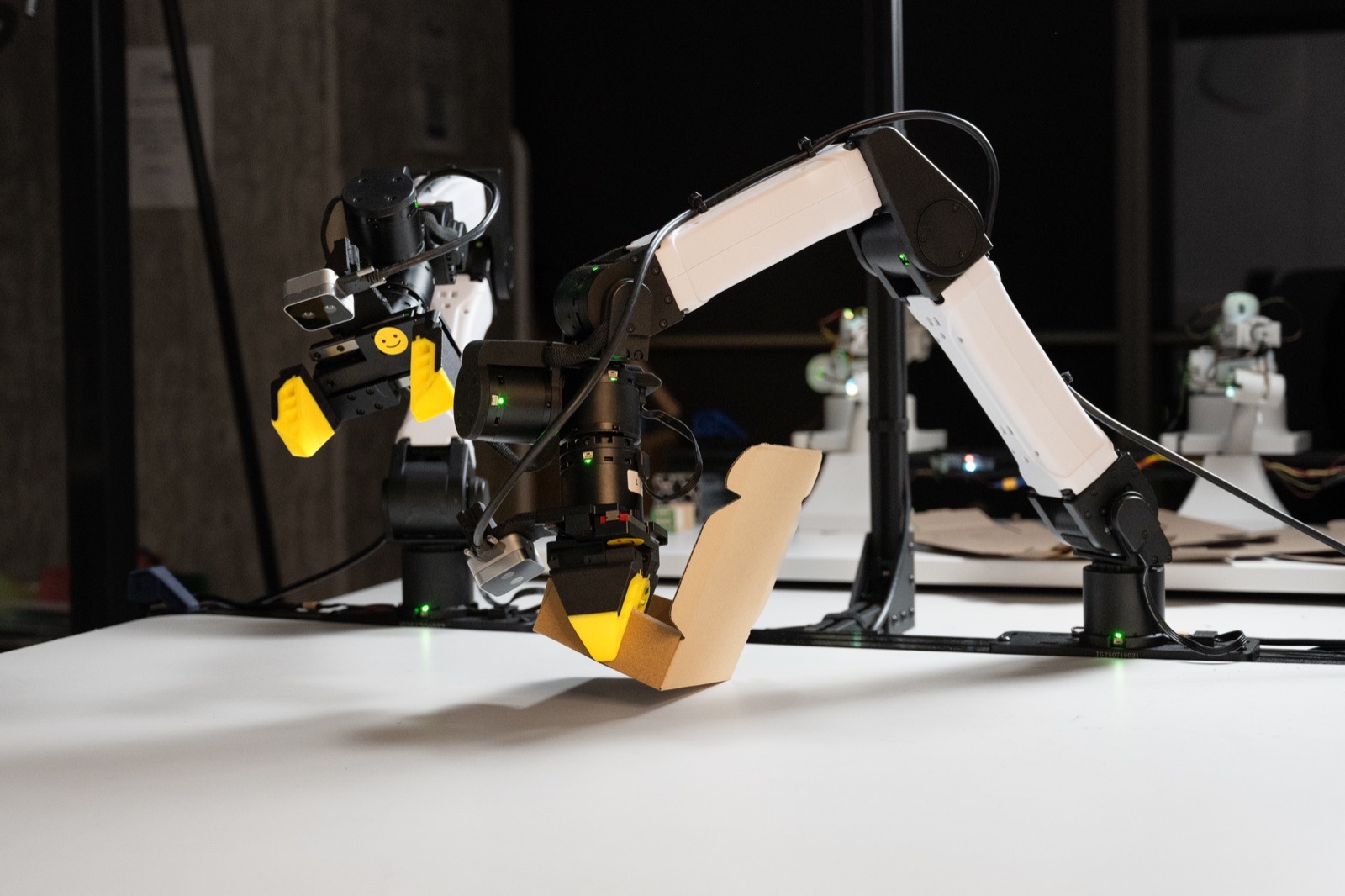} &
        \includegraphics[width=0.247\linewidth]{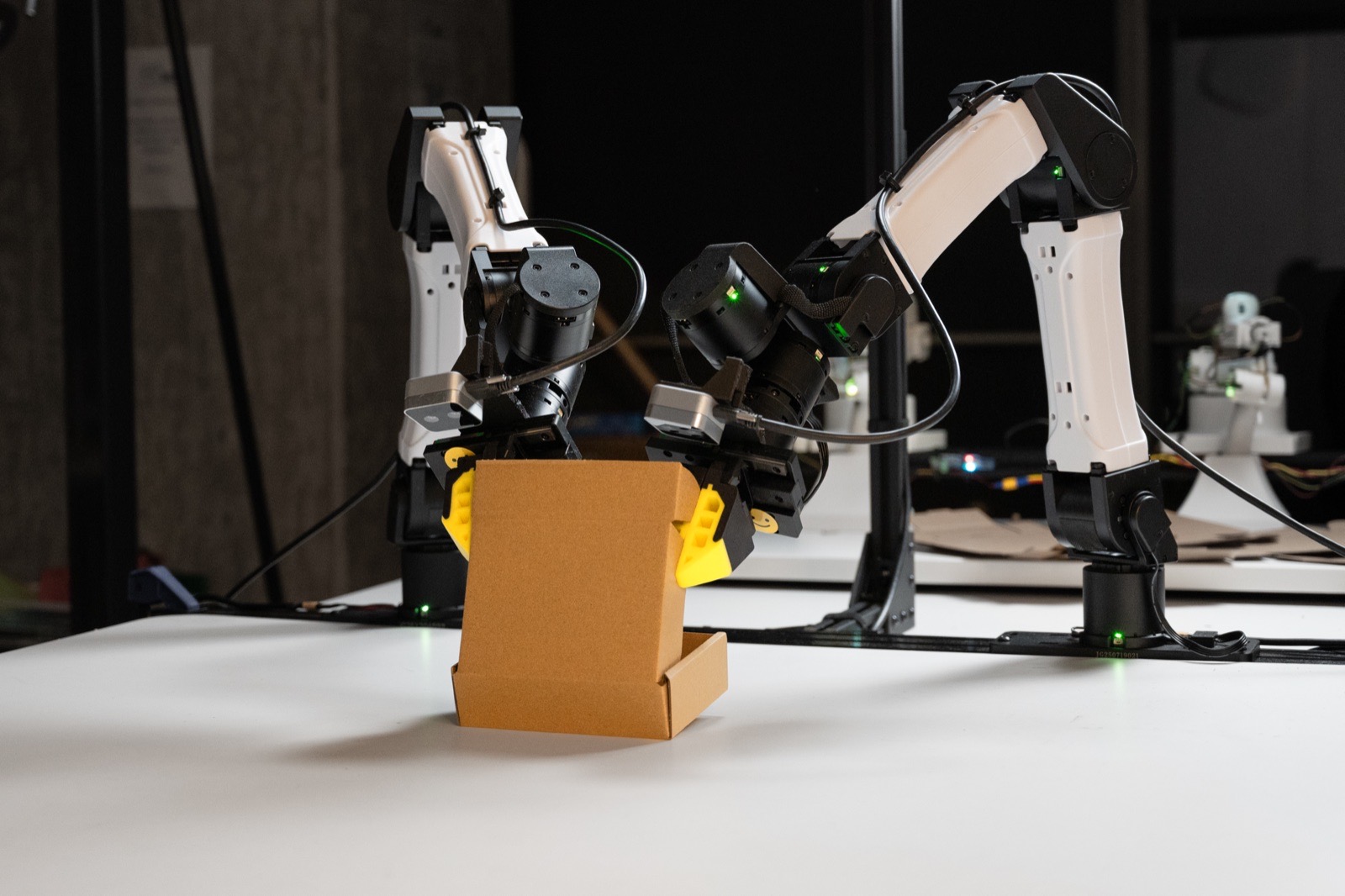} &
        \includegraphics[width=0.247\linewidth]{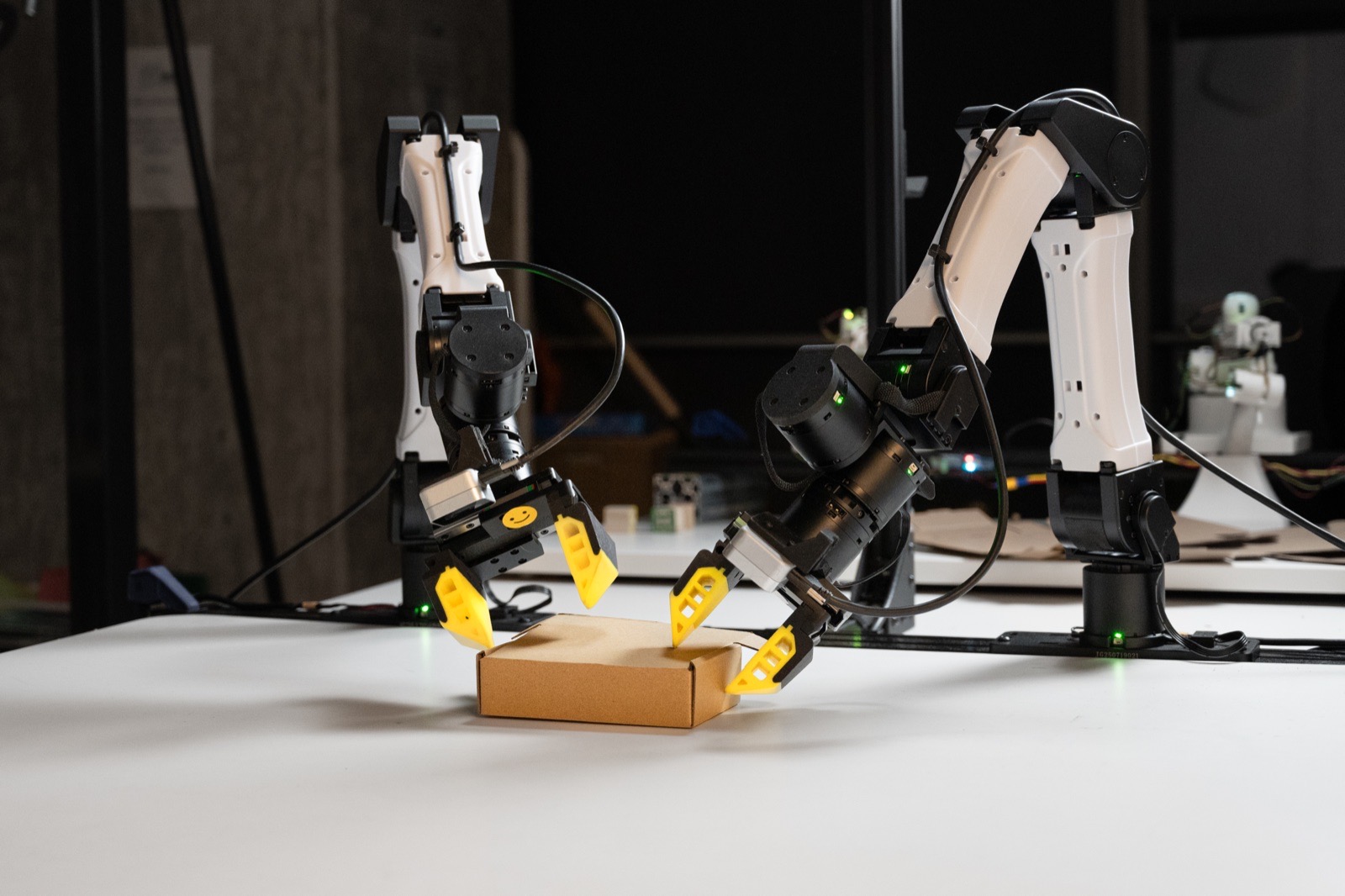} \\[\boxfoldgutter]
        \includegraphics[width=0.247\linewidth]{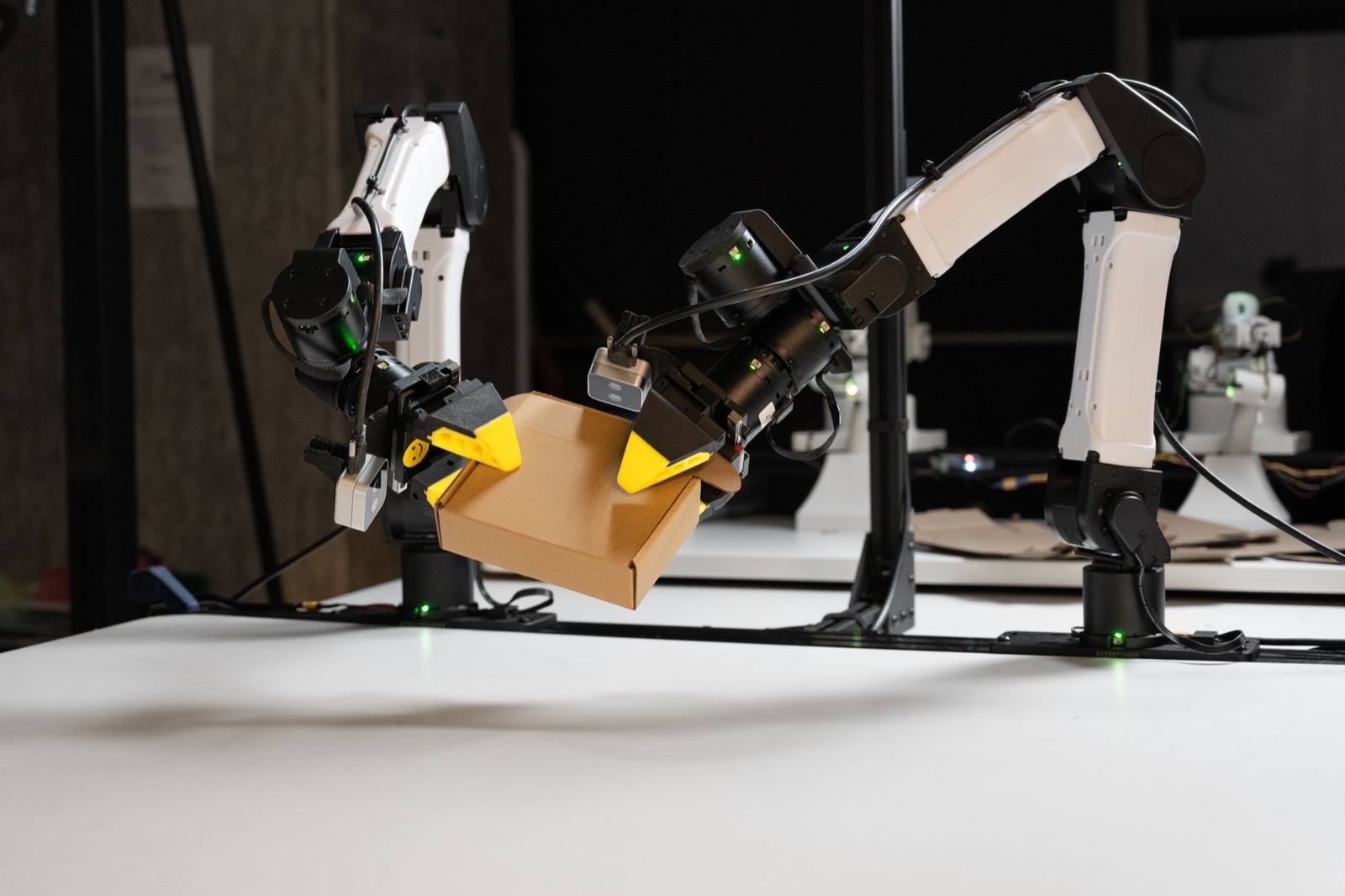} &
        \includegraphics[width=0.247\linewidth]{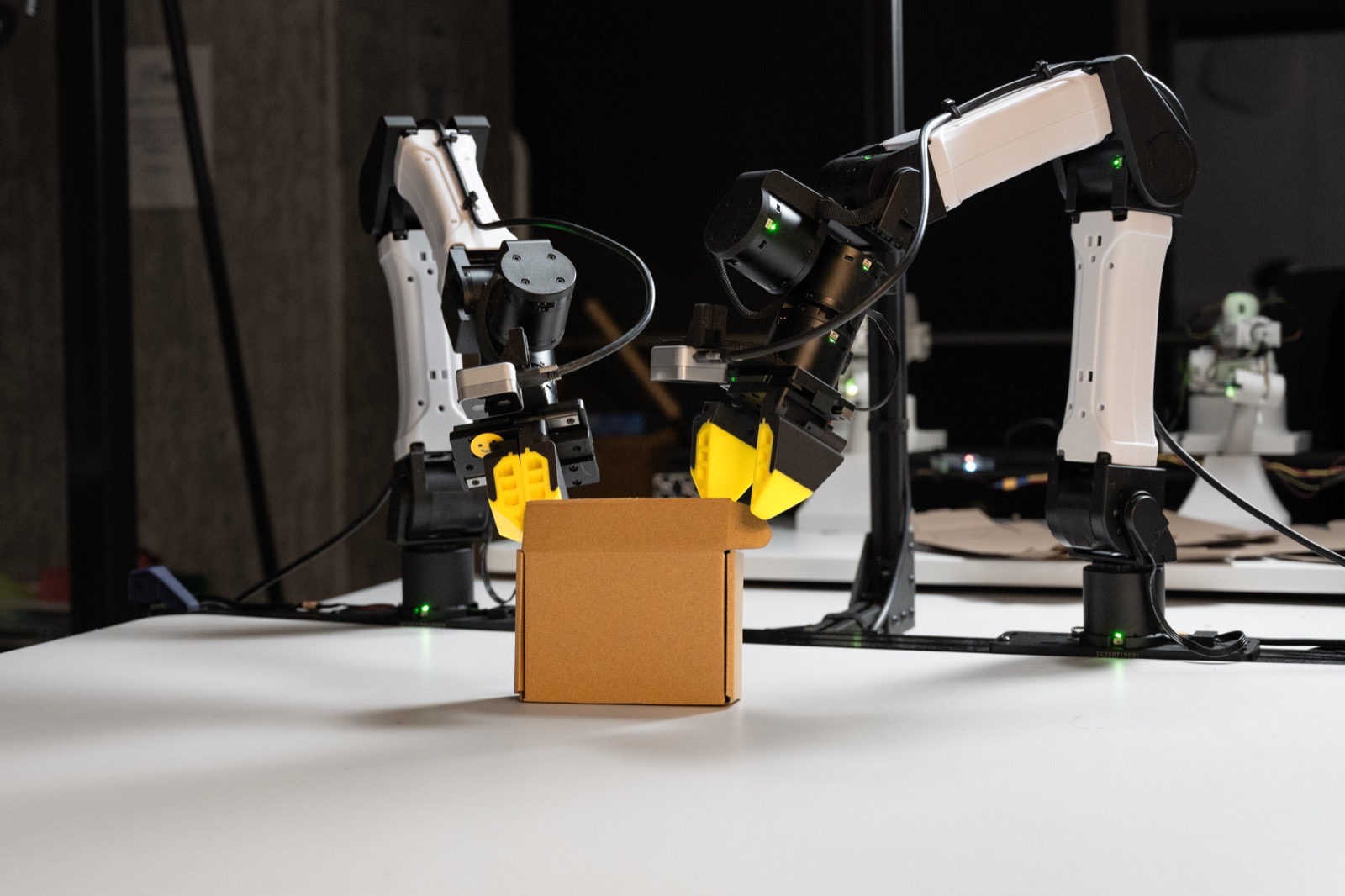} &
        \includegraphics[width=0.247\linewidth]{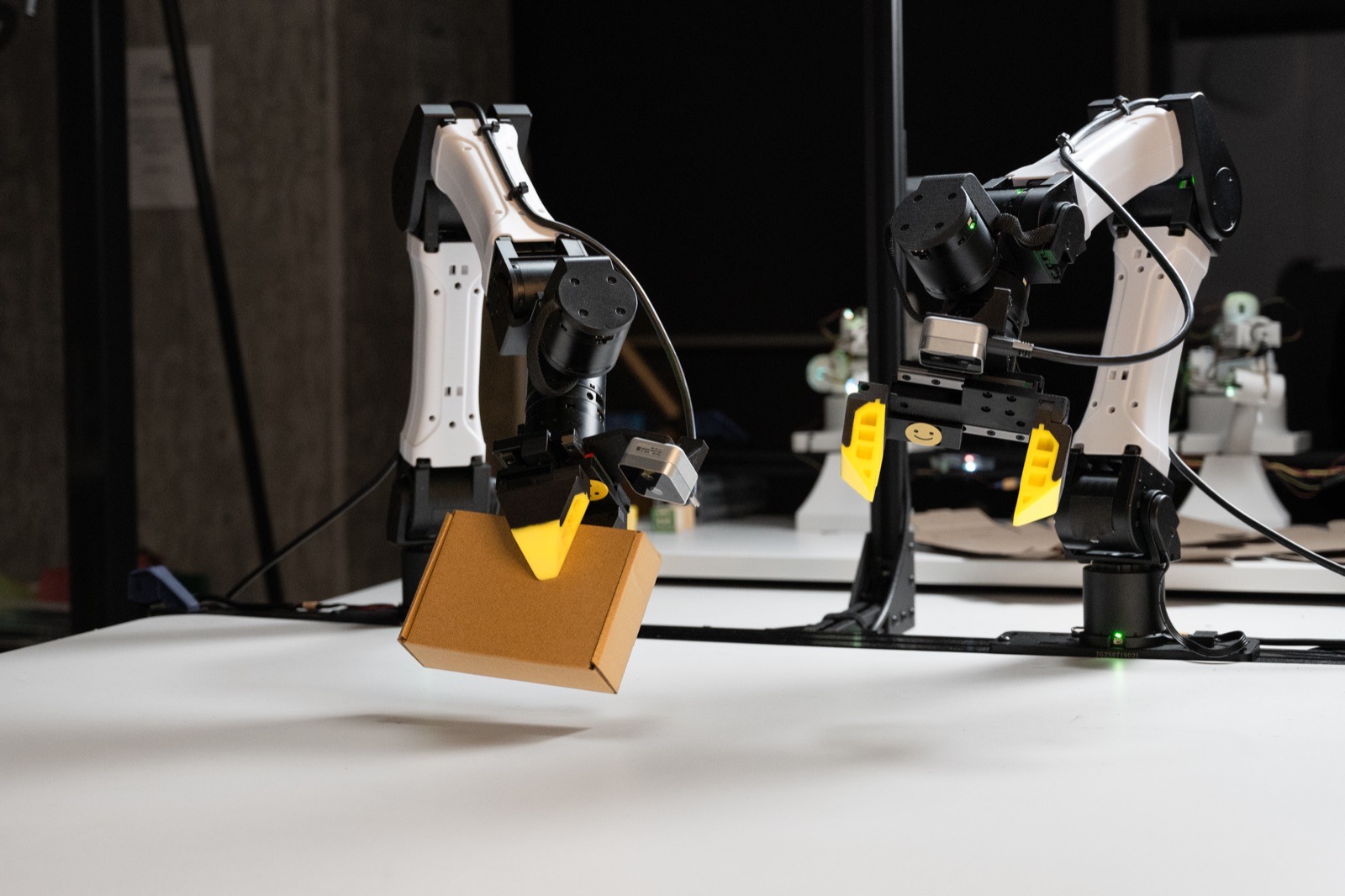} &
        \includegraphics[width=0.247\linewidth]{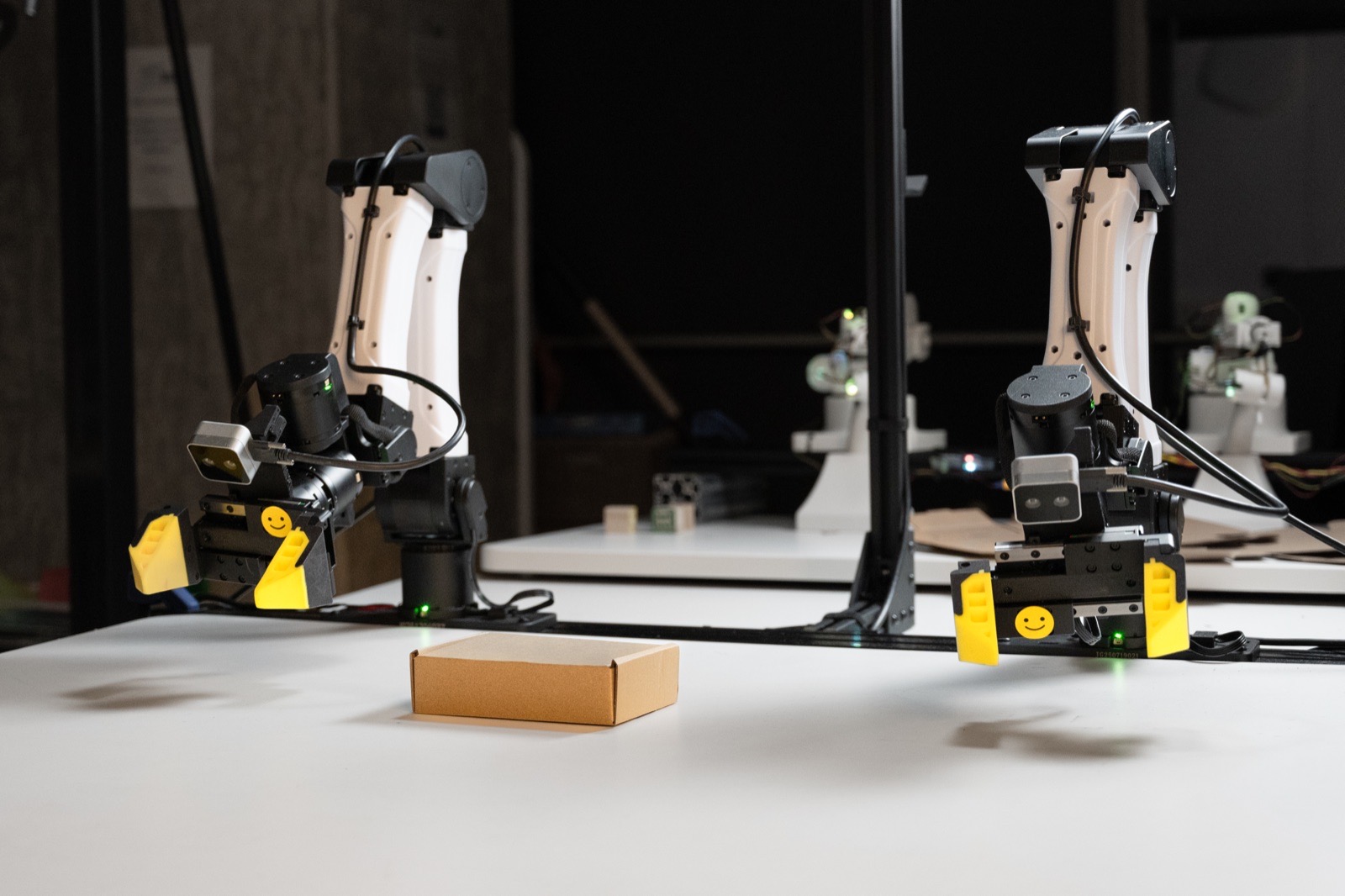}
    \end{tabular}
    \endgroup
    \captionof{figure}{\textbf{The robot autonomously folds a cardboard box and closes the lid.} DAgger intervention data is crucial for achieving good performance on this long-horizon, precise manipulation task.}
    \label{fig:dagger}
\end{center}

To demonstrate the ability of our stack to learn very dexterous tasks, we perform box folding with our policies. We start with the pretrained ABC-DiT model which cannot achieve any real-world success on this dexterous task. Normally, different teleoperators collect demonstrations in different ways, making the data difficult to fit. Due to this, we collect an extra 10 hours of ``collapsed data” which we define as data collected with a relatively stricter standard operating procedure (SOP). We fine-tune the ABC-DiT model on this dataset for 75k steps with a batch size of 720. After training, the model performance is still quite low at only 24\%. While collecting more data under this SOP would likely improve performance, doing so for every dexterous task is prohibitively expensive. One key insight is that the fine-tuned policies exhibit a good understanding of the task, but primarily struggle to make the intermediate adjustments and corrections necessary to complete the task. Instead of collecting data for the entire rollout, we can simply collect recovery behaviors to make these fine-grained adjustments. 

This motivates collecting DAgger intervention data. During collection, we roll out the policy, and intervene when it enters a state where it is struggling and we teleoperate the robot to complete the task. We record data for the whole rollout, not just the human intervention.  Each round of DAgger constitutes roughly 1-1.5 hours of data. During the first round, the ratio of intervention data is around 30\%, which drops down to 15\% during the second round. While the fine-tuning data was collected inside a cage, we opt to collect the DAgger interventions with no cage.

After each round of DAgger, we take the base model from the previous round and continue training for 75k steps. We maintain an 80:10:10 ratio in which 80\% of the data is from the previous round, 10\% is from interventions in the current round, and the other 10\% is from the rest of the episode in the current round. This implicitly overweights interventions while still keeping the rollouts from the whole episode. We find that this ratio is better than training on either whole episodes or interventions alone. This is partly because training on intervention data alone allows the policy to exploit spurious correlations between the source distribution (collected in cages) and the intervention data (collected without a cage), which ultimately worsens the policy. Figure \ref{fig:dagger} shows qualitative results on this task; see the website for videos showcasing the dexterity, and Appendix~\ref{app:dagger_infra} for details of our novel DAgger collection infrastructure.

\section{Related Work}

\subsection{Large-scale open datasets for robotics}

We target the lack of large-scale, dexterous data for low-cost bimanual systems. Early efforts towards data scaling primarily focused on single-arm manipulation. BridgeData-V2~\cite{bridgedatav2} released \textasciitilde 100 hours of robot data across 24 environments on a low-cost WidowX arm. DROID~\cite{droid} provides 350 hours of curated teleoperation data on a single Franka arm across more than 500 scenes, while RH20T~\cite{rh20t} contributes 110k contact-rich trajectories across 41 skills and 5 single-arm platforms. Open X-Embodiment~\cite{openx} aggregates trajectories from multiple embodiments and institutions into a unified source. However, it is a heterogeneous collection with widely varying per-dataset quality, action spaces, and control frequencies.

More recently, the community has shifted focus toward bimanual and dexterous manipulation. The ALOHA ecosystem~\cite{zhao2023learningfinegrainedbimanualmanipulation,fu2024mobilealohalearningbimanual, zhao2024alohaunleashedsimplerecipe} has popularized bimanual teleoperation for contact-rich tasks, though the hardware remains moderately expensive ($\sim$\$20,000) and the released datasets are smaller in scale compared to our collection. Taking a complementary approach to scaling, GR00T N1~\cite{groot_n1} combines 88 hours of real teleoperation data with 780k synthetic trajectories generated via simulation and video models. The MolmoAct2-BimanualYAM Dataset~\cite{fang2026molmoact2actionreasoningmodels} is a bimanual dataset containing 28 unique tasks and 720 hours in total.

In contrast, our dataset contains over 3,500 hours of high-quality, real robot data on a single low-cost bimanual embodiment (our \$8,000 YAM station), spanning 195 tasks with diverse skills and background scenes. The closest dataset to our scale is AgiBot-World~\cite{bu2025agibot_iros}, which features approximately 3,000 hours of real-world bimanual data across roughly 200 tasks. However, AgiBot-World was collected on a substantially more expensive platform (the \$30,000 AgiBot G1). Additionally, while AgiBot-World emphasizes long-horizon macro-manipulation, we also provide data for various fine, dexterous manipulation tasks. We feature tasks such as extracting cards from wallets, folding paper planes, and unlocking doors using a key.

\subsection{Large-scale Behavior Cloning}
Policy architecture and optimization details are critical for training robot policies at scale. While the broad strokes of these designs are publicly available---such as the effectiveness of diffusion objectives for action modeling~\cite{chi2025diffusion} and the benefits of reusing visual features from pre-trained models---the exact implementations vary widely. For instance, the Large Behavior Model (LBM)~\cite{barreiros2025careful} initializes its image backbone with CLIP~\cite{radford2021learning} and uses adaptive layer normalization (adaLN)~\cite{peebles2023scalablediffusionmodelstransformers} to condition its diffusion model. Conversely, other works leverage Vision-Language Model (VLM) backbones: GR00T-N1~\cite{groot_n1} cross-attends to VLM outputs, and $\pi_{0.5}$~\cite{pi05} cross-attends layer-wise to the KV cache of its VLM. %
Despite these high-level insights, finer details of these training recipes are rarely reported, and the underlying training code is largely kept closed-source~\cite{pi0, pi05, barreiros2025careful, brohan2022rt, cheang2024gr}. Because these large-scale efforts are trained and evaluated on proprietary datasets with diverse and often unknown compute budgets, the broader community cannot evaluate these design choices without reproducing deep implementation stacks from scratch.
Our dataset, along with the accompanying infrastructure for large-scale training and evaluation, provides an open-source foundation to answer research questions in manipulation.

\section{Infrastructure}
Beyond the usual data and modeling details, the low-level systems implementations are equally important. To this end, we also release the surrounding infrastructure for this project. Appendix~\ref{app:data_loading} contains the details for efficient data loading from multiple video streams to improve training time. Appendix~\ref{sec:real_robot_setup} details our lightweight and extensible real robot code framework. Appendix~\ref{app:inference_optimization} describes the inference optimizations for both ABC-DiT and ABC-VLA. Appendix~\ref{app:dagger_infra} contains the implementation details for our novel DAgger collection setup with passive leader arms.

\section{Requests for Research}
We present ABC-130K, ABC-Models, and ABC-Sim with the aim of seeding an open-source, reproducible ecosystem for robot learning research. Many interesting open questions remain to be answered for large-scale real-world robot learning, including but not limited to:

\begin{itemize}
    \item \textbf{History conditioning}: policies that condition on a sequence of past observations rather than the current frame alone.
    \item \textbf{Scaling laws for BC}: how performance scales with model and data size for imitation learning.
    \item \textbf{RL finetuning}: improve ABC base models with reinforcement learning beyond the base model.
\end{itemize}
We hope ABC provides a foundation for answering these questions and many others.

\clearpage
\bibliographystyle{unsrtnat}
\bibliography{references}  %

\clearpage

\appendix

\section*{Appendices}
\startcontents[appendices]
\printcontents[appendices]{}{1}[2]{}

\section{ABC-130K Task Taxonomy}
\label{appendix: data}
We organize the 195 tasks in ABC-130K into the following 7 primitive
categories with a goal of grouping the tasks with similar contact modes and precision requirements together. 

\begin{enumerate}

  \item \textbf{Pick-and-Place (67 tasks, 793\,h).} Simple transfers: place cup by coaster, place snacks into paper bag and
  multi-step variants involving packing, organizing, or bussing: pack a student bag, place the mixed dishes into the
  plastic bin. Tasks range from single-object grasps to sequences
  requiring 5--10 objects placed in a target arrangement.

  \item \textbf{Fine Pick-and-Place (39 tasks, 736\,h).} Millimeter-level
  precision tasks: for example, inserting credit cards into holders, building block towers, and pinning/removing brooches, phone case installation, and inserting photos into frames.

    \item \textbf{Folding (36 tasks, 883\,h).} Deformable cloth manipulation spanning
  garment types: t-shirts, long-sleeve shirts, shorts, trousers, skirts,
  tank tops, towels, as well as non-garment deformables such as paper
  boxes and paper planes.
  \item \textbf{Insertion/Ejection (19 tasks, 441\,h).} Peg-in-hole-style skills generalized to everyday objects: plugs into sockets, pens into caps,
  batteries into remote controls, earbuds into charging cases, and keys
  onto keyrings and caps onto bottles.

  \item \textbf{Tool Use (16 tasks, 321\,h).} Tasks requiring an intermediate tool or mechanism: zipping jackets, sealing zip-top bags, locking and
  unlocking with keys, and sweeping with a dustpan.

  \item \textbf{Sorting (8 tasks, 205\,h).} Multi-object classification tasks:
  sorting LEGOs by color, pills into containers, screws and nuts by
  type, and chemistry lab equipment.

  \item \textbf{Tying/Untying (10 tasks, 175\,h).} Knot-tying on plastic bags,
  shoe-lacing, tying bouquets of flowers, and cable/headphone wrapping
  and untangling---long, thin deformables requiring bimanual
  coordination and  contact-rich manipulation.
\end{enumerate}

\begin{figure*}[!t]
    \centering
    \newlength{\tcw}\setlength{\tcw}{0.188\textwidth}
    \newcommand{\labelimg}[2]{%
        \begin{tikzpicture}[inner sep=0pt, outer sep=0pt]
            \node[anchor=south west] (img) at (0,0)
                {\includegraphics[width=\tcw]{#1}};
            \useasboundingbox (img.south west) rectangle (img.north east);
            \node[anchor=north west, font=\sffamily\tiny, text=black,
                  fill=white, fill opacity=0.78, text opacity=1,
                  inner xsep=2pt, inner ysep=1pt, rounded corners=1pt]
                at ([xshift=3pt, yshift=-3pt]img.north west) {#2};
        \end{tikzpicture}}
    \newcommand{\labelimgsq}[2]{%
        \begin{tikzpicture}[inner sep=0pt, outer sep=0pt]
            \node[anchor=south west] (img) at (0,0)
                {\includegraphics[width=\tcw, height=0.75\tcw]{#1}};
            \useasboundingbox (img.south west) rectangle (img.north east);
            \node[anchor=north west, font=\sffamily\tiny, text=black,
                  fill=white, fill opacity=0.78, text opacity=1,
                  inner xsep=2pt, inner ysep=1pt, rounded corners=1pt]
                at ([xshift=3pt, yshift=-3pt]img.north west) {#2};
        \end{tikzpicture}}
    \setlength{\tabcolsep}{2pt}
    \renewcommand{\arraystretch}{1.15}
    \begin{tabular}{@{}ccccc@{}}
        \labelimg{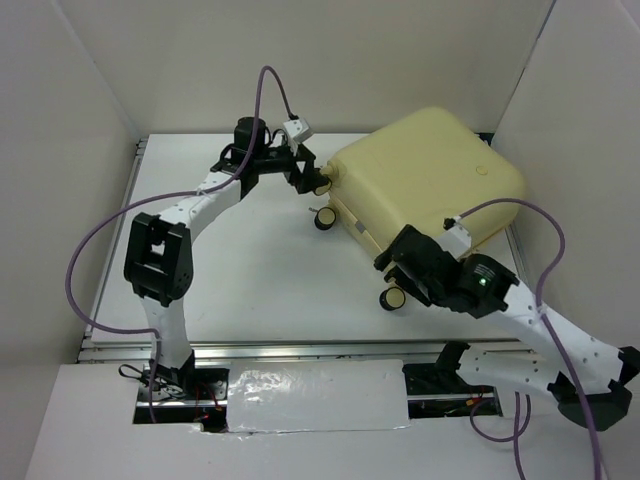}{Packing luggage} &
        \labelimg{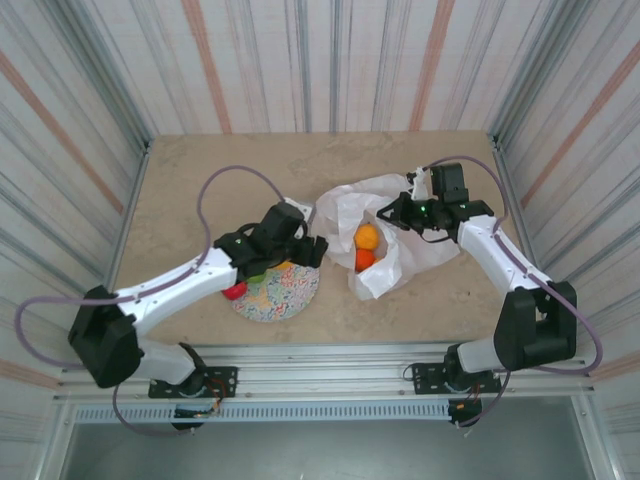}{Place fruits in bag} &
        \labelimg{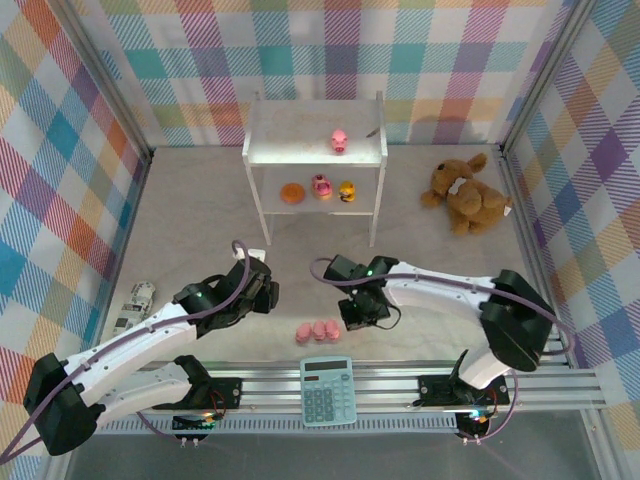}{Organize toys on shelf} &
        \labelimg{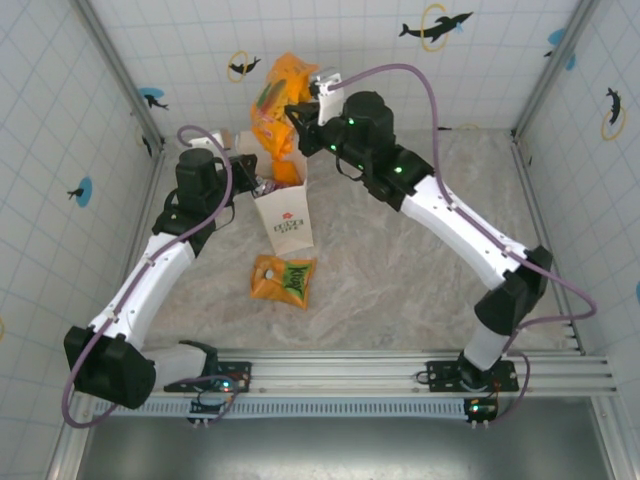}{Place snacks in bag} &
        \labelimg{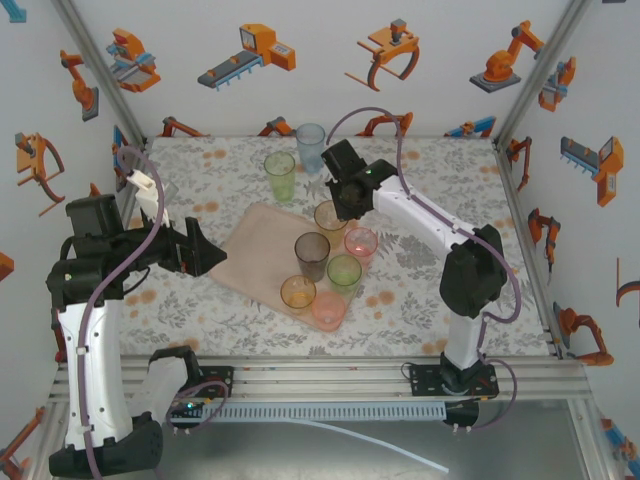}{Place glasses in tray} \\
        \labelimg{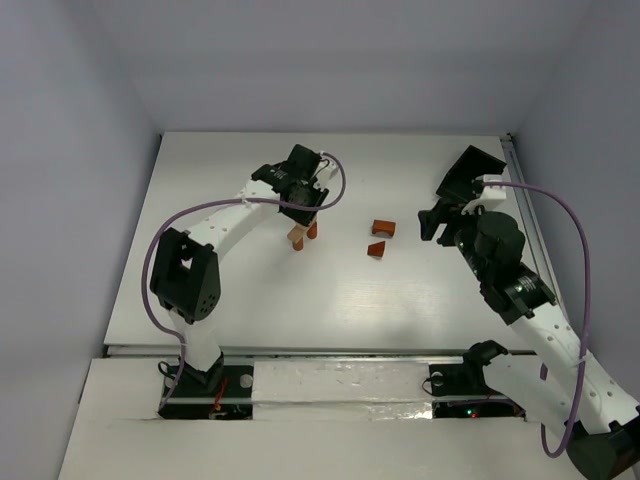}{Build block tower} &
        \labelimg{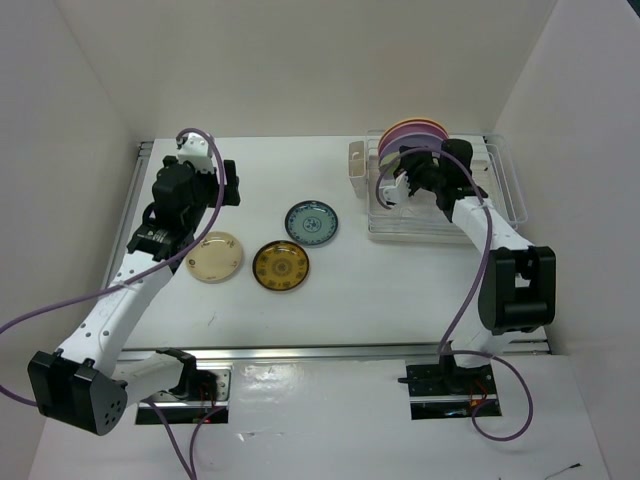}{Load dishrack} &
        \labelimg{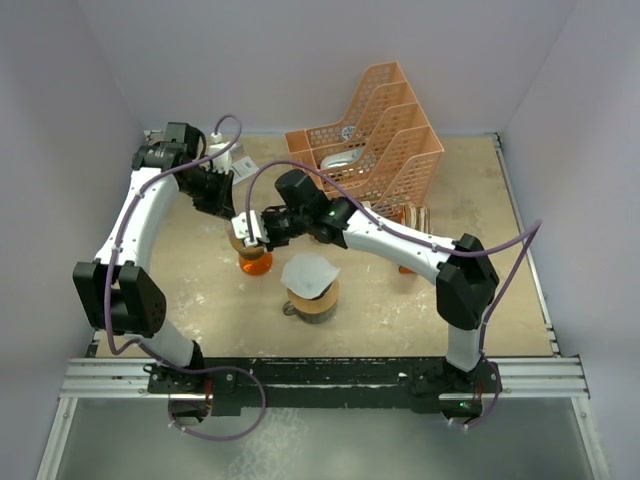}{Place coffee filter} &
        \labelimgsq{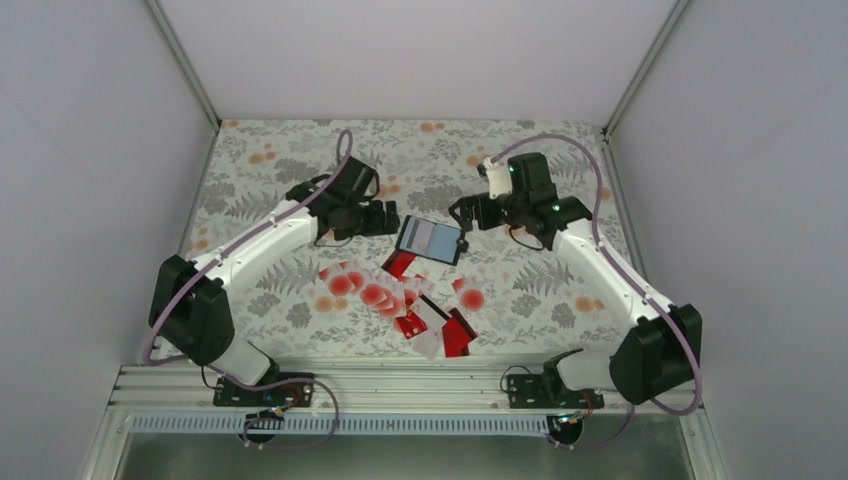}{Insert credit cards} &
        \labelimg{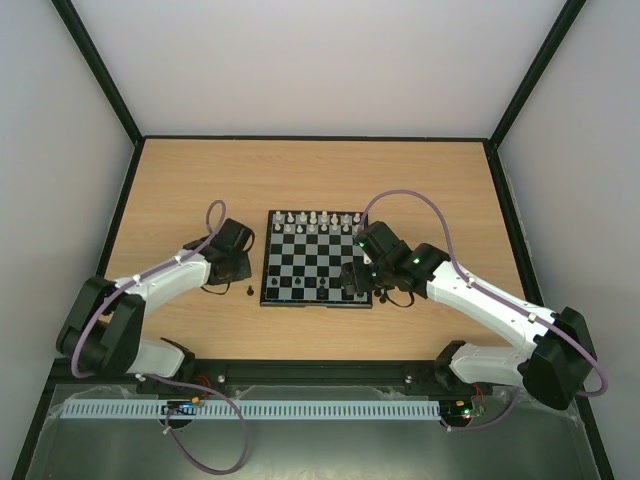}{Set up chess pieces} \\
        \labelimg{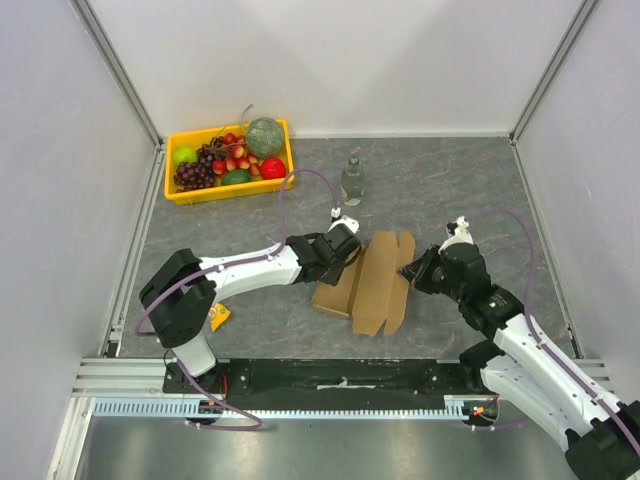}{Fold paper box} &
        \labelimg{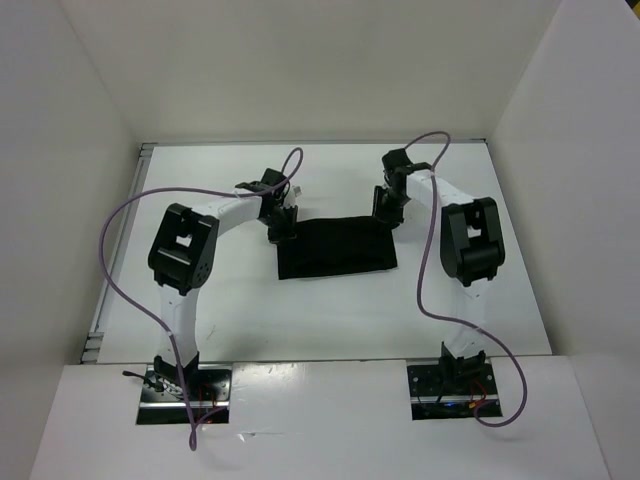}{Fold skirt pile} &
        \labelimg{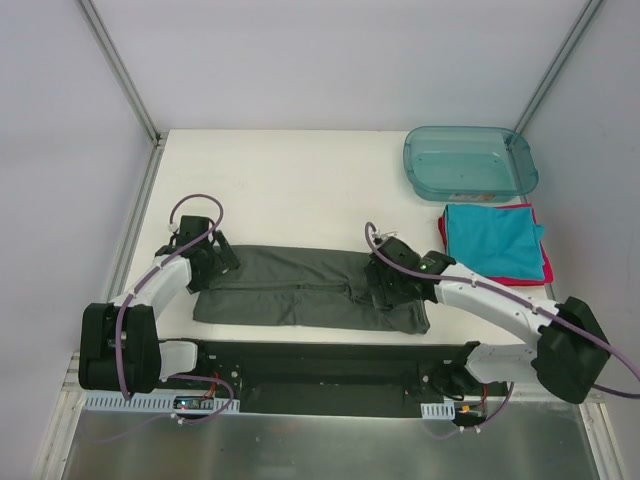}{Fold t-shirt pile} &
        \labelimg{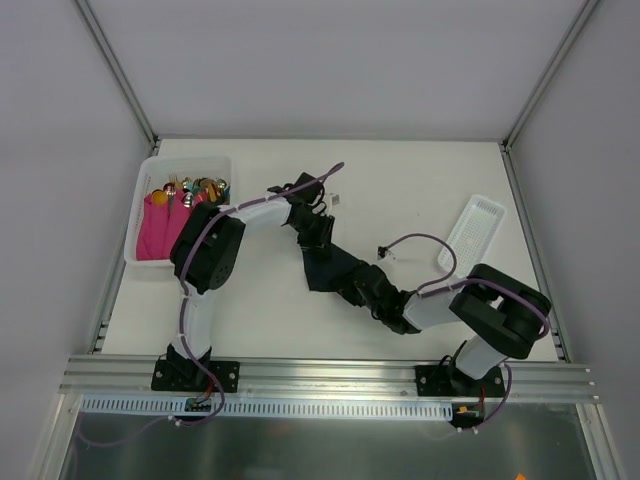}{Roll up utensils} &
        \labelimgsq{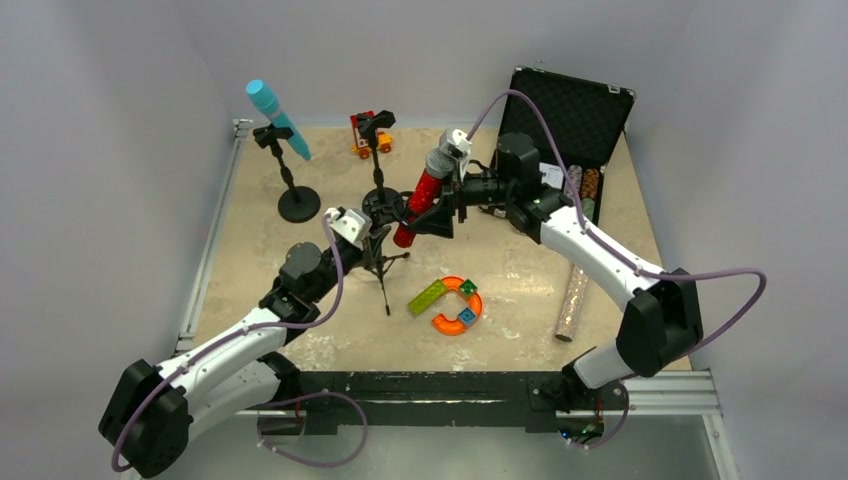}{Attach microphone} \\
        \labelimg{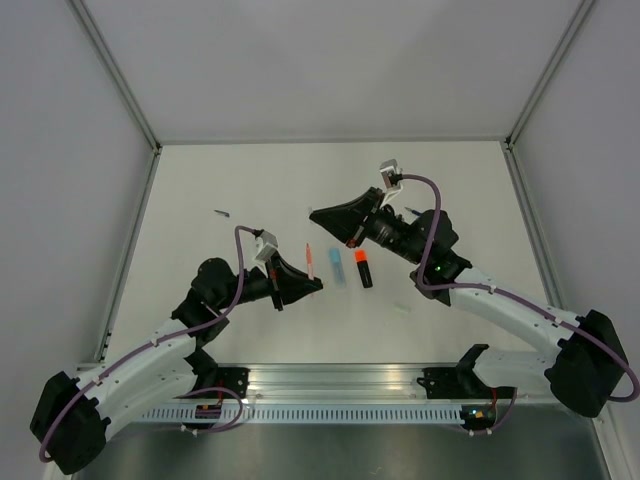}{Cap pens} &
        \labelimg{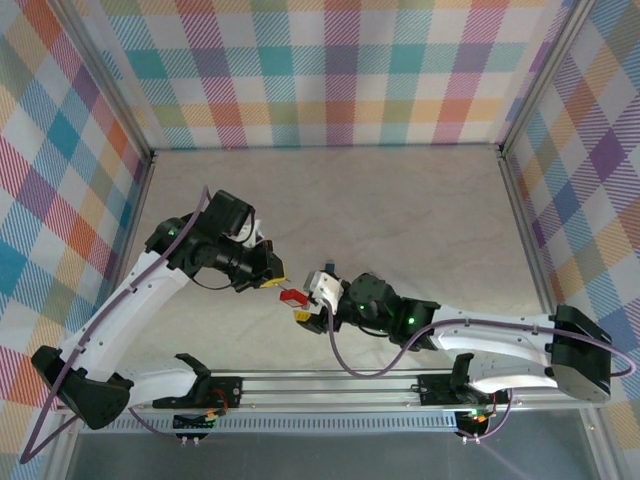}{Lock with key} &
        \labelimg{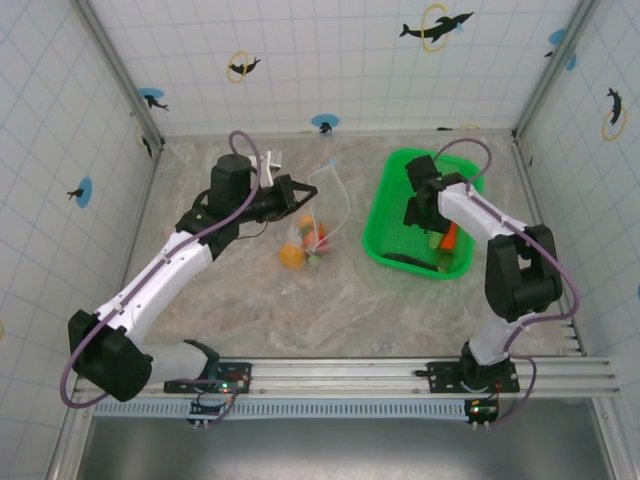}{Seal zip-top bag} &
        \labelimg{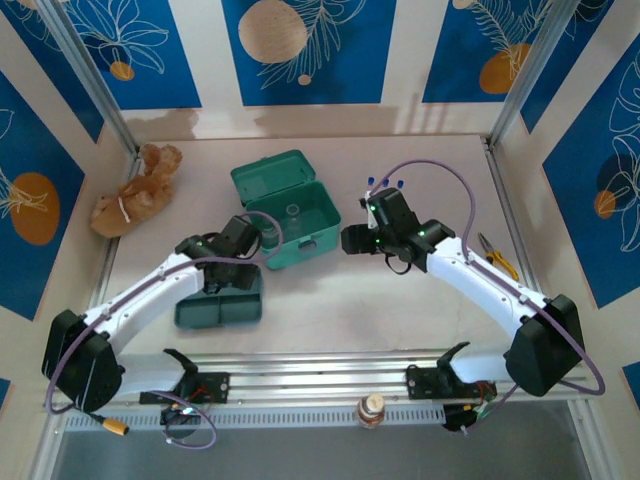}{Organize lab equipment} &
        \labelimg{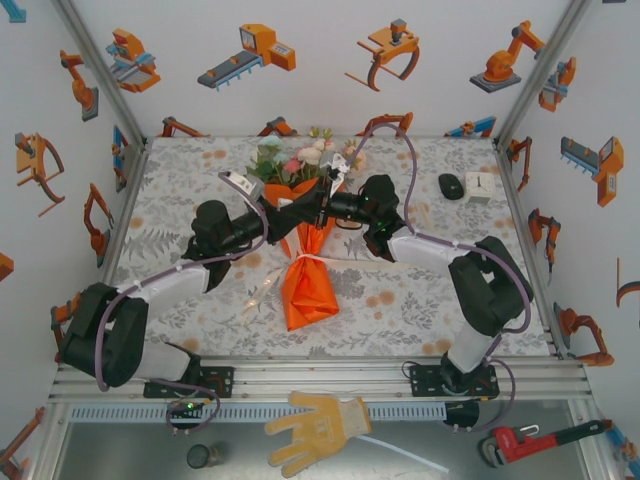}{Tie flower bouquet} \\
    \end{tabular}
    \caption{\textbf{Representative top-camera frames for each task category.} \textbf{Pick-and-Place} (row 1, all 5): \textit{Packing luggage}, \textit{Place fruits in bag}, \textit{Organize toys on shelf}, \textit{Place snacks in bag}, \textit{Place glasses in tray}. \textbf{Fine Pick-and-Place} (row 2, all 5): \textit{Build block tower}, \textit{Load dishrack}, \textit{Place coffee filter}, \textit{Insert credit cards}, \textit{Set up chess pieces}. \textbf{Folding} (row 3, cells 1--4): \textit{Fold paper box}, \textit{Fold skirt pile}, \textit{Fold t-shirt pile}, \textit{Roll up utensils}. \textbf{Insertion/Ejection} (row 3 cell 5, row 4 cell 1): \textit{Attach microphone}, \textit{Cap pens}. \textbf{Tool Use} (row 4, cells 2--3): \textit{Lock with key}, \textit{Seal zip-top bag}. \textbf{Sorting} (row 4 cell 4): \textit{Organize lab equipment}. \textbf{Tying/Untying} (row 4 cell 5): \textit{Tie flower bouquet}.}
    \label{fig:appendix_category_frames}
\end{figure*}

Examples of top-camera frames from each task category are provided in Figure~\ref{fig:appendix_category_frames}. Task names in the dataset are often minor grammatical variants of each other. For example, \textit{sorting legos into containers} and \textit{sort the legos into the containers} are two distinct task descriptions in the dataset that have the same qualitative behavior. We adjust for this duplication in the provided statistics by counting such variants as one task.

\section{Training \& Model Details}
\label{app:training-details}
This appendix provides the implementation details for reproducing our results. We first describe \texttt{abcdl}, our custom dataloader, together with the MP4-plus-binary episode format used to load thousands of hours of video efficiently in distributed training. We then provide the training and architecture details for ABC-DiT and ABC-VLA. Finally, we report the model sizes and FLOP counts. All models are trained on NVIDIA H200 GPUs.

\subsection{Data Loading}
\label{app:data_loading}

\definecolor{MoovBlue}{HTML}{2F6F9F}
\definecolor{NeverGray}{HTML}{ECECEC}
\definecolor{EdgeGray}{HTML}{9A9A9A}
\colorlet{ScanRed}{PrettyMultTwo}
\colorlet{GopGreen}{NPGBlueLight}
\colorlet{InkCol}{PrettyText}

\begin{figure}[!t]
    \centering
    \resizebox{\linewidth}{!}{%
    \begin{tikzpicture}[x=0.92cm, y=0.92cm, font=\rmfamily]
      \def\W{12.605}%
      \def\bh{1.0}%
      \fill[NeverGray, draw=EdgeGray, line width=0.4pt] (0,1.65) rectangle (\W,{1.65+\bh});
      \fill[ScanRed] (0,1.65) rectangle (\W,{1.65+\bh});
      \node[anchor=east, align=right, InkCol] at (-0.45,{1.65+\bh/2}) {naive};
      \fill[NeverGray, draw=EdgeGray, line width=0.4pt] (0,0) rectangle (\W,\bh);
      \fill[MoovBlue] (0.0,0) rectangle (0.13,\bh);     %
      \fill[GopGreen] (2.645,0) rectangle (2.775,\bh);  %
      \node[anchor=east, align=right, InkCol] at (-0.45,{\bh/2}) {after constant\\keyframe encoding};
      \draw[InkCol, line width=0.6pt] (0,{1.65+\bh+0.45}) -- ({\W+0.3},{1.65+\bh+0.45});
      \foreach \x in {0,2,4,6,8,10,12}
        \draw[InkCol] (\x,{1.65+\bh+0.45}) -- (\x,{1.65+\bh+0.65}) node[above, InkCol] {\x};
      \node[InkCol] at ({\W/2},{1.65+\bh+1.25}) {byte offset (MB)};
      \node[anchor=west, draw=GopGreen, line width=1.2pt, inner sep=0pt]
        at ({\W+0.7},1.7) {\includegraphics[height=2.8cm]{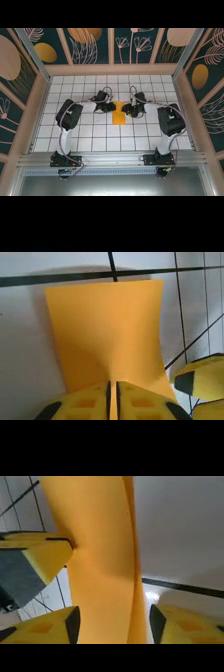}};
      \begin{scope}[shift={(1.6,-1.35)}]
        \fill[ScanRed]  (0,0.45) rectangle (0.35,0.8);   \node[anchor=west, InkCol] at (0.45,0.625) {whole-file index scan};
        \fill[GopGreen] (5.2,0.45) rectangle (5.55,0.8); \node[anchor=west, InkCol] at (5.65,0.625) {one GOP decoded};
        \fill[MoovBlue] (0,0) rectangle (0.35,0.35);     \node[anchor=west, InkCol] at (0.45,0.175) {moov / frame index};
        \fill[NeverGray, draw=EdgeGray] (5.2,0) rectangle (5.55,0.35); \node[anchor=west, InkCol] at (5.65,0.175) {never read};
      \end{scope}
    \end{tikzpicture}%
    }
    \caption{\textbf{Efficient frame access for fast data loading.} Encoding keyframes more frequently and in a manner that allows for an analytically reconstructed frame index makes random frame access nearly free. To read one frame from a video, \texttt{torchcodec}'s default scans the entire file to build its frame index (top). Correctly encoding the file allows us to compute the index analytically, meaning we only need to read the file header plus frames since the closest keyframe, leaving the rest of the file untouched, reducing disk pressure. The decoded frame is three vertically stacked camera views.}
    \label{fig:dataloader-random-access}
\end{figure}

  \begin{figure*}[t]
      \centering
      \begin{minipage}[t]{0.49\linewidth}
          \centering
          \resizebox{\linewidth}{!}{%
              \begin{tikzpicture}
\begin{axis}[
  width=86mm, height=60mm,
  xmode=log, log basis x=2,
  xmin=0.85, xmax=150,
  xtick={1,8,16,32,64,128},
  xticklabels={1,8,16,32,64,128},
  ymin=0, ymax=500,
  ytick={0,100,200,300,400,500},
  xlabel={Dataloader workers},
  ylabel={Decodes / sec},
  grid=major,
  axis line style={PrettyText},
  tick style={PrettyText},
  tick label style={font=\ttfamily\scriptsize, PrettyText},
  label style={font=\small, PrettyText},
  legend style={
    at={(0.03,0.97)}, anchor=north west,
    legend columns=1, draw=gray!35, fill=white, fill opacity=0.92,
    text opacity=1.0, font=\scriptsize, text=PrettyText,
    legend cell align=left, row sep=1pt,
  },
]
\addplot+[color=VLAColor, mark=*,        mark size=1.8pt, line width=1.0pt, mark options={fill=VLAColor}]
  table [x=workers, y=naive, col sep=tab] {figures/pgfplots/data/dataloader_fs_throughput.tsv};
\addlegendentry{naive (GOP 250, no CFR)}
\addplot+[color=LBMColor, mark=diamond*, mark size=2.2pt, line width=1.0pt, mark options={fill=LBMColor}]
  table [x=workers, y=nonnaive, col sep=tab] {figures/pgfplots/data/dataloader_fs_throughput.tsv};
\addlegendentry{non-naive (GOP 30 + CFR)}
\end{axis}
\end{tikzpicture}%
          }
          \captionof{figure}{\textbf{Random-frame decode throughput.}
          Decodes per second versus dataloader workers on a local filesystem. Fixing the encoding options and dataloader args improves throughput.}
          \label{fig:dataloader-throughput}
      \end{minipage}\hfill
      \begin{minipage}[t]{0.49\linewidth}
          \centering
          \resizebox{\linewidth}{!}{%
              \begin{tikzpicture}
\begin{axis}[
  width=86mm, height=60mm,
  ybar, bar width=16pt,
  bar shift=0pt,
  enlarge x limits={abs=28pt},
  ymode=log,
  log origin=infty,
  ymin=0.07, ymax=45,
  ytick={0.1,1,10},
  yticklabels={0.1,1,10},
  ylabel={MB per decode (log)},
  symbolic x coords={naive,nonnaive},
  xtick={naive,nonnaive},
  xticklabels={
    {naive\\(GOP 250, no CFR)},
    {non-naive\\(GOP 30 + CFR)}
  },
  x tick label style={font=\scriptsize, PrettyText, align=center, text width=24mm},
  tick pos=left,
  axis line style={PrettyText},
  tick style={PrettyText},
  tick label style={font=\ttfamily\scriptsize, PrettyText},
  label style={font=\small, PrettyText},
  major grid style={gray!35},
  ymajorgrids,
  nodes near coords,
  point meta=explicit symbolic,
  every node near coord/.append style={font=\scriptsize, PrettyText, anchor=south},
]
\addplot+[fill=VLAColor, draw=PrettyText, line width=0.4pt] coordinates {(naive,9.7473) [9.75]};
\addplot+[fill=LBMColor, draw=PrettyText, line width=0.4pt] coordinates {(nonnaive,0.1404) [0.14]};
\end{axis}
\end{tikzpicture}%
          }
          \captionof{figure}{\textbf{Per-decode read volume.}
          Mean MB requested from the filesystem per decode (log scale). CFR
          frame mapping cuts per-decode reads by ${\sim}70\times$
          (\SI{9.75}{MB}~$\to$~\SI{0.14}{MB}) on average, reducing wall-time on bandwidth-constrained dataloading.}
          \label{fig:dataloader-bytes}
      \end{minipage}
  \end{figure*}

We develop a custom dataloader, \texttt{abcdl} (A Behavior Cloning Dataloader) to facilitate fast loading of training data in our pipeline. Many data loading pipelines do not scale to loading thousands of hours of data in a distributed training setup; hence part of making ABC-130K useful is releasing a dataloader along with our dataset that can facilitate distributed dataloading at such a scale. 

The training data format consists of two files per episode: 1) an MP4 containing the concatenated camera views, and 2) a binary file containing the states and actions.
Our data loading philosophy is to keep our data in an extremely simple format while still facilitating extremely fast on the fly loading of data.  From this perspective, MP4 provides great advantages as it is readable by all manner of video players, IDEs, etc., for convenient viewing without special software, while still being a highly compressed format and providing many format options to facilitate fast dataloading.
Our system facilitates either loading quickly from a locally mounted file system or streaming data from cloud object storage such as S3, simply by changing the backend. In the case of local data loading, files are sharded by episode and downloaded onto individual nodes.
The videos are encoded using the \texttt{H.264} format. We use several format options to make decoding easy:%

\begin{itemize}
    \item \texttt{+faststart} moves the MP4 \texttt{moov} atom to the front of the file meaning that the whole file doesn't need to be read before decoding can begin.
    \item We disable MP4 B-frames so that frames only depend on the most recent keyframe, and not on frames after them.
    \item Crucially, we encode the video with constant GOP (Group of Pictures, i.e. a keyframe + following frames compressed relative to it). This means that the keyframes in the video are in deterministic positions.
\end{itemize}

In practice, we encode videos with \texttt{GOP=30}, corresponding to one keyframe per second for 30-FPS video. A fixed \texttt{GOP} and constant ticks per frame allow us to reconstruct the file index. The reconstructed index is passed to \texttt{torchcodec}'s \texttt{VideoDecoder} object in order to facilitate fast loading (see Figure \ref{fig:dataloader-random-access}). We note that all of these speedups are achieved by using existing APIs in pure Python, keeping our system readable and maintainable. For remote data loading, we make a custom Python file-like object which facilitates reading bytes from the given remote storage source (e.g. \texttt{S3} or GCS), and is passed to \texttt{torchcodec}.

To handle multiple camera streams, we simply stack them into a single MP4 file. This cuts down the number of requests for multiple camera streams by a factor of 3 (assuming 3 cameras, more in the general case). This is important especially when streaming from a remote data source. We also support loading multiple consecutive frames; this can be done at minimal bandwidth / CPU decoding cost as the keyframe for subsequent frames is more likely than not already loaded. 

While providing speed at the low level, our dataloading infrastructure facilitates the ability for practitioners to freely mix datasets from different sources and weight them independently. This is an essential feature for many experiments -- for example for experimenting with different slices of pretraining mixes at the pretraining scale, or for experiments with intervention data at the single task scale.

\subsection{Architecture Details}
\label{app:arch-details}
ABC-DiT, the DINOv3-xattn version in Table~\ref{tab:architecture_ablations}, uses DINOv3~\cite{simeoni2025dinov3} ViT-B. We use 12 query tokens per image and share the weights of the DINO encoder across all three images. We use a modified DiT architecture with 32 layers, a hidden size of 1536, 24 heads, and an MLP ratio of 4.

ABC-VLA, the Pooled adaLN version in Table~\ref{tab:architecture_ablations}, uses a Gemma 3~\cite{kamath2025gemma} 4B backbone. The images are fed into the context through the SigLIP encoder. The robot proprioception is fed through a two-layer MLP with hidden size 256 to produce a 2560-dimensional vector that is fed into the token stream, followed by the task prompt and a \texttt{<action\_start>} token. We use 8 learnable query tokens to attention-pool the tokens of the last layer of the Gemma VLM. This produces 8 512-dimensional vectors that are flattened. The proprioception is also linearly projected to a 512-dimensional embedding. The 4096-dimensional vector corresponding to the VLM features is concatenated with this proprioception embedding as well as a 512-dimensional embedding of the diffusion timestep. This combined 5120-dimensional vector is then fed through a small MLP to be projected down to a 512-dimensional vector and is used for adaLN modulation in the diffusion head. The diffusion head is relatively small with 8 layers, hidden size 512, 8 heads, and an MLP ratio of 4.

\begin{figure}[!t]
    \centering
    \includegraphics[width=\linewidth]{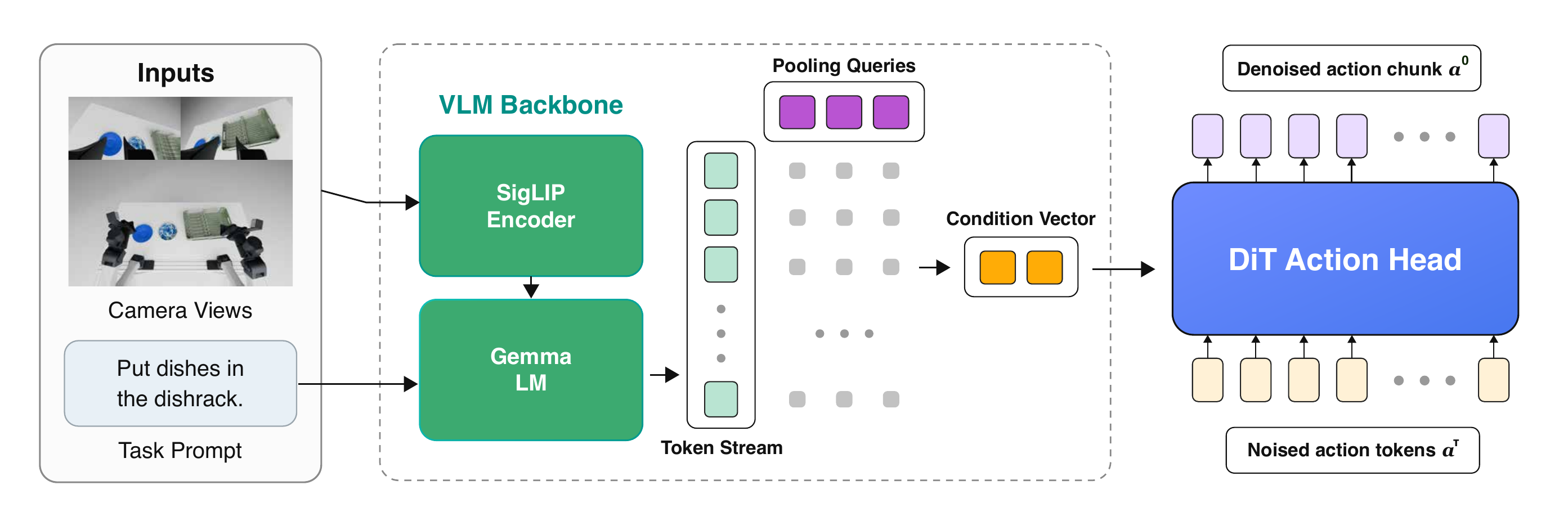}
    \caption{\textbf{ABC-VLA pooled adaLN architecture.} ABC-VLA uses a Gemma~3 VLM backbone, attention-pools the final VLM hidden states into a compact feature vector, and projects the result to an adaLN conditioning vector for the lightweight DiT action head.}
    \label{fig:abc-vla-adaln-architecture}
\end{figure}
\subsection{Training Details}
\label{app:training_details}
For ABC-DiT, we linearly warm up the learning rate for the first 1000 iterations to $1 \times 10^{-4}$ and then keep it constant throughout training. We use the AdamW optimizer~\cite{loshchilov2019decoupledweightdecayregularization} with a weight decay of 0.01. We use a learning rate scale of 0.1 for the vision encoder. We use gradient clipping with a maximum gradient norm of 10. We train on 12 H200 nodes with a batch size of 96 per GPU bringing the global batch size to 9216. We drop out the input proprioception with a probability of 0.1.

For ABC-VLA, we linearly warm up the learning rate for the first 1000 steps to $1 \times 10^{-4}$ and keep it constant throughout training. We use the AdamW optimizer~\cite{loshchilov2019decoupledweightdecayregularization} with weight decay 0.01. We use gradient clipping with max gradient norm 10 and maximum gradient value 100. The model is trained with bf16 mixed precision. We train on 8 H200 nodes with batch size 24 per GPU and gradient accumulation 6, bringing the global batch size to 9216. We drop out the input proprioception with probability 0.1.

\subsection{Architecture Sizes and FLOPS}
\label{app:arch_scaling}

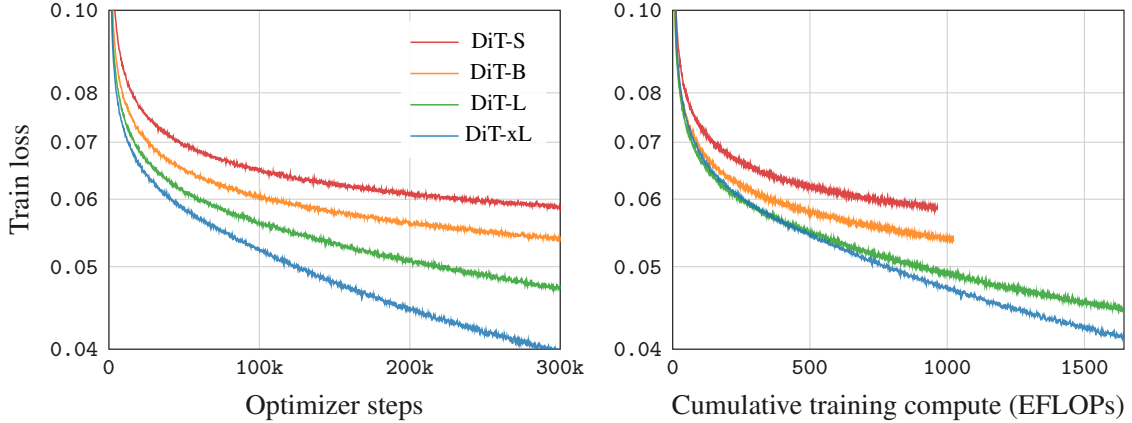
\begin{figure*}[t]
    \centering
    \resizebox{0.95\linewidth}{!}{%
        \begin{tikzpicture}
\begin{groupplot}[
  group style={
    group size=2 by 1,
    horizontal sep=14mm,
    ylabels at=edge left,
  },
  width=72mm, height=58mm,
  ymode=log,
  ymin=0.040, ymax=0.10,
  ytick={0.04,0.05,0.06,0.07,0.08,0.10},
  yticklabels={0.04,0.05,0.06,0.07,0.08,0.10},
  grid=major,
  axis line style={PrettyText},
  tick label style={font=\ttfamily\scriptsize, PrettyText},
  label style={font=\small, PrettyText},
  legend style={
    draw=none,
    fill=white,
    fill opacity=0.92,
    text opacity=1,
    at={(0.97,0.97)},
    anchor=north east,
    font=\scriptsize,
  },
]
\nextgroupplot[
  xlabel={Optimizer steps},
  ylabel={Train loss},
  xmin=0, xmax=300000,
  xtick={0,100000,200000,300000},
  xticklabels={0,100k,200k,300k},
  scaled x ticks=false,
]
\addplot+[color=DiTSColor,  no marks, line width=0.55pt, opacity=0.85]
  table [x=step, y=loss_smooth] {figures/pgfplots/data/dit_scaling_loss_dit_S.tsv};
\addlegendentry{DiT-S}
\addplot+[color=DiTBColor,  no marks, line width=0.55pt, opacity=0.85]
  table [x=step, y=loss_smooth] {figures/pgfplots/data/dit_scaling_loss_dit_B.tsv};
\addlegendentry{DiT-B}
\addplot+[color=DiTLColor,  no marks, line width=0.55pt, opacity=0.85]
  table [x=step, y=loss_smooth] {figures/pgfplots/data/dit_scaling_loss_dit_L.tsv};
\addlegendentry{DiT-L}
\addplot+[color=DiTxLColor, no marks, line width=0.55pt, opacity=0.85]
  table [x=step, y=loss_smooth] {figures/pgfplots/data/dit_scaling_loss_dit_xL_steps.tsv};
\addlegendentry{DiT-xL}

\nextgroupplot[
  xlabel={Cumulative training compute (EFLOPs)},
  xmin=0, xmax=1644,
  xtick={0,500,1000,1500},
  xticklabels={0,500,1000,1500},
  scaled x ticks=false,
]
\addplot+[color=DiTSColor,  no marks, line width=0.55pt, opacity=0.85]
  table [x=compute_eflops, y=loss_smooth] {figures/pgfplots/data/dit_scaling_loss_dit_S.tsv};
\addplot+[color=DiTBColor,  no marks, line width=0.55pt, opacity=0.85]
  table [x=compute_eflops, y=loss_smooth] {figures/pgfplots/data/dit_scaling_loss_dit_B.tsv};
\addplot+[color=DiTLColor,  no marks, line width=0.55pt, opacity=0.85]
  table [x=compute_eflops, y=loss_smooth] {figures/pgfplots/data/dit_scaling_loss_dit_L.tsv};
\addplot+[color=DiTxLColor, no marks, line width=0.55pt, opacity=0.85]
  table [x=compute_eflops, y=loss_smooth] {figures/pgfplots/data/dit_scaling_loss_dit_xL.tsv};
\end{groupplot}
\end{tikzpicture}%
    }
    \caption{\textbf{ABC-DiT model-size scaling.} Train loss versus optimizer steps (left) and versus cumulative training compute (right) for four ABC-DiT sizes trained with identical hyperparameters and global batch size $9{,}216$. At a fixed compute or step budget the larger DiTs reach lower train loss. The total number of parameters in S/B/L/xL are 153M, 290M, 746M, and 1.93B respectively.}
    \label{fig:dit-scaling-steps-vs-flops}
\end{figure*}

Our choices of model size are relatively nonstandard: for ABC-DiT, we use a large action head while keeping the DINOv3 visual backbone comparatively small. This is because the visual backbone is parameter-efficient but compute-intensive: it attends over image patches for three camera views, so its FLOP cost is large relative to its parameter count. This makes it comparatively cheap to add parameters in the action head. Empirically, Figure \ref{fig:dit-scaling-steps-vs-flops} shows that scaling the size of the DiT head is efficient in reducing training loss in terms of both training steps and FLOPs.

Conversely, ABC-VLA places most of its parameters and FLOPs in the Gemma-SigLIP backbone, while its diffusion action head is small. Scaling up the diffusion head in this case would appear to be less effective than simply doing the multiple diffusion draws trick covered in the main paper (i.e. taking multiple diffusion noise + timestep samples and averaging the gradients into the backbone) as the bottleneck is the variance of the diffusion gradients into the VLM rather than the absolute number of parameters.
Table~\ref{tab:flops-param-breakdown} summarizes this split.

\begin{table}[t]
\centering
\small
\setlength{\tabcolsep}{3pt}
\begin{tabular}{lcc|ccc}
\toprule
Model &
\multicolumn{2}{c}{Parameters} &
\multicolumn{3}{c}{Training TFLOPs / sample} \\
\cmidrule(lr){2-3}\cmidrule(lr){4-6}
 & Backbone & Action head & Backbone & Action head & Total \\
\midrule
ABC-DiT &
\cellcolor{VLAColor!7}85.7M &
\cellcolor{VLAColor!18}1.93B &
\cellcolor{LBMColor!12}0.329 &
\cellcolor{LBMColor!12}0.349 &
\cellcolor{LBMColor!12}0.678 \\
ABC-VLA &
\cellcolor{VLAColor!18}4.3B &
\cellcolor{VLAColor!7}44.7M &
\cellcolor{LBMColor!18}6.957 &
\cellcolor{LBMColor!7}0.063 &
\cellcolor{LBMColor!18}7.020 \\
\bottomrule
\end{tabular}
\vspace{1pt}
\caption{\textbf{Parameter and compute split for ABC-DiT and ABC-VLA.} TFLOPs are training Tera-FLOPs per sample, using a $3\times$ forward-pass estimate for forward plus backward and counting one multiply-add as two FLOPs. ABC-VLA uses eight independent diffusion noise/timestep draws per sample. ABC-DiT uses a single diffusion draw per sample because extra draws would repeatedly evaluate its large DiT action head; in ABC-VLA, by contrast, the expensive VLM pass is shared across draws and only the small action head is repeated. 
}
\label{tab:flops-param-breakdown}
\end{table}

\section{Real Robot Setup}
\label{sec:real_robot_setup}

Reproducible robot learning requires a platform that is both capable and easy to replicate. We build ABC around a low-cost bimanual YAM workstation and a lightweight robot codebase for data collection, evaluation, and deployment.

\subsection{Hardware}

\begin{figure}[t]
    \centering
    \begin{subfigure}[t]{0.31\linewidth}
        \centering
        \includegraphics[width=\linewidth]{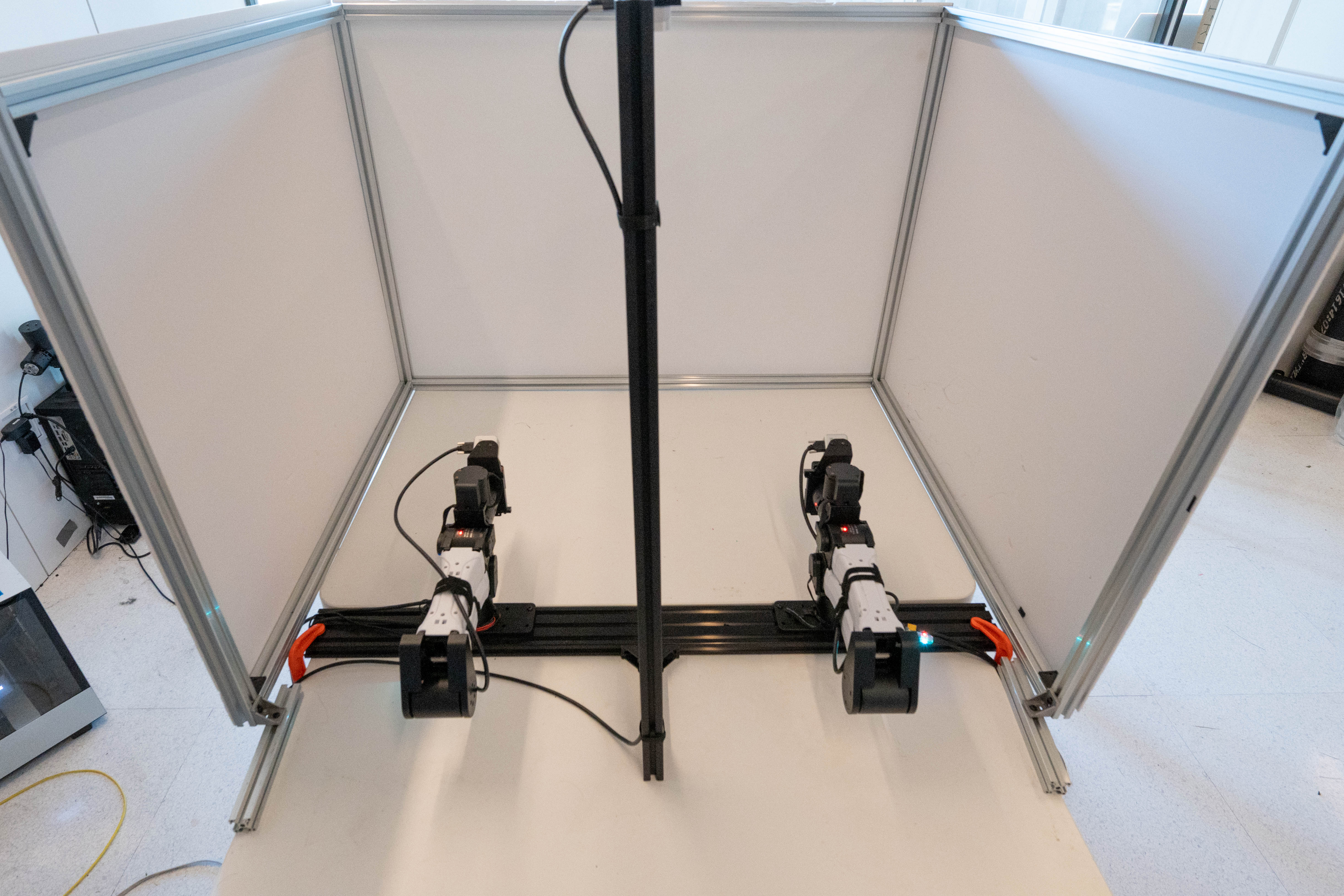}
        \label{fig:workstation}
        \caption{}
    \end{subfigure}
    \hfill
    \begin{subfigure}[t]{0.31\linewidth}
        \centering
        \includegraphics[width=\linewidth]{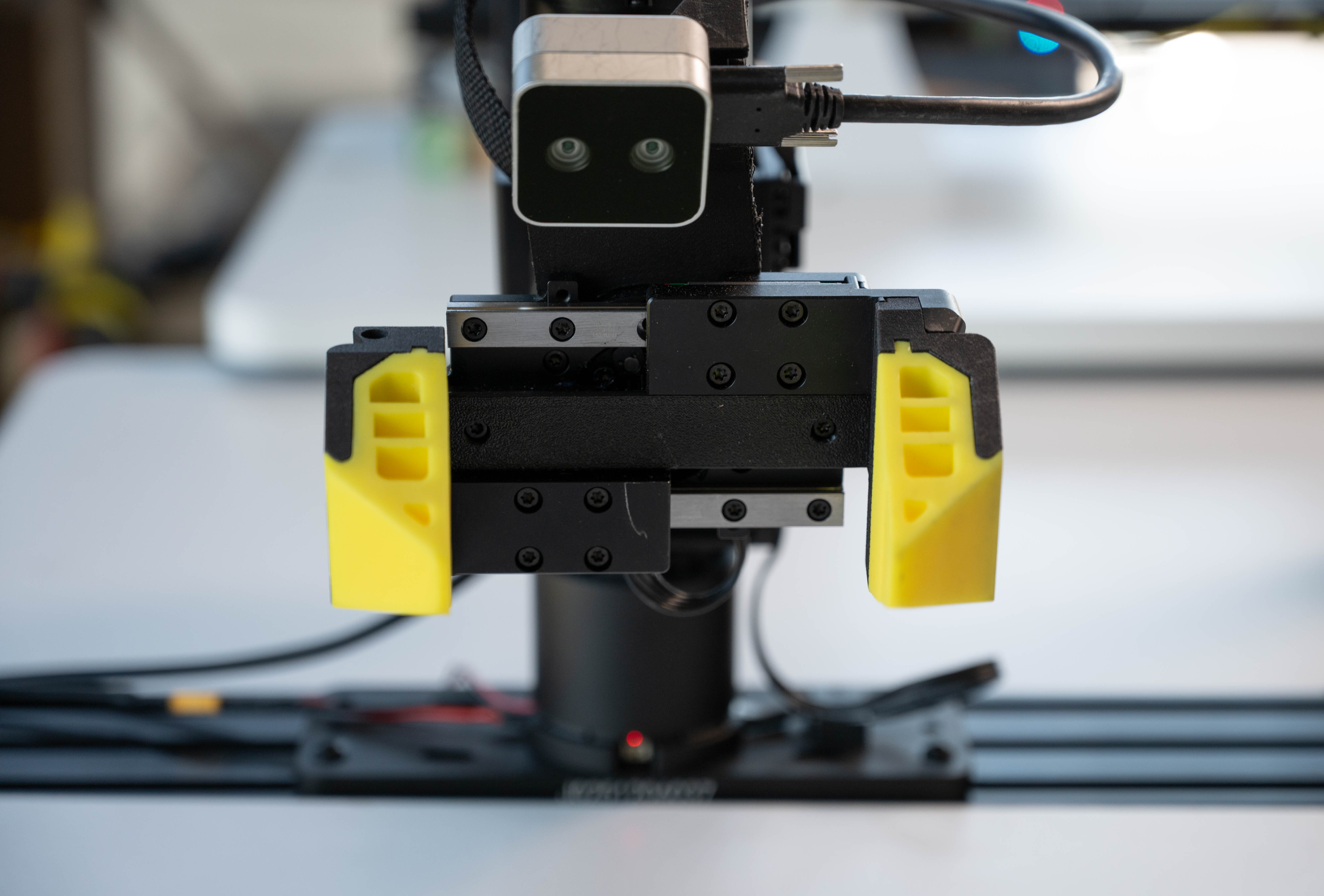}
        \label{fig:gripper-close-up}
        \caption{}
    \end{subfigure}
    \hfill
    \begin{subfigure}[t]{0.31\linewidth}
        \centering
        \includegraphics[width=\linewidth, trim=648 0 0 0, clip]{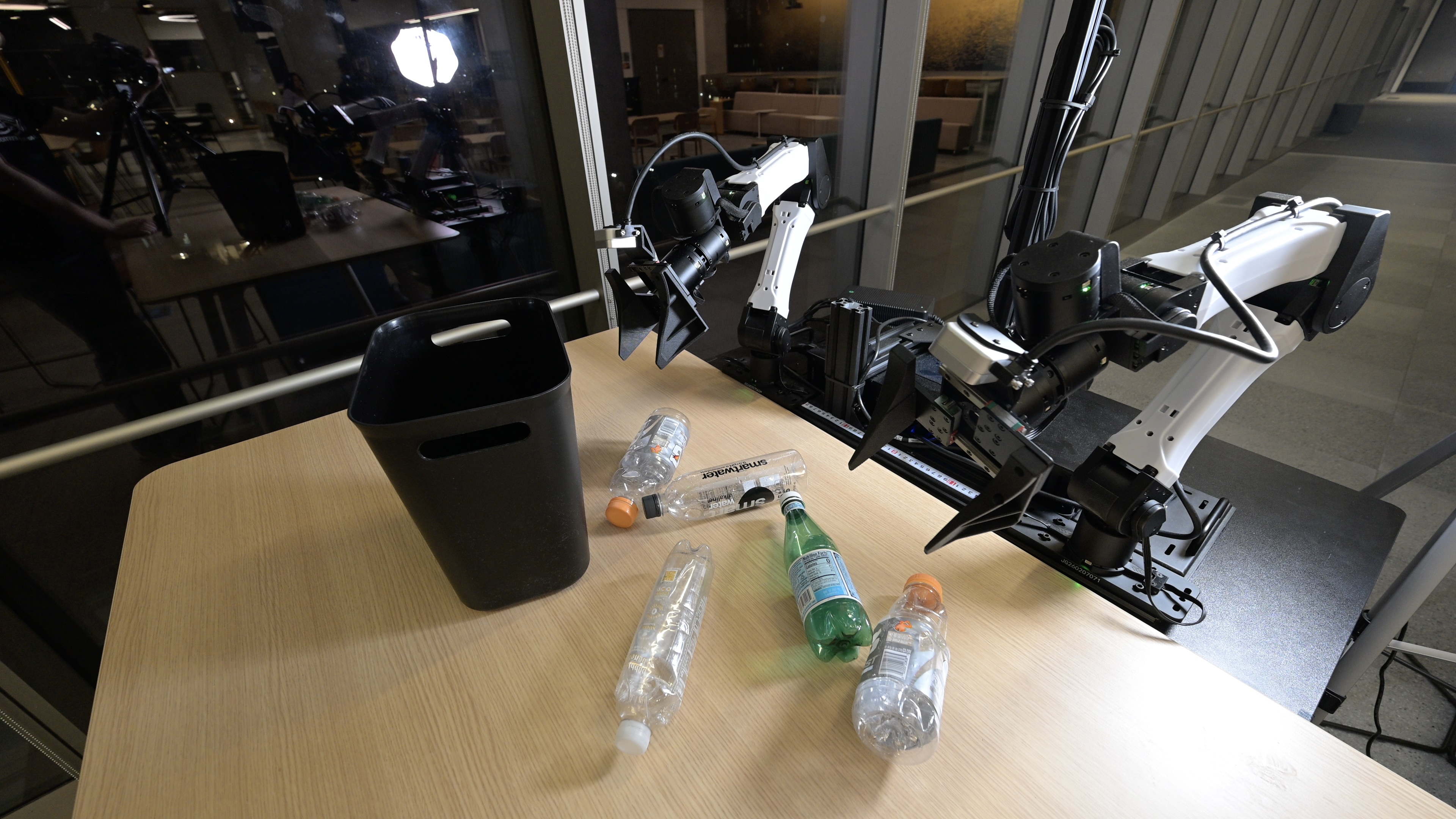}
        \label{fig:placeholder}
        \caption{}
    \end{subfigure}
    \caption{\textbf{Hardware setup.} (a) Bimanual YAM
    workstation with two 6-DoF arms, three RealSense D405 cameras, and
    white enclosure walls. (b) The FlexPoint gripper used in a subset of our dataset. (c) Policies using our dataset can be deployed in in-the-wild settings.}
    \label{fig:setup}
\end{figure}

All data collection and evaluation are conducted on a bimanual platform comprising two \textit{I2RT YAM} arms~\cite{i2rt_sdk}: openly available, low-cost 6-DoF manipulators. The two arms are mounted parallel to each other on a table and enclosed on three sides by white walls; the enclosure narrows background diversity in the training distribution, but (i) reduces variance in evaluation from incidental visual disturbances and (ii) isolates fine-manipulation learning from the confound of background generalization.
We note, however, that despite using only data collected in this caged setup, many of our policies transfer outside the cage and can be deployed in some in-the-wild settings.
We use three Intel RealSense D405 cameras, streaming at 30Hz,  with one being mounted to give a third-person view above the workstation while the other two are mounted as wrist cameras. %

\subsection{Real Robot Code}
\label{sec:real_robot_code}
We implement a custom framework to communicate with our robot, based on the ZMQ backend \cite{zeromq}, which will also be released. Similar to ROS~\cite{ros}, it uses a decentralized node system that relies on PUB/SUB communications. It is meant to be a very lightweight library that is easily extensible to new hardware. Each module of the infrastructure (e.g. GELLO leader arm, YAM follower arm, RealSense cameras, etc.) has its own dedicated node that ticks at a predefined rate and can subscribe and publish to different topics. In order to maintain the tick rate, each node uses Python's \texttt{time.sleep} to coarsely wait for the desired period, and for the final 300 microseconds it employs a busy-spin loop to improve timing precision.

\section{Inference Optimization}
\label{app:inference_optimization}
\begin{figure}[t]
  \centering
  \resizebox{\linewidth}{!}{%
      \input{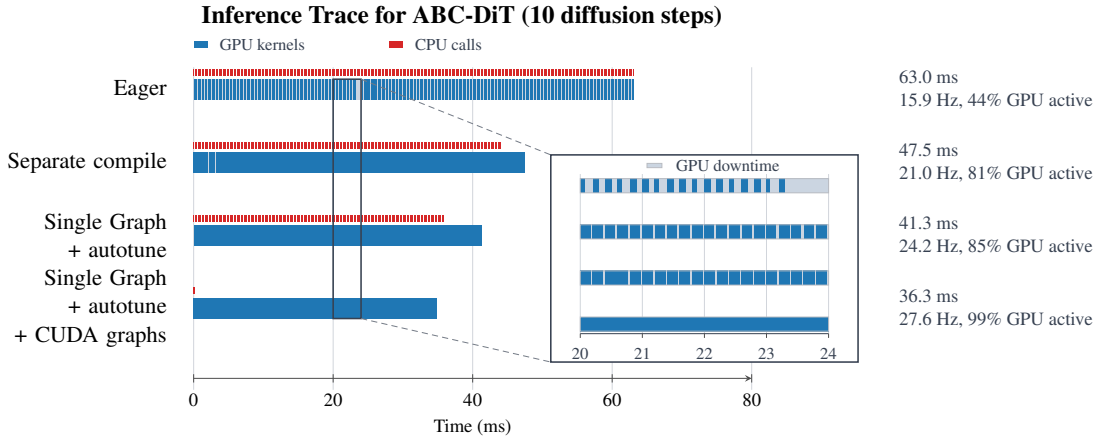}%
  }
  \caption{\textbf{Inference optimizations for ABC-DiT.}
  Profiling traces for ABC-DiT inference with 10 diffusion denoising steps. All timings are measured on an NVIDIA GeForce RTX 5090.}
  \label{fig:abc-dit-inference-trace}
\end{figure}

\begin{figure}[t]
  \centering
  \resizebox{\linewidth}{!}{%
      \input{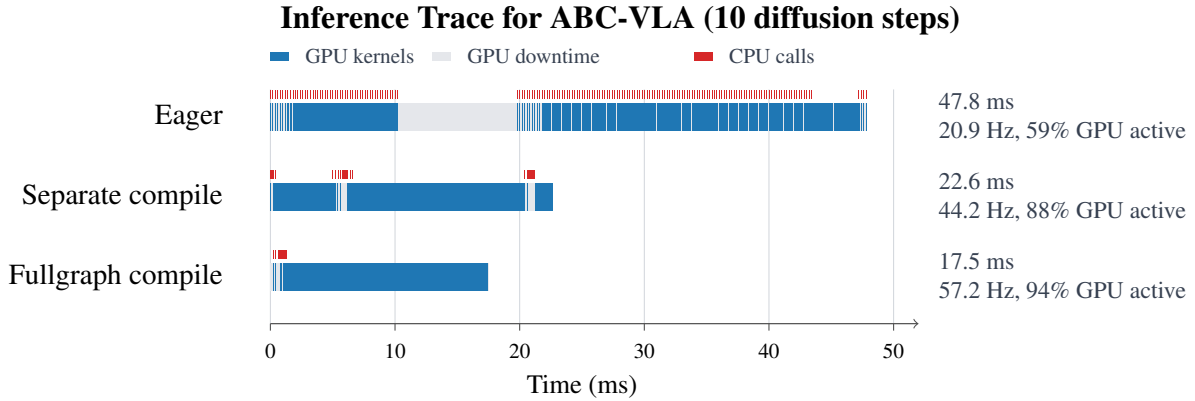}%
  }
  \caption{\textbf{Inference optimizations for ABC-VLA.}
  Profiling traces for ABC-VLA inference with 10 diffusion denoising steps. All timings are measured on an NVIDIA GeForce RTX 5090.}
  \label{fig:abc-vla-inference-trace}
\end{figure}

We implement various optimizations to reduce the inference latency of both ABC-DiT and ABC-VLA. For the purposes of this section, all timing numbers are measured on an NVIDIA GeForce RTX 5090. The main techniques include conducting inference at bf16, caching the visual embeddings/VLM features to avoid recomputation every diffusion step, and leveraging \texttt{torch.compile}. While the first two are fairly ubiquitous, \texttt{torch.compile} performance can vary depending on how it is used. When wrapping the model in a compile block, we pass in \texttt{fullgraph=True}. This causes compilation to fail on graph breaks, including host synchronization points, which lets us identify and modify code paths where there are CPU-side dependencies that limit compiler optimization. We use this to iteratively go through our model and remove any synchronization points until the entire inference pass compiles as a single graph. We show how the various optimizations reduce latency for ABC-DiT in Figure~\ref{fig:abc-dit-inference-trace} and for ABC-VLA in Figure~\ref{fig:abc-vla-inference-trace}.

For ABC-DiT, our baseline (eager mode) uses bf16 and cached visual features, but no PyTorch compilation. Looking at the trace, we can see there are many CPU-side kernel launch calls throughout the inference duration of the model, as well as the fact that there are many small GPU kernels -- each one carrying its own launch overhead. The first optimization involves just compiling the DINO model and the DiT head separately. Compilation reduces the latency by fusing multiple kernels together. Fusing kernels means that there are fewer that need to be individually launched, hence lower CPU overhead. The next optimization on top of that is to compile both the DINO encoder and DiT head together instead of creating two separate graphs, and using kernel autotuning. This lets the compiler benchmark various kernel implementations and launch configurations for the actual tensor shapes on the target GPU. The last optimization does all of this with CUDA graph capture. This is a technique whereby memory locations are kept static, and during the initial graph capture phase, PyTorch records all of the kernels launched by the model. Subsequent operations replay those kernels rather than having the CPU launch each kernel separately. This dramatically reduces overhead which can be seen by how the number of CPU-side kernel launch calls diminishes. In the end, these optimizations improve latency from 63 ms down to 36.3 ms.

For ABC-VLA, the eager baseline also uses bf16 with cached embeddings but exhibits a major gap due to GPU downtime. The separate compile involves compiling the SigLIP, VLM, and DiT head separately with max autotuning and CUDA graph capture. Here, the main overhead is seen between these three main modules. When the full inference path is compiled together, much of the overhead is then eliminated. Most of the remaining CPU-side overhead occurs at the beginning likely due to input staging and graph setup. These optimizations combined bring down the latency from 47.8 ms down to 17.5 ms. One perhaps counterintuitive aspect of the inference results is that ABC-VLA is faster than ABC-DiT despite having more than twice as many total parameters. This is because the repeated component during diffusion is much smaller for ABC-VLA given the AdaLN design choice. In ABC-DiT, the diffusion head has roughly 1.93B parameters and must be evaluated at every diffusion step. In ABC-VLA, the VLM is large, but it is only run once to produce cached features; the diffusion head itself has only 45M parameters and is then run for the 10 diffusion steps. As a result, the cost of ABC-VLA inference is dominated by a single VLM forward pass plus lightweight diffusion-head evaluations, which is faster than running a roughly 2B-parameter transformer 10 times in ABC-DiT.

\section{Single-task Finetuning}
\label{app:single-task-finetuning}

\subsection{Task description and evaluation}
We provide here detailed task description, evaluation protocol, and finetuning
recipe for the downstream experiments in Section~\ref{sec:finetuning}. We provide the task names and descriptions for the tasks we used for finetuning. 
\begin{enumerate}
    \item \textbf{Take credit cards out of the card holder:} The task includes opening up a wallet, scrolling across it to find a filled slot, taking the card out of the slot, placing it in the container and continuing. This task requires high precision and fine motor skills.
    \item \textbf{Sort legos into the container:} The task includes picking up LEGO pieces, breaking them up into atomic pieces if they are stuck together, and then placing them into the correctly colored container. This task requires strong visual representations for planning.
    \item \textbf{Insert pens into pen caps:} The task is to pick up the pen and the cap with different arms and then insert the pen into the cap with the correct orientation and then press hard. 
    \item \textbf{Unscrew bottle caps:} The task is to pick up a bottle and unscrew its cap using the other arm.
\end{enumerate}
We provide a time budget of 2 minutes for each task, except for credit card removal, for which we used a budget of 3 minutes. Table \ref{tab:rubrics} provides the evaluation rubric we used for these tasks.

\subsection{Finetuning recipe}
ABC-DiT is cheaper to train and therefore is our base model for single-task finetuning except for the credit-card removal task. We find that performance for ABC-VLA finetuning is markedly better for this task compared to that of ABC-DiT. We finetune
both model families with AdamW using a learning rate of $1\mathrm{e}{-5}$ for both the vision backbone and the action expert, with weight decay
$1\mathrm{e}{-2}$. We use a batch size of $1{,}440$ for ABC-DiT and $192$ for
ABC-VLA using 1 H200 node for training. We keep all other training hyperparameters the same as in pretraining. We train for $200{,}000$ gradient steps and visually inspect checkpoints every $50{,}000$ steps, selecting the best checkpoint for evaluation.

\section{DAgger Infrastructure}
\label{app:dagger_infra}

Most robot teleoperation data, including the data in this paper, is collected using leader arms \citep{zhao2023learning, wu2024gello}. While active leader arms exist \cite{liu2025factr, oh2026factr2}, they tend to be more complex to set up, requiring system identification and advanced control algorithms to make them smooth. Meanwhile, alternative methods like VR require additional hardware. We release a new method to collect intervention data using no additional hardware other than the conventional passive leader-arm setup. The way this works is that when an intervention happens, we run forward kinematics on the pose of the leader and follower arms, and continuously on the leader arms during the course of the intervention. We then take the delta in SE3 pose between the intervention pose and the current pose of the leader arms. We then compute the pose made up of the pose of the follower end effector + the leader SE3 delta for the follower arms, and then use IK to get follower arm joints. We use mink for IK \cite{Zakka_Mink_Python_inverse_2026}. The infrastructure to do such interventions will be released along with the rest of the code for this project. 

We implement a state machine for data collection to facilitate collection of data. When an episode starts, the policy is executing. The operator can press a button (either on the leader arms if they have such a button, or a cheap commercially available foot pedal) to transition to the teleoperation mode. At any time the operator can press another button to transition back to the policy.

When transitioning from an intervention back to the policy, we condition the policy on the last few actions executed during the intervention as part of the RTC prefix. This not only makes the transition smoother, but makes it easier for the operator to select a specific mode out of the diffusion policy if it is not obvious from the Markovian observations alone what the next action should be (e.g. whether to retry, say, folding a flap or go to the next action during box folding).

\section{Evaluation Details}
\label{sec:evaluation}
When we report the progress of policies on various tasks, we use the rubric in Table~\ref{tab:rubrics}. Each score is normalized by the maximum score in order to give a measure of progress.

  \begin{table*}[htbp]
    \centering
    \small
    \setlength{\tabcolsep}{4pt}
    \renewcommand{\arraystretch}{1.15}
    \caption{\textbf{Evaluation task rubrics.} \textit{Max Score} is the maximum achievable score for the task; reported progress is the achieved score normalized by this maximum. \textit{Timeout} is the maximum trial duration.}
    \label{tab:rubrics}
    \begin{tabularx}{\textwidth}{@{}
      >{\raggedright\arraybackslash}p{4.8cm}
      >{\raggedright\arraybackslash}p{2.0cm}
      c
      >{\raggedright\arraybackslash}p{1.6cm}
      >{\raggedright\arraybackslash}X
    @{}}
      \toprule
      Task & Setting & Max Score & Timeout & Scoring Rubric \\
      \midrule
      \texttt{throw plastic bottles in bin} &
      6 bottles, 1 bin &
      6 &
      120 s &
      +1 for each bottle inside the bin \\

      \texttt{insert pens into the pen caps} &
      1 pen, 1 cap &
      3 &
      120 s &
      +1 for picking up both pen and cap; +2 for inserting the pen into the cap \\

      \texttt{load plates into tabletop dish rack} &
      2 plates, 1 dish rack &
      6 &
      180 s &
      Per plate: +1 for picking up the plate; +1 for placing it inside the dish rack; +1 for placing it correctly \\

    \texttt{sorting legos into containers} &
      2 green/yellow bricks, 2 bins (with bricks in them) &
      4 &
      120 s &
      +1 for each properly deposited lego piece \\
      
    \texttt{unscrew bottle caps} &
      1 bottle&
      3 &
      120 s &
      +1 for lifting the bottle, +2 for unscrewing the cap \\
      
    \texttt{take credit cards out of the card holder} &
      1 wallet filled with credit cards and 1 container &
      5 &
      180 s &
      +1 for opening wallet, +1 for removing card, +1 for placing in container \\
    \texttt{folding paper box} &
      1 paper box &
      5 &
      180 s &
      +1 for picking up box, +1 for folding each of the sides, +1 for closing the box lid, +1 for folding in the flaps\\

    \texttt{folding tshirt pile and stacking} &
      1 t-shirt &
      5 &
      390 s &
      +1 pick \& place; +1 flatten; +1 for $\geq$1 proper fold; +1 full fold (roughly square); +1 placed in corner. Fold quality rated low/mid/high; on repeated flatten--fold cycles, score only the best fold (no double counting) \\
    
    \bottomrule
    \end{tabularx}
  \end{table*}

\section{Policy Conditioning}
\label{app:conditioning}

We study three mechanisms for shaping the behavior of our policies at inference time. \emph{Operator-ID conditioning} (Section~\ref{sec:operator-conditioning}) addresses stylistic multimodality across teleoperators, allowing the policy to be steered towards the chosen operator's style at deployment. \emph{Action prefix conditioning} (Section~\ref{sec:action-prefix-conditioning}) examines how the length of the executed-action prefix used by Real-Time Chunking trades off motion smoothness against visual responsiveness. \emph{Subtask conditioning} (Section~\ref{sec:subtask-conditioning}) looks at resolving temporal ambiguity within a task via language prompting. Across all three, the common theme is shifting the policy's effective conditioning balance toward the signals most relevant to the current decision.

All experiments in this section are evaluated on the t-shirt folding task. We chose it
because it highlights all three conditioning mechanisms we study: substantial variation in
operator quality (Section~\ref{sec:operator-conditioning}); a multi-stage structure with visually
similar states in trajectories, so that the correct next action is often ambiguous from a
single observation (Section~\ref{sec:subtask-conditioning}); and a large pool of demonstrations
($\sim$268.64 hours total). Every t-shirt-folding episode is subtask-annotated, compared to
44\% of ABC-130K overall, which makes it particularly suited to the subtask-conditioning
study.

For these experiments we use the CLIP cross-attention variant of ABC-DiT rather than the DINOv3 backbone used elsewhere: once finetuned, the CLIP cross-attention policy was the most reliable on this task, so that the conditioning effects we report are not confounded by a weak policy. The base model is pretrained on the internal 7,000-hour corpus, then finetuned on a filtered subset of the released t-shirt-folding data: for each operator we keep only the later-collected half of their episodes, which tend to reflect more skillful execution as operators grow more practiced over the course of collection. We finetune for 100k steps and evaluate on two unseen shirts (small/blue and large/yellow) with five randomly crumpled initial configurations per shirt, each initialized at approximately the same workspace location ($n=10$ per condition, maximum trial duration 390~s). Each rollout is scored on a 5-point rubric (Section~\ref{sec:evaluation}); fold quality is rated low~/~medium~/~high. All evaluation data, including the rollouts' actions, states, and image observations, are released for reader inspection.

\begin{figure}[!t]
    \centering
    \includegraphics[width=1.0\linewidth]{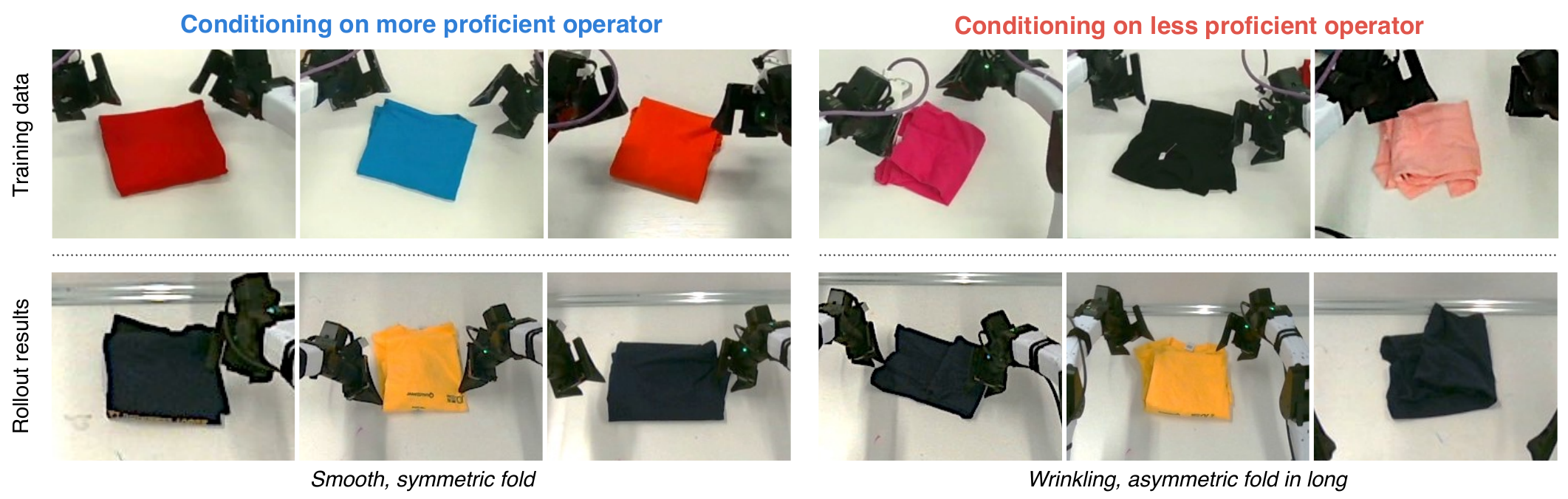}
    \caption{\textbf{Impact of Operator conditioning.}
\textit{Top row}: folds from training demonstrations contributed by Op-0 (more proficient operator) left and Op-1 (less proficient operator) right. \textit{Bottom row}: representative folds from the same trained policy under Op-0 (left) and Op-1 (right) conditioning. Although the policy is the same, varying only the operator prompt produces qualitatively distinct fold outcomes matching the training distribution of the conditioning operator. Quantitative results across all rollouts are reported in Table~\ref{tab:operator-conditioning}.}

    \label{fig:operator_data_rollout}
\end{figure}

\subsection{Operator-ID conditioning}
\label{sec:operator-conditioning}

The natural baseline for operator-ID conditioning is to filter the training data to the demonstrations of a single high-quality operator. We evaluate this baseline by finetuning the ABC-DiT model on the subset of the t-shirt folding corpus contributed by the highest-volume operator (op~0, 19.5 hours across 1{,}183 episodes). This filtered checkpoint achieves a mean rubric score of 3.3 with only 2 of 10 completions and a mean trial duration of 369 s (Table~\ref{tab:operator-conditioning}); it performs worse on every metric than the unconditioned all-operator baseline (3.8, 4/10, 302 s). We trained the filtered model to 30k steps rather than 100k, after observing overfitting when trained longer.

Operator-ID conditioning allows test-time conditioning on desired behavior without data filtering. We finetune the ABC-DiT policy on the all-operator corpus with operator-ID text appended to the task prompt (see Figure~\ref{fig:operator_data_rollout} for representative training demonstrations from Op-0 and Op-1). We evaluate three inference conditions on the same checkpoint:
\begin{enumerate}
    \item Conditioning on Op-0, the highest-volume operator, whose training demonstrations are characterized by short, deliberate execution (mean episode duration 59 s)
    \item Conditioning on the task prompt alone, which marginalizes over operators by reverting to the training-time dropout target
    \item Conditioning on Op-1, a long-duration operator with 226 episodes of training data whose demonstrations average 205 s per episode and exhibit lower fold quality
\end{enumerate}

Marginalized inference improves over the unconditioned baseline, indicating that operator-ID conditioning at training time is strictly beneficial even when the operator channel is not used at inference. Conditioning on Op-0 yields the highest-quality folds with the highest completion rate, while conditioning on Op-1 both yields longer duration and lower-quality folds. We also note that Op-0-conditioned rollouts frequently re-execute earlier subtasks, e.g., re-flattening a partially folded garment that fails some implicit quality threshold, consistent with stricter stage-transition criteria in that operator's training demonstrations; Op-1-conditioned rollouts advance through transitions more readily and reproduce that operator's lower fold quality (Figure~\ref{fig:operator_data_rollout}).

\begin{table}[t]
\centering
\caption{\textbf{Operator-ID conditioning affects t-shirt folding behavior at inference time.} T-shirt folding evaluation across the operator-filtered baseline and operator-ID conditioning inference modes. Rubric scores are out of 5; mean trial duration is reported in minutes:seconds with timeouts counted as the maximum trial duration (390 s); quality counts (H/M/L) are among completed trials. All conditioning rows use the same trained checkpoint, varying only the inference-time prompt.}
\label{tab:operator-conditioning}
\small
\setlength{\tabcolsep}{3pt}
\begin{tabularx}{\linewidth}{@{}p{0.26\linewidth}p{0.24\linewidth}cccc@{}}
\toprule
Training data & Inference prompt & Mean score & Completions & Mean time & H / M / L \\
\midrule
\multicolumn{6}{l}{\textit{Baselines}} \\
All-operator corpus & task only & 3.8 & 4 / 10 & 302 s & 2 / 2 / 0 \\
Op-0 filtered corpus & task only & 3.3 & 2 / 10 & 369 s & 1 / 0 / 1 \\
\midrule
\multicolumn{6}{l}{\textit{Operator-ID conditioning (same checkpoint)}} \\
All-operator corpus & task + Op-0 & \textbf{4.6} & \textbf{8 / 10} & \textbf{237 s} & \textbf{4 / 3 / 1} \\
All-operator corpus & task only (marginalize) & 4.4 & 6 / 10 & 247 s & 1 / 4 / 1 \\
All-operator corpus & task + Op-1 & 4.0 & 5 / 10 & 277 s & 1 / 1 / 3 \\
\bottomrule
\end{tabularx}

\vspace{0.5em}
\end{table}

\subsection{Action prefix conditioning}
\label{sec:action-prefix-conditioning}

\begin{figure}[!t]
    \centering
    \begin{subfigure}[t]{0.49\linewidth}
        \centering
        \includegraphics[width=\linewidth]{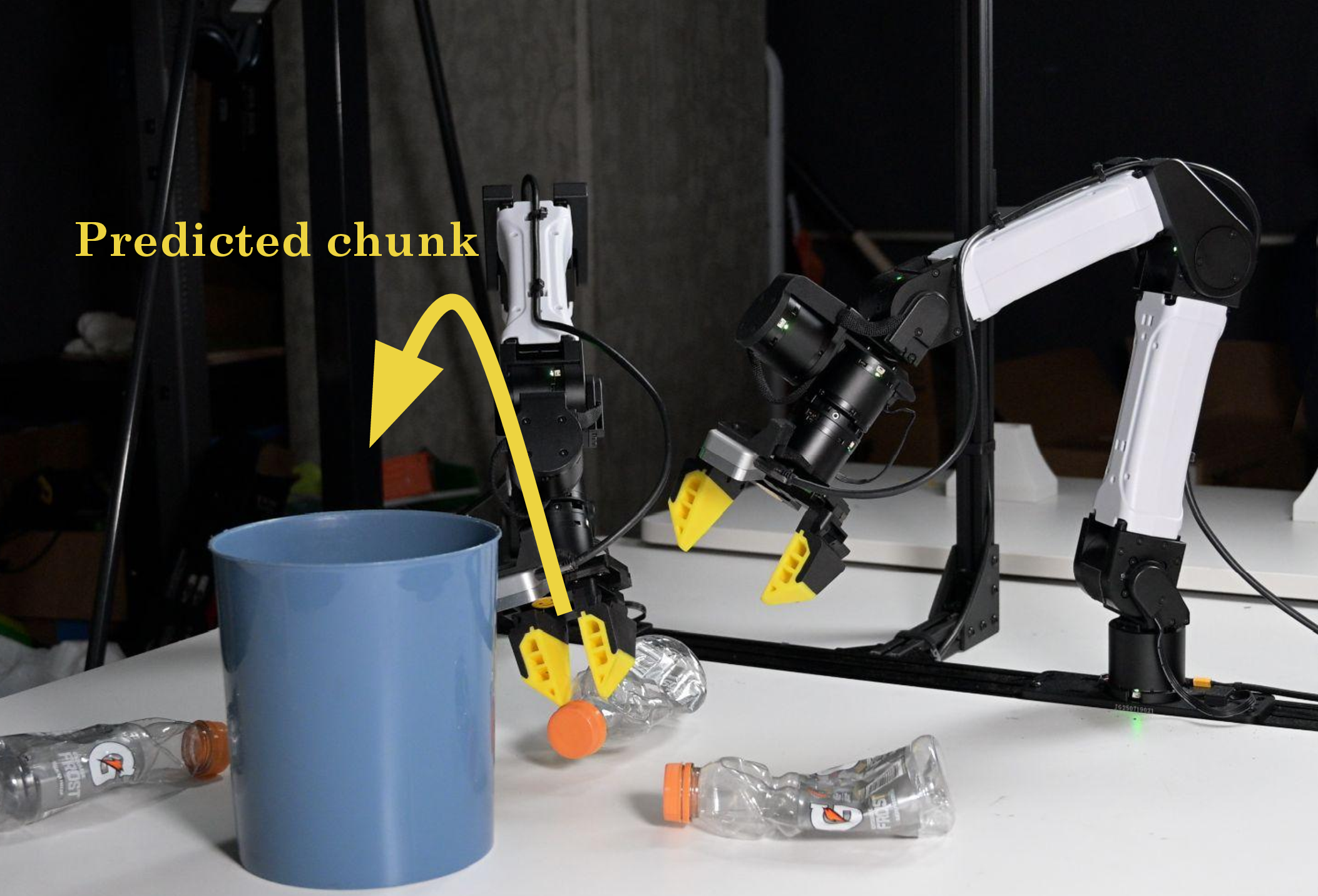}
        \label{fig:impact-of-prefixing-predicted}
    \end{subfigure}
    \hfill
    \begin{subfigure}[t]{0.49\linewidth}
        \centering
        \includegraphics[width=\linewidth]{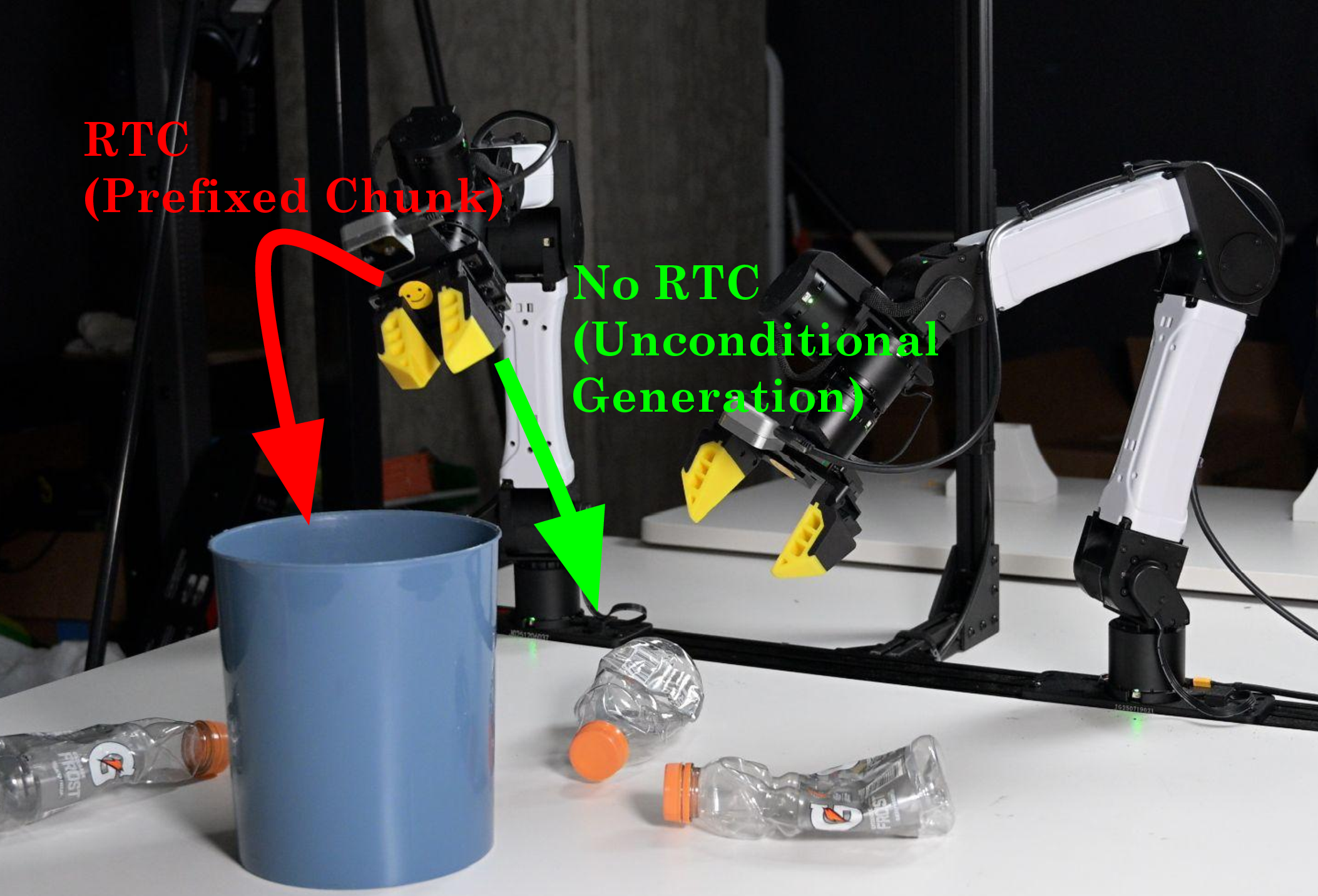}
        \label{fig:impact-of-prefixing-rtc}
    \end{subfigure}
    \caption{\textbf{Prefix conditioning can suppress visual responsiveness.} Prefixing chunk generation on recent actions can bias the generated motion toward the recently executed trajectory, while the unconditioned generation remains more responsive to the current visual observation. \textbf{Left:} In the first chunk, the robot fails to pick the bottle. \textbf{Right:} In the next chunk, we compare RTC prefix-conditioned and unconditional generations. In this example, the model conditioned on the prefix continues the trajectory into the bin, ignoring the visual observation that it has missed the grasp, while the unconditional generation, while less smooth, correctly attends to the visual observations.}
    \label{fig:impact-of-prefixing}
\end{figure}

All of our policies are trained with training-time Real-Time Chunking~\citep{black2025trainingtimeactionconditioningefficient}, in which each generated action chunk is conditioned on a prefix of recently-executed actions. The action prefix improves inter-chunk motion smoothness by anchoring the next chunk to the trajectory currently being executed.

We observe that while this improves the smoothness of policies, it can also hurt their performance, and policies can tend to overfit to the prefix and partially ignore visual observations in favor of continuing the existing trajectory. Quantitative evidence of this is provided in Table \ref{tab:action-prefix}, and a qualitative explanation of the phenomenon is illustrated in Figure \ref{fig:impact-of-prefixing}. We find that the simplest method to combat this is reducing the prefix length at the cost of some smoothness; more thorough characterization of the issue and investigation of other methods to mitigate these effects are left to future work.

\begin{table}[t]
\centering
\caption{\textbf{Shorter action prefixes improve t-shirt folding performance.} Effect of inference-time action prefix length on t-shirt folding under Op-0 conditioning, using the same trained checkpoint as the Op-0 row in Table~\ref{tab:operator-conditioning}. Same evaluation protocol; only the prefix length differs. Mean trial duration is reported in seconds with timeouts counted as the maximum (390 s); quality counts (H/M/L) are among completed trials.}
\label{tab:action-prefix}
\small
\begin{tabular}{@{}lcccc@{}}
\toprule
Action prefix & Mean score & Completions & Mean time & H / M / L \\
\midrule
Prefix = 4 & 3.9 & 5 / 10 & 309 s & 3 / 2 / 0 \\
Prefix = 1 & \textbf{4.6} & \textbf{8 / 10} & \textbf{237 s} & \textbf{4 / 3 / 1} \\
\bottomrule
\end{tabular}
\end{table}

\subsection{Subtask conditioning}
\label{sec:subtask-conditioning}

Multi-stage manipulation tasks introduce a second form of multi-modality, orthogonal to the operator-stylistic variation in Section~\ref{sec:operator-conditioning}. Depending on which stage of a task the policy is in, the same task prompt corresponds to distinct action distributions. Task-prompt conditioning alone often cannot disambiguate these stages from a single visual observation without history, because intermediate and terminal states can look similar, while the correct next action depends on which stage the task has actually reached. For example, a crumpled t-shirt and a completed fold could appear similar depending on the lighting, cloth texture, fold quality, etc., so a policy conditioned on the task prompt alone may regrasp an already-folded shirt, flatten it, and begin folding again---and sometimes begins re-flattening partway through a fold. Subtask conditioning addresses this by exposing the current subtask through language, using the same independent-dropout scheme as operator-ID conditioning; at inference the subtask can be set by a high-level planner, classifier, or human operator, letting the policy commit to the stage-appropriate action distribution rather than averaging over stage-conditional modes.

To evaluate subtask prompting, we train a small subtask classifier---whose architecture is inspired by the stage-aware reward model of SARM~\cite{chen2026sarm}---on ABC-130K, and use it to set the subtask prompt for our ABC-DiT policy at inference time. We initialize each trial with a half-folded shirt to probe the failure mode described above. Unlike the earlier rollouts, which ended once the shirt was placed in the corner, we hold the arms in the reset pose for 10\,s after placement to observe the policy's subsequent behavior. Without subtask prompting, the policy exhibits the re-flattening failure in 5 of 10 trials; in 4 of these cases, it regrabs an already-folded or already-placed t-shirt. With subtask prompting, the policy does not exhibit this behavior, although it fails to complete the task within the time limit in one trial due to a missed grasp and subsequent recovery errors.

\subsection{Conditioning during pretraining}
\label{sec:conditioning-pretraining}
We stochastically condition our models with all of the above approaches during pretraining. We set the length of the action prefix to condition the policy by uniformly sampling a value from 0 to 7 for every training example. We condition on the operator metadata and subtask annotations, each with a probability of 0.2. Since subtask metadata is present only for a subset of the data, we use it for 20\% of the time it is actually available.

\end{document}